\documentclass{article} 
\usepackage{iclr2027_conference,times}

\usepackage[utf8]{inputenc} 
\usepackage[T1]{fontenc}    
\usepackage{hyperref}       
\usepackage{url}            
\usepackage{booktabs}       
\usepackage{amsfonts}       
\usepackage{nicefrac}       
\usepackage{microtype}      
\usepackage{xcolor}         

\usepackage{subcaption}
\usepackage{xcolor}
\usepackage{graphicx}   
\usepackage{amsmath}

\usepackage{amsthm}
\newtheorem{lemma}{Lemma}[section]

\usepackage{multirow}
\usepackage{float}

\title{Predict Backbone, Graft Dynamics: A Decoupled  Paradigm for Time Series Forecasting}

\author{
Lin Zhang\textsuperscript{1}
\quad
Zilin Li\textsuperscript{2}
\quad
Jingmei Zhao\textsuperscript{1}
\quad
Jiayang Wu\textsuperscript{1}
\\[0.6em]
{\normalfont\small
\textsuperscript{1}Southwestern University of Finance and Economics}
\\
{\normalfont\small
\textsuperscript{2}Georgia Institute of Technology}
\\[0.4em]
{\normalfont\small
\texttt{1230202j1001@smail.swufe.edu.cn}
\quad
\texttt{zli3381@gatech.edu}}
\\
{\normalfont\small
\texttt{zhaojingmei@swufe.edu.cn}
\quad
\texttt{225081200014@smail.swufe.edu.cn}}
}

\iclrfinalcopy

\begin{document}

\maketitle

\lhead{Preprint}

\begin{abstract}
Modern long-term forecasters are typically optimized for point-wise accuracy, which rewards dominant low-frequency backbones but often suppresses high-utility local dynamics such as abrupt variations, transient surges, and turning points. 
We propose \emph{Predict Backbone, Graft Dynamics}, a decoupled forecasting paradigm that separates stable structural prediction from grounded dynamic restoration. 
We instantiate it as \textbf{GRAFT}, where a base forecaster predicts the reliable trajectory backbone, and a dedicated grafting path restores historically recurrent dynamics through retrieval-based evidence, phase-aware release, channel-wise control, and restoration-cooperative optimization. 
Unlike generic post-hoc revision methods that correct residual errors, GRAFT formulates the second stage as grounded dynamic grafting, where historical response evidence is calibrated into transferable dynamics and selectively grafted onto the backbone under phase-, channel-, and confidence-aware constraints. 
Across the full evaluation grid of seven benchmarks, two metrics, four horizons, and six backbones, 
GRAFT improves over the corresponding backbone in 87\% of all comparisons, 
with average MSE/MAE reductions of 4.6\%/3.9\%.
Dynamic-window analyses show that GRAFT restores local dynamics while preserving the stable backbone structure. 
These results suggest a new principle for high-utility forecasting: predict the backbone, then graft the dynamics.
\end{abstract}



\section{Introduction}
\newlength{\panelheight}
\setlength{\panelheight}{8.2cm}   

Long-term time series forecasting (LTSF) has advanced through larger model capacity and increasingly specialized architectural inductive biases, yielding strong point-wise accuracy on benchmarks such as ETT and Electricity~\citep{wu2021autoformer,wu2023timesnet,zhou2021informer}. 
Yet recent empirical studies suggest that progress is entering a saturation regime, with gains in MSE or MAE often confined to the third decimal place~\citep{wang2025accuracy}. 
We argue that this saturation reflects more than diminishing returns from architectural refinement; it exposes a mismatch between point-wise objectives and forecasting utility. 
Because global error minimization is dominated by high-energy, low-frequency periodic components, forecasters are rewarded for preserving stable backbone structures while receiving weak incentives to recover sparse local dynamics. 
As a result, models can achieve strong aggregate accuracy but still produce smooth, conservative predictions that miss structural breaks, transient surges, and turning points, which are often decisive for practical forecasting value.

\begin{figure}[t]
    \centering
    \begin{minipage}[t][\panelheight][t]{0.64\textwidth}
        \vspace{0pt}
        \centering
        \parbox[c][2.55cm][c]{0.045\linewidth}{%
            \centering
            \rotatebox{90}{\scriptsize \textbf{Across models}}%
        }%
        \begin{minipage}[c]{0.94\linewidth}
            \centering
            \begin{subfigure}[t]{0.24\linewidth}
            \centering
            \includegraphics[width=\linewidth]{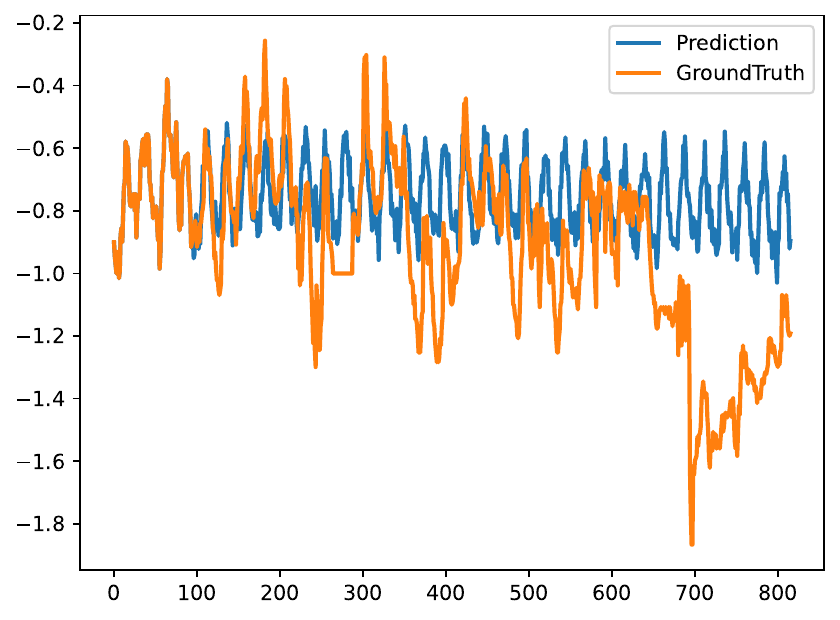}
            \caption{\tiny\begin{tabular}[t]{@{}l@{}} PatchTST \\[-1pt] MSE:0.516 \end{tabular}}
            \end{subfigure}
            \hfill
            \begin{subfigure}[t]{0.24\linewidth}
            \centering
            \includegraphics[width=\linewidth]{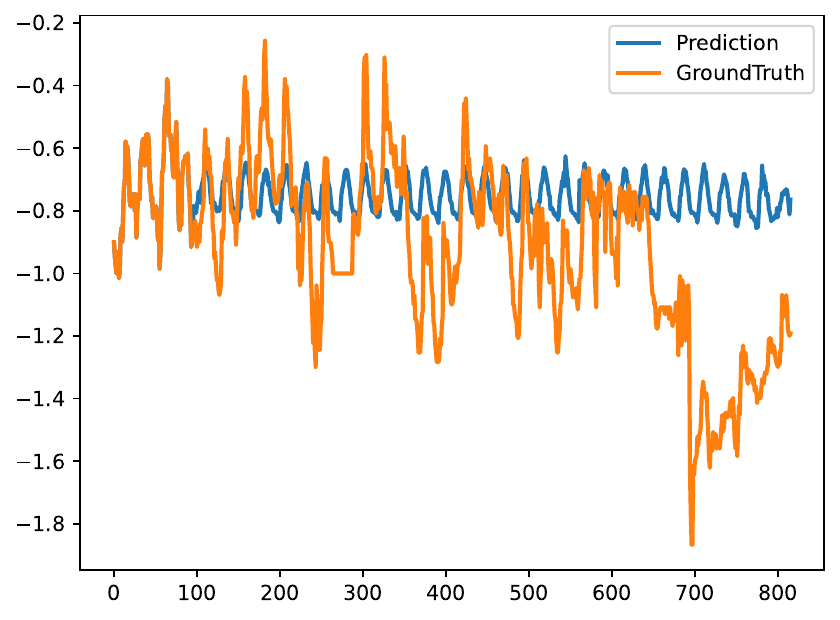}
            \caption{\tiny\begin{tabular}[t]{@{}l@{}} TimesNet \\[-1pt] MSE:0.518 \end{tabular}}
            \end{subfigure}
            \hfill
            \begin{subfigure}[t]{0.24\linewidth}
            \centering
            \includegraphics[width=\linewidth]{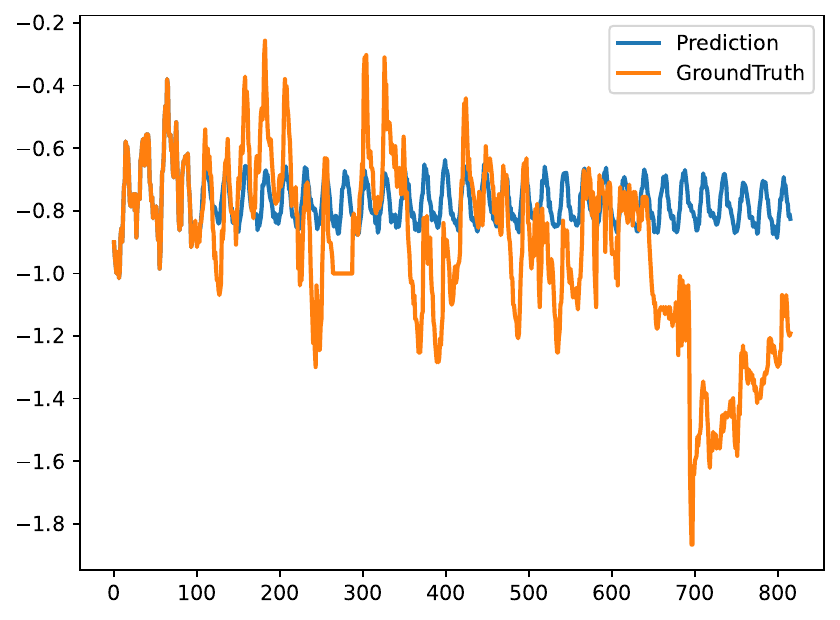}
            \caption{\tiny\begin{tabular}[t]{@{}l@{}} TimeMixer \\[-1pt] MSE:0.497 \end{tabular}}
            \end{subfigure}
            \hfill
            \begin{subfigure}[t]{0.24\linewidth}
            \centering
            \includegraphics[width=\linewidth]{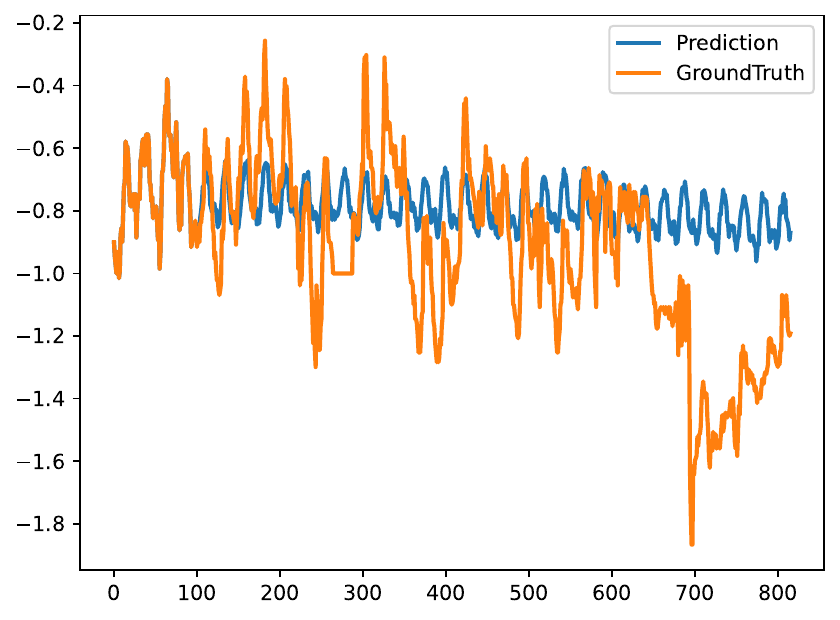}
            \caption{\tiny\begin{tabular}[t]{@{}l@{}} Times2D \\[-1pt] MSE:0.489 \end{tabular}}
            \end{subfigure}
            \end{minipage}
        \vspace{1pt}
        \noindent\rule{0.98\linewidth}{0.25pt}
        \vspace{1pt}

        \parbox[c][2.55cm][c]{0.045\linewidth}{%
            \centering
            \rotatebox{90}{\scriptsize \textbf{Across datasets}}%
        }%
        \begin{minipage}[c]{0.94\linewidth}
            \centering
            \begin{subfigure}[t]{0.24\linewidth}
                \centering
                \includegraphics[width=\linewidth]{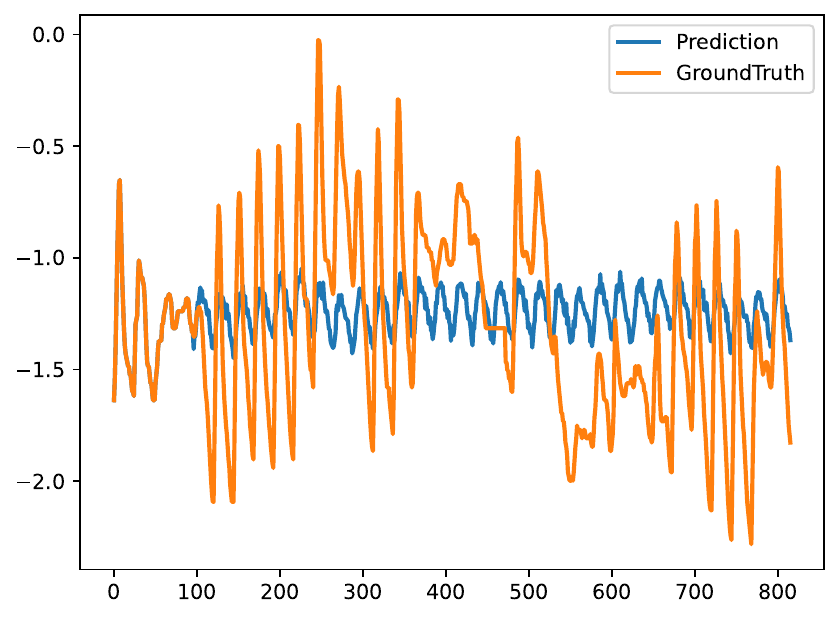}
                \caption{\tiny{ETTh2}}
            \end{subfigure}
            \hfill
            \begin{subfigure}[t]{0.24\linewidth}
                \centering
                \includegraphics[width=\linewidth]{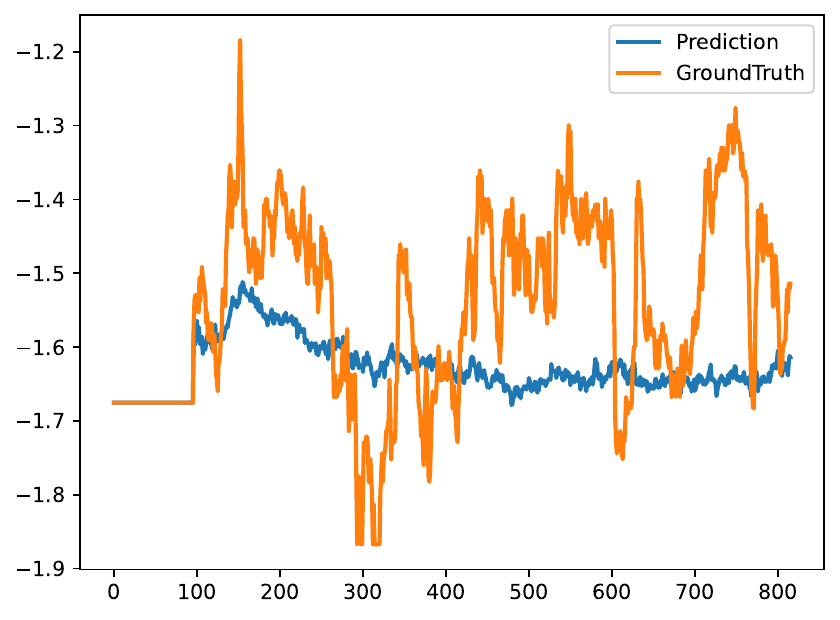}
                \caption{\tiny{ETTm1}}
            \end{subfigure}
            \hfill
            \begin{subfigure}[t]{0.24\linewidth}
                \centering
                \includegraphics[width=\linewidth]{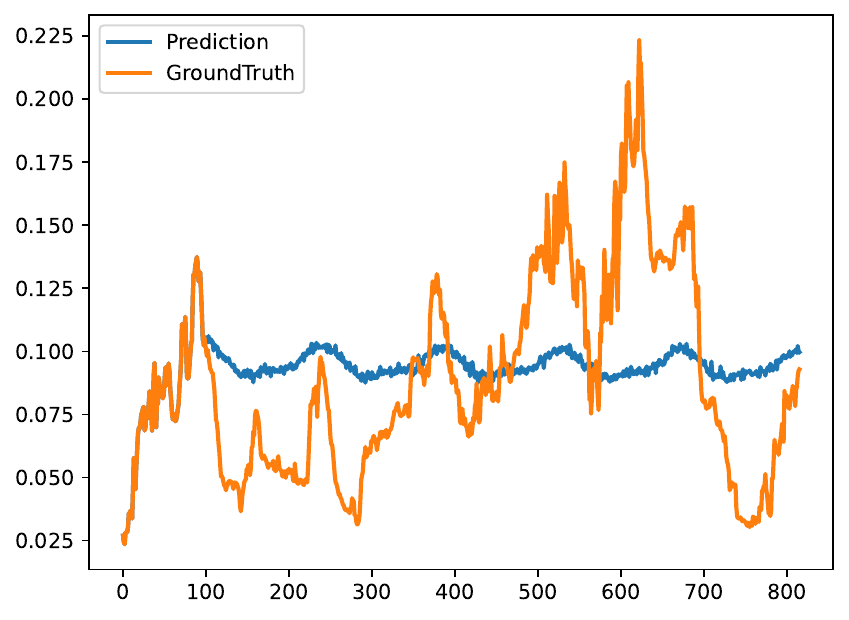}
                \caption{\tiny{Weather}}
            \end{subfigure}
            \hfill
            \begin{subfigure}[t]{0.24\linewidth}
                \centering
                \includegraphics[width=\linewidth]{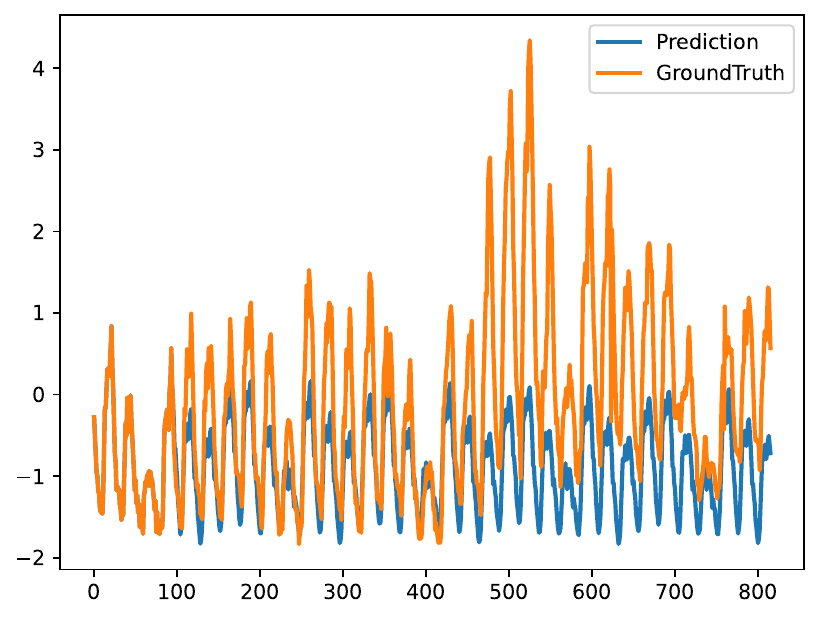}
                \caption{\tiny{Electricity}}
            \end{subfigure}
        \end{minipage}
        \vspace{0pt}
        {\footnotesize \textbf{(1)} The MSE/MAE trap across models and datasets.}
    \end{minipage}
    \hfill
    \begin{minipage}[t][\panelheight][t]{0.34\textwidth}
        \vspace{0pt}
        \centering

        \raisebox{0.15cm}{\includegraphics[
            width=0.98\linewidth,
            height=4.8cm,
        ]{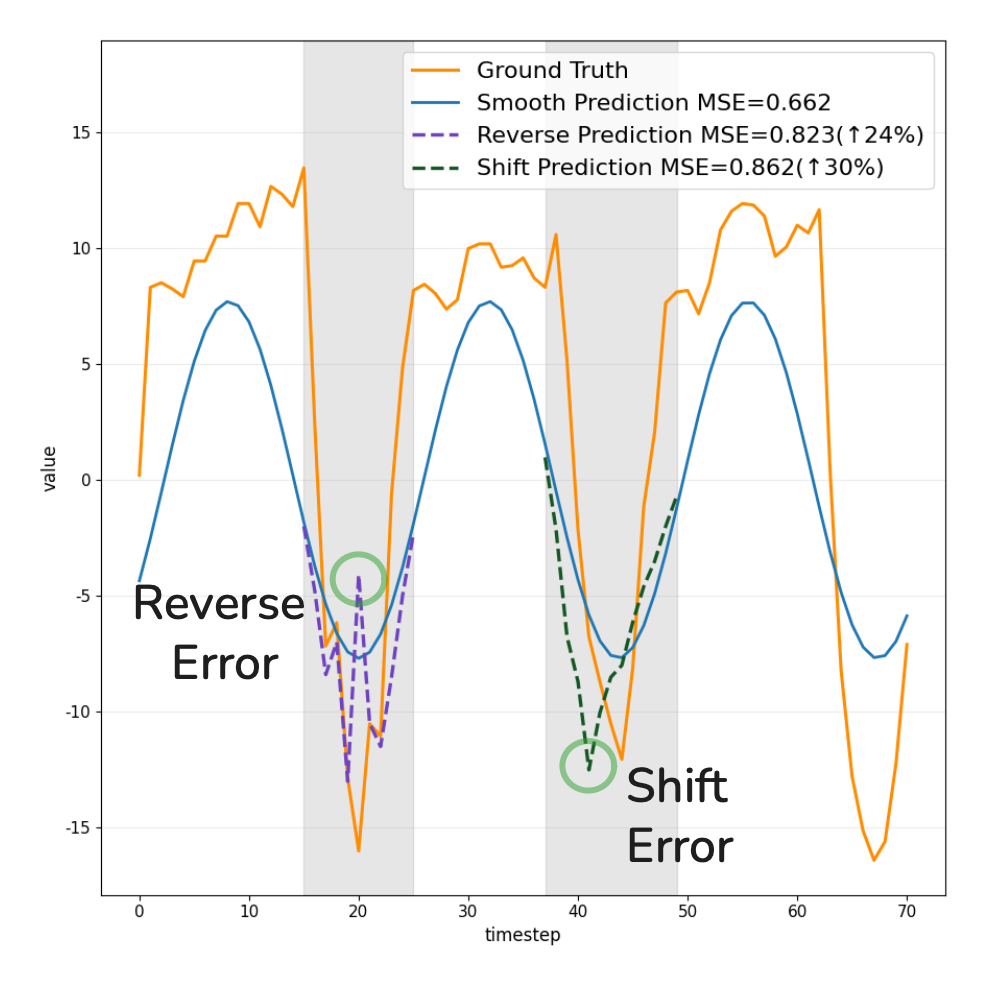}}
        \vspace{48pt}
        {\footnotesize \textbf{(2)} Point-wise loss penalty.}
    \end{minipage}
    \vspace{-75pt}
\caption{
The MSE/MAE trap and its optimization cause. \textbf{Left}: Point-wise optimized forecasters smooth out local mutations across architectures (on the same ETTh1 segment) and datasets. Orange denotes the ground truth, and blue denotes the prediction. \textbf{Right}: Misaligned dynamic recovery can be penalized more heavily than smooth conservative forecasting.
}
\label{fig:motivation_and_penalty}
\vspace{-2em}
\end{figure}

This discrepancy gives rise to a \emph{performance-utility paradox}: 
forecasters can achieve competitive aggregate errors while failing to recover the local events that matter most in practice. 
Fig.~\ref{fig:motivation_and_penalty} illustrates that this phenomenon is not tied to a specific architecture or dataset. 
Across diverse backbones and benchmarks, predictions tend to preserve stable low-frequency patterns but smooth out high-frequency dynamics, structural mutations, and abrupt regime shifts. 
Such behavior reflects an implicit preference for steady-state dynamics over rare but high-impact events. 
In power grids, for example, sudden load surges are often more consequential than routine daily cycles~\citep{zhang2014review}; in financial markets, trend reversals can be more decision-critical than regular price oscillations~\citep{jiang2023re,liqingtkde2018}. 
Although these localized mutations contribute little to aggregate point-wise error, they often carry disproportionate forecasting utility and risk relevance~\citep{oppenheim1999discrete,fei2025amplifier,nematirad2025times2d}.

Why does this failure persist across architectures and datasets?
We attribute it to two coupled biases in the standard end-to-end forecasting paradigm: sparse dynamics are \emph{under-rewarded} by the objective and \emph{high-risk} to optimize.
First, \textit{spectral reward bias} makes dominant periodic structures the most loss-efficient target. 
Standard LTSF benchmarks often have skewed power spectral densities, with most energy concentrated in low-frequency seasonal components. 
Together with the frequency principle of neural networks, which suggests that low-frequency patterns are learned earlier during optimization~\citep{rahaman2019spectral}, point-wise objectives naturally reward models for fitting stable periodic backbones. 
In contrast, local dynamics occupy short temporal regions; although information-dense and practically important, they contribute little to aggregate MSE or MAE while requiring higher modeling complexity. 
They are therefore easily treated as negligible noise.
Second, \textit{optimization penalty bias} makes imperfect dynamic recovery risky. 
Local mutations are sparse and uncertain in timing, direction, and magnitude~\citep{Lin2026GFMixer}. 
The right part of Fig.~\ref{fig:motivation_and_penalty} shows that fitting an abrupt transition improves the objective only when its temporal location and amplitude are accurately aligned. 
A minor synchronization error can trigger a double penalty: the model is punished for missing the true event and for generating a false one~\citep{wang2025time}. 
Thus, point-wise losses can penalize imperfect dynamic predictions more heavily than smooth conservative ones, biasing gradient descent toward a variance-minimizing mean~\citep{le2019shape}.

Together, these biases explain why stronger end-to-end architectures may still converge to conservative forecasts: sparse dynamics offer limited aggregate reward but incur high optimization risk (see Appendix~\ref{app:mse_reward_risk}). 
The bottleneck is therefore not merely model capacity, but the entanglement of stable backbone prediction and sparse dynamic recovery under a single point-wise objective.

This tension reveals a fundamental misalignment between optimization incentives and forecasting utility. 
Standard point-wise paradigms drive predictors toward a conservative compromise: they prioritize the low-frequency backbone for stable loss reduction, while suppressing high-value local mutations to avoid disproportionate penalties. 
Overcoming this bottleneck requires decoupling two conflicting roles, namely stable structural anchoring and dynamic transition recovery, rather than further escalating architectural complexity.

We therefore propose \emph{Predict Backbone, Graft Dynamics} (PBGD), a decoupled forecasting paradigm that reframes prediction as a structured composition of distinct temporal processes. 
Under this paradigm, a backbone forecaster first provides a reliable structural anchor, while a dedicated grafting path restores suppressed local dynamics through historically grounded and constraint-controlled dynamic transfer. 
We instantiate PBGD with \textbf{GRAFT}, a model-agnostic framework that retrieves dynamically homologous historical responses, calibrates their transferable dynamics under temporal-phase and channel-wise constraints, and selectively grafts them onto the backbone forecast through confidence-aware release. 
Unlike generic post-hoc correction, GRAFT does not merely learn an unconstrained residual error mapping.
Instead, it formulates the second stage as grounded dynamic grafting, where historical evidence is transformed into controlled local dynamic compensation.
Our contributions are summarized as follows:
\begin{itemize}
\setlength{\itemsep}{2pt}
\setlength{\parskip}{0pt}
\setlength{\parsep}{0pt}

\item \textbf{A utility-oriented reformulation of LTSF.} 
We revisit long-term forecasting through the lens of optimization-utility mismatch and identify a systematic \emph{high-score but low-utility} phenomenon: point-wise objectives incentivize modern forecasters to preserve dominant periodic backbones while suppressing sparse but decision-critical local dynamics. 
This analysis motivates a new forecasting principle: \emph{Predict Backbone, Graft Dynamics}.

\item \textbf{A decoupled framework for grounded dynamic grafting.}
We propose \textbf{GRAFT}, a decoupled forecasting framework that combines backbone prediction with constrained dynamic grafting.
It retrieves homologous historical responses, transforms them into transferable local dynamics, and grafts them onto the backbone forecast under phase-, channel-, and confidence-aware constraints, enabling standalone forecasting as well as model-agnostic enhancement of existing LTSF models.

\item \textbf{Extensive validation of accuracy and dynamic recovery.}
Across the full evaluation grid, GRAFT improves over the corresponding backbone in 85.7\% of comparisons, with average MSE/MAE reductions of 3.88\%/3.00\%. 
Dynamic-window analyses and ablations show that these gains come from targeted recovery of surges and turning points, rather than indiscriminate high-frequency injection.
\end{itemize}

\vspace{-0.8em}
\section{Related Work}
\vspace{-0.8em}

Existing efforts most relevant to GRAFT can be grouped into two lines: 
\emph{architectural temporal disentanglement} and \emph{backbone-level enhancement with dynamic-aware correction}. 
The former designs end-to-end architectures that decompose or mix temporal components within the predictor, while the latter improves forecasting through retrieval, adaptation, alignment, correction, compact temporal representations, or objective reshaping. 
GRAFT is related to both lines, but differs by explicitly decoupling stable backbone prediction from grounded local dynamic restoration.

\textbf{Architectural temporal disentanglement.}
Methods such as Autoformer~\citep{wu2021autoformer}, FEDformer~\citep{zhou2022fedformer}, and ETSformer~\citep{woo2022etsformer} decompose time series into trend and seasonal components, while recent models such as TimeMixer~\citep{wang2023timemixer} further exploit multi-scale mixing or auxiliary reconstruction. 
These designs improve the modeling of dominant temporal structures such as trends, seasonality, and periodic patterns. 
However, they are still trained as unified end-to-end forecasters under point-wise objectives. 
Consequently, the optimization signal can remain dominated by stable low-frequency patterns, leaving sparse but high-utility local dynamics under-recovered. 
GRAFT takes a complementary route: it keeps the backbone forecast as a structural anchor and introduces a separate grafting path to restore local dynamics suppressed by point-wise optimization.

\textbf{Backbone-level enhancement and dynamic-aware correction.}
Recent methods increasingly improve forecasting without fully replacing the backbone, either by augmenting predictions through retrieval, adaptation, alignment, correction, or compact temporal bases, or by reshaping the training objective to better capture dynamic behavior. 
For example, GTR~\citep{cao2026gtr} extends the input context through global temporal retrieval,
RAFT~\citep{han2025retrieval} retrieves historical patterns as additional forecasting information, PIR~\citep{liu2025improving} performs instance-aware forecast revision, TimeAlign~\citep{hu2026bridging} improves prediction consistency via temporal alignment and auxiliary reconstruction, 
$\delta$-Adapter~\citep{liang2026the} applies lightweight input- or output-side adaptation, and TimeBase~\citep{huang2025timebase} represents temporal evolution with compact bases. 
Meanwhile, shape- and time-aware losses, label-alignment strategies, and frequency-aware objectives have shown that point-wise supervision may lead to temporal distortion, misaligned trajectories, or neglected low-energy components~\citep{fei2025amplifier,wang2025time,le2019shape}. 
These works show that retrieval, adaptation, correction, and objective redesign can improve forecasting.
GRAFT instead asks a more fundamental question: \emph{what does point-wise forecasting systematically fail to recover?}
Sparse local dynamics are weakly rewarded by global objectives and difficult to infer from the current observation alone, while homologous historical responses provide grounded evidence of their evolution.
GRAFT therefore treats dynamic recovery as a distinct forecasting responsibility: the backbone preserves the structural trajectory, while a separate grafting path restores recurrent local dynamics.
Thus, retrieval and enhancement are not contributions by themselves, but mechanisms supporting a new division of forecasting labor---\emph{predict the backbone, then graft the dynamics}.

\section{Methodology}
\begin{figure}[ht]
    \centering
    \includegraphics[width=1\linewidth,height=0.20\textheight]{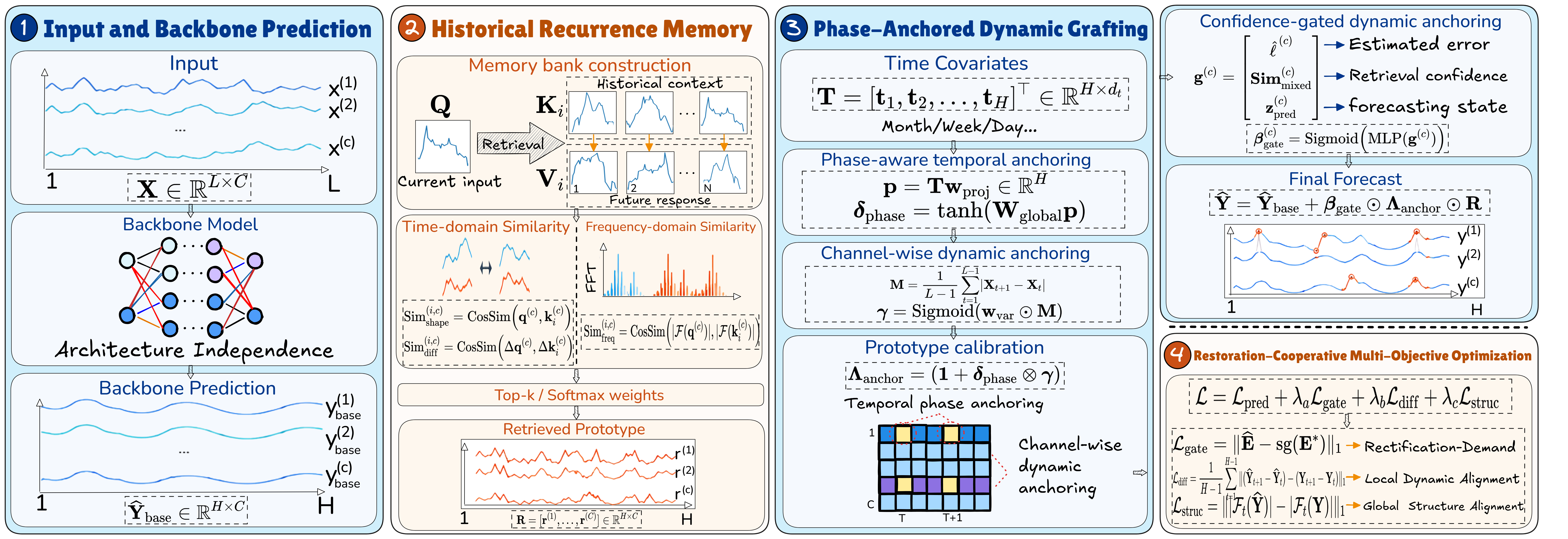}
    \caption{Overview of GRAFT.}
    \label{fig:method}
    \vspace{-1.0em}
\end{figure}

Given a look-back window 
$\mathbf{X}\in\mathbb{R}^{L\times C}$, long-term forecasting predicts 
$\mathbf{Y}\in\mathbb{R}^{H\times C}$ over a horizon of length $H$. 
We denote by 
$\mathbf{T}=[\mathbf{t}_1,\ldots,\mathbf{t}_H]^\top\in\mathbb{R}^{H\times d_t}$ 
the future temporal covariates, where each $\mathbf{t}_h$ encodes the temporal phase of the $h$-th future step, such as hour-of-day, day-of-week, or other calendar features. 
In GRAFT, $\mathbf{T}$ serves as a set of future phase anchors for determining when historical dynamics are temporally compatible with the prediction horizon.

GRAFT instantiates the principle of \emph{Predict Backbone, Graft Dynamics}. 
A backbone forecaster first predicts a stable structural trajectory, while a dedicated grafting branch restores historically grounded local dynamics that are often suppressed by point-wise objectives:
\begin{equation}
\widehat{\mathbf{Y}}
=
\widehat{\mathbf{Y}}_{\mathrm{base}}
+
\widehat{\mathbf{Y}}_{\mathrm{dyn}},
\qquad
\widehat{\mathbf{Y}}_{\mathrm{base}}
=
f_{\theta}(\mathbf{X}),
\qquad
\widehat{\mathbf{Y}}_{\mathrm{dyn}}
=
\mathcal{G}_{\phi,\psi}
\left(
\mathbf{X},
\mathbf{T},
\widehat{\mathbf{Y}}_{\mathrm{base}};
\mathcal{M}
\right),
\end{equation}
where $\mathcal{M}$ is a historical recurrence memory and 
$\mathcal{G}_{\phi,\psi}$ is a retrieval-grounded dynamic grafting operator.
Specifically, GRAFT retrieves dynamically homologous historical responses from $\mathcal{M}$ and aggregates them into a response prototype:
\begin{equation}
\mathbf{R}
=
\operatorname{Retrieve}
\left(
\mathbf{X};\mathcal{M}
\right),
\qquad
\mathbf{R}\in\mathbb{R}^{H\times C}.
\end{equation}
The prototype $\mathbf{R}$ provides historical response evidence, but is not directly used as an alternative forecast. 
It is first mapped into a transferable dynamic basis:
\begin{equation}
\mathbf{R}_{\mathrm{dyn}}
=
\mathcal{T}_{\psi}
\left(
\mathbf{R}\mid \mathbf{X},\mathbf{T}
\right)
=
\boldsymbol{\Lambda}_{\mathrm{anchor}}
\left(
\mathbf{X},\mathbf{T};\psi
\right)
\odot
\mathbf{R}.
\end{equation}
Here, $\mathcal{T}_{\psi}$ maps the retrieved prototype $\mathbf{R}$ into a transferable dynamic basis $\mathbf{R}_{\mathrm{dyn}}$ under the current input and future temporal context.

Finally, GRAFT releases the transferable dynamic basis as the predicted dynamic component through a confidence-aware gate:
\begin{equation}
\widehat{\mathbf{Y}}_{\mathrm{dyn}}
=
\boldsymbol{\beta}_{\mathrm{gate}}
\left(
\mathbf{X},
\widehat{\mathbf{Y}}_{\mathrm{base}};\phi
\right)
\odot
\mathbf{R}_{\mathrm{dyn}}
=
\underbrace{
\left[
\boldsymbol{\beta}_{\mathrm{gate}}
\left(
\mathbf{X},
\widehat{\mathbf{Y}}_{\mathrm{base}};\phi
\right)
\odot
\boldsymbol{\Lambda}_{\mathrm{anchor}}
\left(
\mathbf{X},\mathbf{T};\psi
\right)
\right]
}_{\text{learned dynamic transfer field}}
\odot
\mathbf{R}.
\label{equ:4}
\end{equation}

Here, $\boldsymbol{\beta}_{\mathrm{gate}}\in(0,1)^{H\times C}$ 
is a confidence-aware release gate that decides whether dynamic restoration should be activated based on the backbone prediction state and query-induced retrieval reliability, which is detailed in Sec.~\ref{sec:phase_anchoring}.
The product 
$\boldsymbol{\beta}_{\mathrm{gate}}\odot\boldsymbol{\Lambda}_{\mathrm{anchor}}$ forms a learned dynamic transfer field, which controls the compatibility and release strength of the retrieved evidence at each future step and channel. 
Thus, $\widehat{\mathbf{Y}}_{\mathrm{dyn}}$ is not the retrieved prototype itself, but a gated and transformed dynamic prediction grounded in $\mathbf{R}$.
Thus, GRAFT turns historical recurrence into controlled dynamic restoration rather than copying retrieved trajectories or learning unconstrained residual corrections. 
Retrieval determines \emph{what} dynamics may be useful, temporal phase information constrains \emph{when} they are compatible, channel-wise control regulates \emph{how much} evidence can be transferred, and confidence gating decides \emph{whether} intervention is needed. 
We next describe Historical Recurrence Memory, Phase-Anchored Dynamic Grafting, and Restoration-Cooperative Optimization.

\subsection{Historical Recurrence Memory}
\label{sec:historical_recurrence}

Although abrupt variations, short-term deviations, and turning points are often smoothed out by point-wise optimization, they are not necessarily random. 
In real-world time series, similar historical contexts may be followed by homologous future responses~\citep{ning2026tsrag}. 

\textbf{Memory bank.}
Inspired by strategies that explicitly retrieve historical instances~\citep{han2025retrieval,liu2025improving}, we organize historical context-response pairs into a key-value memory bank:
\begin{equation}
\mathcal{M}
=
\left\{
(\mathbf{K}_i,\mathbf{V}_i)
\right\}_{i=1}^{N},
\end{equation}
where $N$ denotes the total number of available historical instances, $\mathbf{K}_i\in\mathbb{R}^{L\times C}$ denotes a historical look-back context and 
$\mathbf{V}_i\in\mathbb{R}^{H\times C}$ stores the corresponding full future response.
Given the current input $\mathbf{X}$, we treat it as the query $\mathbf{Q} = \mathbf{X} \in\mathbb{R}^{L\times C}$ and retrieve historical keys that are dynamically homologous to $\mathbf{Q}$.
The associated values then serve as historical responses,
which are aggregated into the retrieved prototype $\mathbf{R}$ for subsequent dynamic grafting.
Note that we store full future responses rather than residualized targets because local dynamics depend on their surrounding level, trend, scale, and phase, whereas residualization commits the memory to the approximation errors of a particular forecaster.
A controlled comparison of the two memory targets is provided in Appendix~\ref{app:full_response_memory}.

\textbf{Pseudo-similarity and homology retrieval.}

Reliable grafting requires historical keys whose associated futures provide transferable dynamic evidence.
Naive point-wise retrieval may select \emph{pseudo-similar} keys that appear close to the query but differ in shape, local variation, or temporal evolution, thereby contaminating the prototype $\mathbf{R}$ with incompatible dynamics.
To retrieve dynamically transferable prototypes, we measure homology between the $c$-th channel of the query $\mathbf{q}^{(c)}\in\mathbb{R}^{L}$ and the $i$-th historical key $\mathbf{k}_i^{(c)}\in\mathbb{R}^{L}$ across three complementary views:

\begin{equation}
\resizebox{\linewidth}{!}{%
$\mathrm{Sim}_{\mathrm{shape}}^{(i,c)} \!=\! \mathrm{CosSim}(\mathbf{q}^{(c)}, \mathbf{k}_i^{(c)}), \;
\mathrm{Sim}_{\mathrm{diff}}^{(i,c)} \!= \!\mathrm{CosSim}(\Delta\mathbf{q}^{(c)}, \Delta\mathbf{k}_i^{(c)}), \;
\mathrm{Sim}_{\mathrm{freq}}^{(i,c)} \!=\! \mathrm{CosSim}(|\mathcal{F}(\mathbf{q}^{(c)})|, |\mathcal{F}(\mathbf{k}_i^{(c)})|)$%
}
\end{equation}
where $\Delta$ is the discrete temporal difference operator and $\mathcal{F}(\cdot)$ denotes the 1D Discrete Fourier Transform.
The three views target different failure modes of raw-value retrieval:
\emph{shape} similarity enforces global contour consistency,
\emph{differential} similarity captures local slope transitions and abrupt-change tendencies,
and \emph{spectral} similarity matches periodic composition while reducing sensitivity to minor temporal shifts.
Finally, we fuse these similarities with channel-specific learnable weights to obtain $\mathrm{Sim}_{\mathrm{mixed}}^{(i,c)}$, and aggregate the top-$K$ associated responses per channel into the prototype $\mathbf{R}\in\mathbb{R}^{H\times C}$.
\begin{equation}
\mathcal I^{(c)}
=\operatorname{TopK}_{i}\!\left(\mathrm{Sim}_{\mathrm{mixed}}^{(i,c)}\right),
\qquad
\mathbf R_{:,c}
=
\sum_{i\in\mathcal I^{(c)}}
\frac{\exp\!\left(\mathrm{Sim}_{\mathrm{mixed}}^{(i,c)}\right)}
{\sum_{j\in\mathcal I^{(c)}}\exp\!\left(\mathrm{Sim}_{\mathrm{mixed}}^{(j,c)}\right)}
\mathbf V_i^{(c)} .
\end{equation}

Details are in Appendix~\ref{app:leakage_free_retrieval} and Appendix~\ref{app:prototype_construction}.



\subsection{Phase-Anchored Dynamic Grafting}
\label{sec:phase_anchoring}

The retrieved prototype $\mathbf{R}$ is not directly added as a residual to the backbone forecast $\widehat{\mathbf{Y}}_{\mathrm{base}}$. 
Since even relevant historical dynamics can be harmful when released at the wrong phase, scale, or confidence level, GRAFT controls \emph{when}, \emph{how much}, and \emph{whether} to graft them.



\textbf{Phase-aware temporal anchoring.}
Dynamic variations are often tied to phase-dependent temporal windows rather than arbitrary future positions.
Thus, we use future time covariates 
$\mathbf{T}=[\mathbf{t}_1,\dots,\mathbf{t}_H]^\top\in\mathbb{R}^{H\times d_t}$ 
not as ordinary auxiliary inputs, but as phase anchors for dynamic release.
We first project these covariates into a latent phase signal 
$\mathbf{p}=\mathbf{T}\mathbf{w}_{\mathrm{proj}}\in\mathbb{R}^{H}$.
However, isolated indicators only indicate where each step lies in the calendar.
Our key insight is that the \emph{expected temporal window} and \emph{modulation direction} of dynamic events are constrained by global phase regularities; e.g., Monday morning may anchor the window of a traffic rush and imply a positive directional surge.
We therefore model cross-step phase dependency as:

\begin{equation}
\boldsymbol{\delta}_{\mathrm{phase}}
=
\tanh
\left(
\mathbf{W}_{\mathrm{global}}\mathbf{p}
\right),
\qquad
\boldsymbol{\delta}_{\mathrm{phase}}\in[-1,1]^H.
\end{equation}
Here, $\mathbf{W}_{\mathrm{global}}\in\mathbb{R}^{H\times H}$ captures horizon-level phase correlations, and $\boldsymbol{\delta}_{\mathrm{phase}}$ acts as a step-wise instruction that amplifies phase-compatible dynamics and suppresses phase-inconsistent release.

\textbf{Channel-wise dynamic anchoring.}
While phase modulation determines \emph{when} dynamics may be released, channels differ in how much correction they can tolerate. 
To avoid over-injecting dynamics into stable variables, we estimate a channel-specific correction budget from recent variation intensity in the look-back window $\mathbf{X}\in\mathbb{R}^{L\times C}$:


\begin{equation}
\mathbf{M}
=
\frac{1}{L-1}
\sum_{t=1}^{L-1}
\left|
\mathbf{X}_{t+1}
-
\mathbf{X}_{t}
\right|,
\qquad
\boldsymbol{\gamma}
=
\operatorname{Sigmoid}
\left(
\mathbf{w}_{\mathrm{var}}\odot\mathbf{M}
\right).
\end{equation}
Here, $\mathbf{M}\in\mathbb{R}^{1\times C}$ measures recent local variation intensity, 
$\mathbf{w}_{\mathrm{var}}\in\mathbb{R}^{1\times C}$ is a learnable channel-wise sensitivity vector, and 
$\boldsymbol{\gamma}\in(0,1)^{1\times C}$ acts as a channel-wise dynamic boundary, enabling stronger release for volatile channels and suppressing unnecessary grafting for stable ones.

Finally, we combine temporal phase anchoring and channel-wise boundary control to calibrate the retrieved prototype:
\begin{equation}
\mathbf{R}_{\mathrm{dyn}}
=
\boldsymbol{\Lambda}_{\mathrm{anchor}}
\odot
\mathbf{R}
=
\left(
\mathbf{1}
+
\boldsymbol{\delta}_{\mathrm{phase}}
\otimes
\boldsymbol{\gamma}
\right)
\odot
\mathbf{R},
\end{equation}
where $\otimes$ denotes the outer product.
This calibration determines where retrieved dynamics are phase-compatible and how much release is permissible for each channel.

\textbf{Confidence-gated dynamic grafting.}
We introduce a \textit{Step-wise Confidence Gate} that independently decides whether the calibrated dynamic basis should be grafted at each future step.

Consistent with the compact notation in Eq.~\ref{equ:4}, 
the gate is implemented using features derived from the current query, the backbone forecast, and the query-induced retrieval confidence. 
For channel $c$, we form
\[
\mathbf{g}^{(c)}
=
[
\hat{\ell}^{(c)}
\parallel
\mathbf{Sim}_{\mathrm{mixed}}^{(c)}
\parallel
\mathbf{z}_{\mathrm{pred}}^{(c)}
],
\]
where $\hat{\ell}^{(c)} \in \mathbb{R}^{1}$ encodes the estimated backbone error, 
$\mathbf{Sim}_{\mathrm{mixed}}^{(c)} \in \mathbb{R}^{K}$ stacks the fused homology scores of the top-$K$ retrieved keys, and 
$\mathbf{z}_{\mathrm{pred}}^{(c)} \in \mathbb{R}^{d_{\mathrm{model}}}$ denotes the projected feature of 
$\widehat{\mathbf{Y}}_{\mathrm{base}}^{(c)}$. 
The retrieval-confidence scores are computed from the same query--memory matching process induced by $\mathbf{X}$, rather than introduced as independent inputs beyond Eq.~\ref{equ:4}.
$\parallel$ denotes feature-wise concatenation.
To achieve fine-grained step-wise control, a shared network directly projects this channel-specific context to the future horizon length $H$:
\begin{equation}
\boldsymbol{\beta}_{\mathrm{gate}}^{(c)}
=
\operatorname{Sigmoid}
\left(
\operatorname{MLP}
(
\mathbf{g}^{(c)}
)
\right),
\qquad
\widehat{\mathbf{Y}}
=
\widehat{\mathbf{Y}}_{\mathrm{base}}
+
\boldsymbol{\beta}_{\mathrm{gate}}
\odot
\mathbf{R}_{\mathrm{dyn}},
\end{equation}
where $\boldsymbol{\beta}_{\mathrm{gate}}^{(c)} \in (0,1)^{H}$ acts as the step-wise gate for channel $c$. 
By stacking these channel-wise outputs, we obtain the full gating matrix $\boldsymbol{\beta}_{\mathrm{gate}}=[\boldsymbol{\beta}_{\mathrm{gate}}^{(1)},\dots,\boldsymbol{\beta}_{\mathrm{gate}}^{(C)}]\in(0,1)^{H\times C}$, which explicitly aligns with the calibrated dynamic basis $\mathbf{R}_{\mathrm{dyn}}$ for element-wise dynamic grafting.
In this way, the gate releases retrieved dynamics only when the backbone is likely to require correction and the retrieved evidence is sufficiently reliable.

\subsection{Restoration-Cooperative Multi-Objective Optimization}
\label{sec:restoration_optimization}

Although historical retrieval provides candidate dynamics and phase anchoring controls their release, optimizing the whole framework only with point-wise forecasting losses may still pull the grafting branch toward conservative smoothing. 
We therefore introduce a restoration-cooperative objective that explicitly supervises three aspects of controlled dynamic grafting: whether intervention is needed, what local changes should be restored, and how the dominant structure should be preserved:
\begin{equation}
\mathcal{L}
=
\mathcal{L}_{\mathrm{pred}}
+
\lambda_a \mathcal{L}_{\mathrm{gate}}
+
\lambda_b \mathcal{L}_{\mathrm{diff}}
+
\lambda_c \mathcal{L}_{\mathrm{struc}} .
\end{equation}

\textbf{Demand-Aware Gate Supervision ($\mathcal{L}_{\mathrm{gate}}$).} 
Unnecessary grafting on already reliable backbone predictions introduces noise. We train a lightweight MLP to estimate the backbone's intrinsic predictive error, 
$\mathbf{E}^{*} = \frac{1}{H} \sum_{t=1}^{H} |\widehat{\mathbf{Y}}_{\mathrm{base}, t} - \mathbf{Y}_{t}| \in \mathbb{R}^{C}$, which serves as an explicit \emph{rectification demand}. 
The loss is defined as $\mathcal{L}_{\mathrm{gate}} = \| \widehat{\mathbf{E}} - \operatorname{sg}(\mathbf{E}^{*}) \|_{1}$, 
where $\widehat{\mathbf{E}}$ is the MLP estimation. Crucially, the stop-gradient operation ($\operatorname{sg}$) treats the backbone error as a fixed target, preventing the model from trivially minimizing this loss by actively degrading the backbone's performance.

\textbf{Local-Global Dynamic Alignment ($\mathcal{L}_{\mathrm{diff}}, \mathcal{L}_{\mathrm{struc}}$).} To make sparse local dynamics visible to optimization, $\mathcal{L}_{\mathrm{diff}}$ supervises first-order temporal changes: $\mathcal{L}_{\mathrm{diff}} = \frac{1}{H-1} \sum_{t=1}^{H-1} \| (\widehat{\mathbf{Y}}_{t+1} - \widehat{\mathbf{Y}}_{t}) - (\mathbf{Y}_{t+1} - \mathbf{Y}_{t}) \|_{1}$. This explicitly encourages the recovery of slope variations and turning points neglected by $\mathcal{L}_{\mathrm{pred}}$. However, optimizing local changes alone can induce high-frequency fragmentation. To regularize this, $\mathcal{L}_{\mathrm{struc}}$ aligns the Fourier spectral amplitudes: $\mathcal{L}_{\mathrm{struc}} = \| |\mathcal{F}_{t}(\widehat{\mathbf{Y}})| - |\mathcal{F}_{t}(\mathbf{Y})| \|_{1}$,
where $\mathcal{F}_{t}(\cdot)$ denotes the one-dimensional discrete Fourier transform along the temporal dimension.
This ensures that local variations are restored without distorting the dominant periodicity and global trajectory, complementing $\mathcal{L}_{\mathrm{diff}}$ and mirroring the spectral consistency inherent in our historical retrieval design. 
Details are in Appendix \ref{app:optimization_details}.



\section{Experiments}

We evaluate GRAFT on seven standard LTSF benchmarks: ETTh1, ETTh2, ETTm1, ETTm2, Weather, Electricity, and Traffic~\citep{wu2021autoformer,wu2023timesnet,zhou2021informer}.
These datasets span different sampling frequencies, variable dimensions, periodic structures, and local variations.
To assess the model-agnostic applicability of GRAFT, we evaluate it on eight forecasting backbones spanning diverse architectural designs. We include three recent backbones published in 2026: PMDformer~\citep{hu2026pmdformer} (ICLR 2026), a patch-based Transformer; Dynamic-TMoE~\citep{zhu2026dynamic_tmoe} (ICML 2026), a drift-aware dynamic mixture-of-experts model; and XLinear~\citep{chen2026xlinear} (AAAI 2026), a lightweight MLP-based model. We also evaluate GRAFT on five established backbones covering Transformer-based, MLP/linear-based, and CNN-based architectures: PatchTST~\citep{nie2023a}, TimeMixer~\citep{wang2023timemixer}, DLinear~\citep{zeng2023transformers}, TimesNet~\citep{wu2023timesnet}, and PDF~\citep{dai2024periodicity}.
All results are reported over prediction horizons $H\in\{96,192,336,720\}$.
Detailed experimental protocols, implementation details, computational overhead, and leakage-free memory construction are provided in Appendix~\ref{app:experimental_details}.


\subsection{Main Results}
\label{sec:main_results}

\begin{table}[t]
\centering
\caption{
Forecasting performance under long-term settings.
Results are averaged over $H\in\{96,192,336,720\}$.
For each Base/GRAFT pair, the better result is in \textbf{bold}.
Red positive $\Delta\%$ indicates improvement of GRAFT over the corresponding backbone.
$\Delta\%=100(\overline{\mathrm{Base}}-\overline{\mathrm{GRAFT}})/\overline{\mathrm{Base}}$.
Boldface and $\Delta\%$ are determined before rounding.
}
\label{tab:main_results1}
\scriptsize
\renewcommand{\arraystretch}{0.6}
\setlength{\tabcolsep}{1.2pt}
\resizebox{\textwidth}{!}{%
\begin{tabular}{cc|ccc|ccc|ccc|ccc|ccc|ccc|ccc|ccc}
\toprule
\multicolumn{2}{c|}{\multirow{3}{*}{Dataset}} & \multicolumn{9}{c|}{2026 strong baselines} & \multicolumn{3}{c|}{Transformer} & \multicolumn{6}{c|}{MLP} & \multicolumn{6}{c}{CNN} \\
\cmidrule(lr){3-11}\cmidrule(lr){12-14}\cmidrule(lr){15-20}\cmidrule(lr){21-26}
\multicolumn{2}{c|}{} & \multicolumn{3}{c|}{PMDformer} & \multicolumn{3}{c|}{Dynamic-TMoE} & \multicolumn{3}{c|}{XLinear} & \multicolumn{3}{c|}{PatchTST} & \multicolumn{3}{c|}{TimeMixer} & \multicolumn{3}{c|}{DLinear} & \multicolumn{3}{c|}{TimesNet} & \multicolumn{3}{c}{PDF} \\
\cmidrule(lr){3-5}\cmidrule(lr){6-8}\cmidrule(lr){9-11}\cmidrule(lr){12-14}\cmidrule(lr){15-17}\cmidrule(lr){18-20}\cmidrule(lr){21-23}\cmidrule(lr){24-26}
\multicolumn{2}{c|}{} & Base & GRAFT & $\Delta$\% & Base & GRAFT & $\Delta$\% & Base & GRAFT & $\Delta$\% & Base & GRAFT & $\Delta$\% & Base & GRAFT & $\Delta$\% & Base & GRAFT & $\Delta$\% & Base & GRAFT & $\Delta$\% & Base & GRAFT & $\Delta$\% \\
\midrule
\multirow{2}{*}{ETTh1} & MSE & 0.397 & \textbf{0.391} & \textcolor{red}{1.5} & 0.431 & \textbf{0.420} & \textcolor{red}{2.7} & 0.425 & \textbf{0.413} & \textcolor{red}{2.8} & 0.465 & \textbf{0.426} & \textcolor{red}{8.4} & 0.442 & \textbf{0.417} & \textcolor{red}{5.5} & 0.465 & \textbf{0.420} & \textcolor{red}{9.7} & 0.463 & \textbf{0.429} & \textcolor{red}{7.4} & 0.410 & \textbf{0.401} & \textcolor{red}{2.1} \\
 & MAE & 0.420 & \textbf{0.419} & \textcolor{red}{0.3} & 0.432 & \textbf{0.427} & \textcolor{red}{1.1} & 0.428 & \textbf{0.425} & \textcolor{red}{0.8} & 0.452 & \textbf{0.434} & \textcolor{red}{3.9} & 0.434 & \textbf{0.425} & \textcolor{red}{2.0} & 0.461 & \textbf{0.427} & \textcolor{red}{7.4} & 0.455 & \textbf{0.436} & \textcolor{red}{4.2} & 0.425 & \textbf{0.422} & \textcolor{red}{0.9} \\
\midrule
\multirow{2}{*}{ETTh2} & MSE & 0.336 & \textbf{0.331} & \textcolor{red}{1.6} & 0.370 & \textbf{0.362} & \textcolor{red}{2.2} & 0.359 & \textbf{0.353} & \textcolor{red}{1.7} & 0.382 & \textbf{0.361} & \textcolor{red}{5.4} & 0.383 & \textbf{0.375} & \textcolor{red}{2.1} & 0.563 & \textbf{0.350} & \textcolor{red}{37.9} & 0.403 & \textbf{0.376} & \textcolor{red}{6.7} & \textbf{0.335} & 0.337 & -0.4 \\
 & MAE & 0.381 & \textbf{0.381} & \textcolor{red}{0.1} & 0.400 & \textbf{0.392} & \textcolor{red}{2.1} & 0.391 & \textbf{0.386} & \textcolor{red}{1.2} & 0.404 & \textbf{0.389} & \textcolor{red}{3.7} & 0.404 & \textbf{0.398} & \textcolor{red}{1.7} & 0.520 & \textbf{0.382} & \textcolor{red}{26.5} & 0.422 & \textbf{0.400} & \textcolor{red}{5.2} & 0.384 & \textbf{0.383} & \textcolor{red}{0.1} \\
\midrule
\multirow{2}{*}{ETTm1} & MSE & 0.347 & \textbf{0.338} & \textcolor{red}{2.6} & 0.379 & \textbf{0.371} & \textcolor{red}{2.0} & 0.372 & \textbf{0.365} & \textcolor{red}{2.0} & 0.400 & \textbf{0.381} & \textcolor{red}{4.7} & 0.387 & \textbf{0.372} & \textcolor{red}{4.1} & 0.404 & \textbf{0.371} & \textcolor{red}{8.0} & 0.408 & \textbf{0.392} & \textcolor{red}{3.9} & 0.356 & \textbf{0.341} & \textcolor{red}{4.2} \\
 & MAE & 0.372 & \textbf{0.372} & \textcolor{red}{0.2} & 0.394 & \textbf{0.388} & \textcolor{red}{1.7} & 0.390 & \textbf{0.385} & \textcolor{red}{1.5} & 0.410 & \textbf{0.393} & \textcolor{red}{4.1} & 0.402 & \textbf{0.389} & \textcolor{red}{3.1} & 0.408 & \textbf{0.383} & \textcolor{red}{6.0} & 0.414 & \textbf{0.402} & \textcolor{red}{3.0} & 0.380 & \textbf{0.373} & \textcolor{red}{1.8} \\
\midrule
\multirow{2}{*}{ETTm2} & MSE & 0.246 & \textbf{0.244} & \textcolor{red}{0.7} & 0.272 & \textbf{0.272} & \textcolor{red}{0.1} & \textbf{0.270} & 0.270 & -0.1 & \textbf{0.281} & 0.281 & -0.2 & 0.274 & \textbf{0.272} & \textcolor{red}{0.7} & 0.353 & \textbf{0.296} & \textcolor{red}{16.0} & 0.297 & \textbf{0.291} & \textcolor{red}{1.9} & 0.255 & \textbf{0.252} & \textcolor{red}{1.5} \\
 & MAE & \textbf{0.305} & 0.306 & -0.2 & \textbf{0.321} & 0.321 & -0.2 & \textbf{0.316} & 0.317 & -0.2 & 0.328 & \textbf{0.327} & \textcolor{red}{0.3} & 0.323 & \textbf{0.322} & \textcolor{red}{0.3} & 0.400 & \textbf{0.331} & \textcolor{red}{17.4} & 0.333 & \textbf{0.330} & \textcolor{red}{0.9} & 0.314 & \textbf{0.309} & \textcolor{red}{1.8} \\
\midrule
\multirow{2}{*}{Weather} & MSE & 0.217 & \textbf{0.216} & \textcolor{red}{0.6} & 0.245 & \textbf{0.243} & \textcolor{red}{0.8} & 0.237 & \textbf{0.236} & \textcolor{red}{0.5} & 0.259 & \textbf{0.254} & \textcolor{red}{1.6} & 0.245 & \textbf{0.244} & \textcolor{red}{0.1} & \textbf{0.265} & 0.266 & -0.5 & 0.258 & \textbf{0.252} & \textcolor{red}{2.3} & \textbf{0.230} & 0.235 & -2.1 \\
 & MAE & \textbf{0.251} & 0.251 & -0.1 & 0.273 & \textbf{0.270} & \textcolor{red}{0.9} & 0.268 & \textbf{0.266} & \textcolor{red}{0.5} & \textbf{0.281} & 0.283 & -1.0 & 0.275 & \textbf{0.274} & \textcolor{red}{0.3} & 0.315 & \textbf{0.290} & \textcolor{red}{8.2} & 0.285 & \textbf{0.281} & \textcolor{red}{1.5} & \textbf{0.265} & 0.267 & -0.8 \\
\midrule
\multirow{2}{*}{Electricity} & MSE & 0.148 & \textbf{0.146} & \textcolor{red}{1.1} & 0.175 & \textbf{0.173} & \textcolor{red}{1.4} & 0.168 & \textbf{0.166} & \textcolor{red}{1.3} & 0.215 & \textbf{0.201} & \textcolor{red}{6.2} & 0.184 & \textbf{0.182} & \textcolor{red}{1.2} & 0.225 & \textbf{0.187} & \textcolor{red}{16.9} & 0.194 & \textbf{0.184} & \textcolor{red}{5.3} & 0.164 & \textbf{0.162} & \textcolor{red}{1.5} \\
 & MAE & 0.241 & \textbf{0.239} & \textcolor{red}{0.4} & 0.270 & \textbf{0.268} & \textcolor{red}{1.0} & 0.263 & \textbf{0.258} & \textcolor{red}{1.9} & 0.303 & \textbf{0.288} & \textcolor{red}{5.0} & 0.274 & \textbf{0.271} & \textcolor{red}{1.1} & 0.318 & \textbf{0.275} & \textcolor{red}{13.6} & 0.293 & \textbf{0.283} & \textcolor{red}{3.3} & 0.257 & \textbf{0.253} & \textcolor{red}{1.4} \\
\midrule
\multirow{2}{*}{Traffic} & MSE & 0.378 & \textbf{0.374} & \textcolor{red}{1.0} & 0.482 & \textbf{0.476} & \textcolor{red}{1.2} & 0.463 & \textbf{0.451} & \textcolor{red}{2.6} & 0.492 & \textbf{0.487} & \textcolor{red}{1.0} & 0.512 & \textbf{0.502} & \textcolor{red}{2.1} & 0.624 & \textbf{0.536} & \textcolor{red}{14.1} & 0.654 & \textbf{0.628} & \textcolor{red}{4.0} & 0.453 & \textbf{0.444} & \textcolor{red}{1.9} \\
 & MAE & 0.234 & \textbf{0.234} & \textcolor{red}{0.1} & 0.291 & \textbf{0.290} & \textcolor{red}{0.4} & 0.295 & \textbf{0.289} & \textcolor{red}{2.0} & 0.316 & \textbf{0.314} & \textcolor{red}{0.5} & 0.305 & \textbf{0.298} & \textcolor{red}{2.1} & 0.384 & \textbf{0.328} & \textcolor{red}{14.6} & 0.342 & \textbf{0.329} & \textcolor{red}{3.7} & 0.285 & \textbf{0.277} & \textcolor{red}{2.7} \\
\bottomrule
\end{tabular}%
}
\vspace{-1.0em}
\end{table}

Table~\ref{tab:main_results1} summarizes performance averaged over four prediction horizons
$H\in\{96,192,336,720\}$ across eight representative backbones, with each setting corresponding to one dataset--backbone pair.
Across the 56 settings, GRAFT improves at least one metric in 54/56 (96.4\%) settings and improves both MSE and MAE in 47/56 (83.9\%), with average relative reductions of 3.85\% and 2.97\%, respectively.
The complete horizon-wise evaluation further shows lower MSE in 200/224 (89.3\%) and lower MAE in 184/224 (82.1\%) comparisons
(Appendix~\ref{app:full_results}).

The results further support the backbone-agnostic effectiveness of GRAFT.
Across the three recent 2026 strong backbones, PMDformer, Dynamic-TMoE, and XLinear, GRAFT reduces MSE in 20/21 (95.2\%) and MAE in 17/21 (81.0\%) dataset--backbone settings.
Consistent improvements are also observed across Transformer-, MLP-, and CNN-based forecasters, indicating that recoverable local dynamics remain complementary to a broad range of backbone representations.
Additional analyses excluding DLinear and reporting the complete backbone- and horizon-wise results are provided in Appendix~\ref{app:full_results}.

From a dataset perspective, GRAFT improves both metrics for all eight backbones on ETTh1, ETTm1, Electricity, and Traffic.
Results are more mixed on ETTm2 and Weather.
Appendix~\ref{app:main_result_interpretation} examines these differences in relation to dataset structural profiles and the historical transferability of local dynamics.

\begin{table}[h]
\centering
\vspace{-0.5em} 
\caption{
Head-to-head comparison with plug-and-play forecasting enhancement methods
under a shared PatchTST backbone.
Results are obtained from independent training runs.
}
\label{tab:main_results2}

\scriptsize
\renewcommand{\arraystretch}{0.72}
\setlength{\tabcolsep}{3.5pt}
\setlength{\aboverulesep}{1pt}
\setlength{\belowrulesep}{1pt}

\resizebox{0.88\textwidth}{!}{
\begin{tabular}{l|c|c|c|c}
\toprule
\textbf{MSE/MAE}
& \textbf{PatchTST}
& \textbf{+GRAFT (Ours)}
& \textbf{+PIR (2025)}
& \textbf{+TimeAlign (2026)} \\
\midrule

\textbf{ETTh1}       
& 0.465 / 0.452
& \textbf{0.426} / \textbf{0.434}
& 0.437 / 0.439
& 0.453 / 0.441 \\

\textbf{ETTh2}
& 0.382 / 0.404
& \textbf{0.361} / \textbf{0.389}
& 0.375 / 0.400
& 0.382 / 0.397 \\

\textbf{ETTm1}
& 0.400 / 0.410
& \textbf{0.381} / 0.393
& 0.383 / 0.397
& 0.392 / \textbf{0.392} \\

\textbf{ETTm2}
& \textbf{0.280} / 0.328
& 0.281 / \textbf{0.327}
& 0.282 / 0.330
& 0.295 / 0.336 \\

\textbf{Weather}
& 0.259 / 0.281
& \textbf{0.254} / 0.283
& \textbf{0.254} / \textbf{0.277}
& 0.263 / 0.278 \\

\textbf{Electricity}
& 0.214 / 0.303
& \textbf{0.201} / \textbf{0.288}
& 0.202 / 0.289
& 0.205 / 0.289 \\

\textbf{Traffic}
& 0.491 / 0.316
& \textbf{0.486} / \textbf{0.314}
& 0.487 / \textbf{0.314}
& 0.507 / 0.321 \\
\bottomrule
\end{tabular}
}

\vspace{-0.8em}
\end{table}

\subsection{Comparison with Enhancements under a Fixed-Backbone Protocol}


We compare GRAFT with a directly relevant retrieval-based revision via PIR~\citep{liu2025improving} and alignment-based enhancement via TimeAlign~\citep{hu2026bridging}, using PatchTST~\citep{nie2023a} as the shared backbone.
All methods use the same data splits, input lengths, prediction horizons, and forecasting protocol.
GRAFT achieves the best or tied-best performance in 11 of 14 dataset--metric comparisons (Table~\ref{tab:main_results2}), supporting its effectiveness under a controlled backbone setting.
Protocol details are provided in Appendix~\ref{app:comparison_baselines}.


\vspace{-0.3em}
\begin{table}[h]
    \centering
    \caption{
    Ablation of GRAFT on TimeMixer, 
    \textbf{Naive Prototype}: vanilla cosine retrieval without phase- and channel-aware anchoring;
    \textbf{Residual MLP}: MLP-based residual adapter;
    \textbf{Residualized Memory}: backbone residuals as memory values.
    Details are provided in Appendix~\ref{app:ablation_discussion}.
    Results are obtained from independent training runs.
    }
    \label{tab:ablation}
    \scriptsize
    \renewcommand{\arraystretch}{0.5}
    \resizebox{\textwidth}{!}{
    \begin{tabular}{l |cccccccccc}
        \toprule
        \textbf{Datasets} & \multicolumn{2}{c}{\textbf{ETTh1}} & \multicolumn{2}{c}{\textbf{ETTh2}} & \multicolumn{2}{c}{\textbf{ETTm1}} & \multicolumn{2}{c}{\textbf{Electricity}} & \multicolumn{2}{c}{\textbf{Traffic}} \\
        \cmidrule(lr){2-3} \cmidrule(lr){4-5} \cmidrule(lr){6-7} \cmidrule(lr){8-9} \cmidrule(lr){10-11}
        \textbf{Method} & \textbf{MSE} & \textbf{MAE} & \textbf{MSE} & \textbf{MAE} & \textbf{MSE} & \textbf{MAE} & \textbf{MSE} & \textbf{MAE} & \textbf{MSE} & \textbf{MAE} \\
        \midrule
        \textbf{GRAFT} & \textbf{0.417} & \textbf{0.425} & \textbf{0.375} & \textbf{0.397} & \textbf{0.372} & \textbf{0.389} & \textbf{0.182} & \textbf{0.271} & \textbf{0.502} & \textbf{0.298} \\
        \midrule
        \textbf{w/ Naive Prototype} & 0.428 & 0.430 & 0.379 & 0.399 & 0.381 & 0.396 & 0.183 & 0.272 & 0.506 & 0.300 \\
        \textbf{w/o Confidence-Aware Gate} & 0.434 & 0.433 & 0.381 & 0.400 & 0.383 & 0.401 & 0.185 & 0.274 & 0.512 & 0.306 \\
        \textbf{w/o Multi-Objective} & 0.423 & 0.429 & 0.378 & 0.401 & 0.379 & 0.395 & 0.184 & 0.273 & 0.509 & 0.303 \\
        \midrule
        \textbf{w/ Residual MLP} & 0.472 
        & 0.467 & 0.387 & 0.408 & 0.386 & 0.401 & 0.186 & 0.277 & 0.514 & 0.309
        \\
        \textbf{w/ Residualized Memory}
        & 0.423 & 0.428
        & 0.376 & \textbf{0.397}
        & 0.380 & 0.396
        & 0.185 & 0.272
        & 0.505 & 0.300 \\
        \midrule
        \textbf{TimeMixer Backbone} & 0.442 & 0.434  & 0.383 & 0.404 & 0.387 & 0.402 & 0.184  & 0.274  & 0.512 & 0.305 \\
        \bottomrule
    \end{tabular}
    }
    \vspace{-1em}
\end{table}

\subsection{Ablation Studies}
\label{sec:ablation_studies}

We ablate GRAFT on TimeMixer to assess three key components: historical prototype construction, confidence-aware release, and restoration-cooperative optimization.
We use TimeMixer as the ablation backbone because it is a recent strong architecture and shows stable gains in Table~\ref{tab:main_results1}, making it suitable for diagnosing component-level contributions.
We report the main ablation on datasets with relatively stable response transferability, where component effects can be more clearly attributed, as discussed in Sec.~\ref{sec:main_results}. 
Table~\ref{tab:ablation} shows that the full model generally performs best, and weakening any component leads to performance degradation.
Notably, replacing complete historical responses with backbone residuals is less consistent, supporting our choice to preserve predictor-independent evidence and determine transferable dynamics only at grafting time.
Moreover, the residual MLP often underperforms the original backbone, showing that the gains do not arise from an auxiliary correction path itself.
Together, these results support GRAFT's central division of labor: retrieve historical evidence, selectively graft compatible dynamics, and leave structural prediction to the backbone.
Details are in Appendix~\ref{app:ablation_discussion}.

\subsection{Dynamic Recovery Analysis}

 \begin{figure}[!t]
 \centering

 \begin{subfigure}[c]{0.49\textwidth}
     \centering
     \raisebox{0.15cm}{\includegraphics[width=\textwidth]{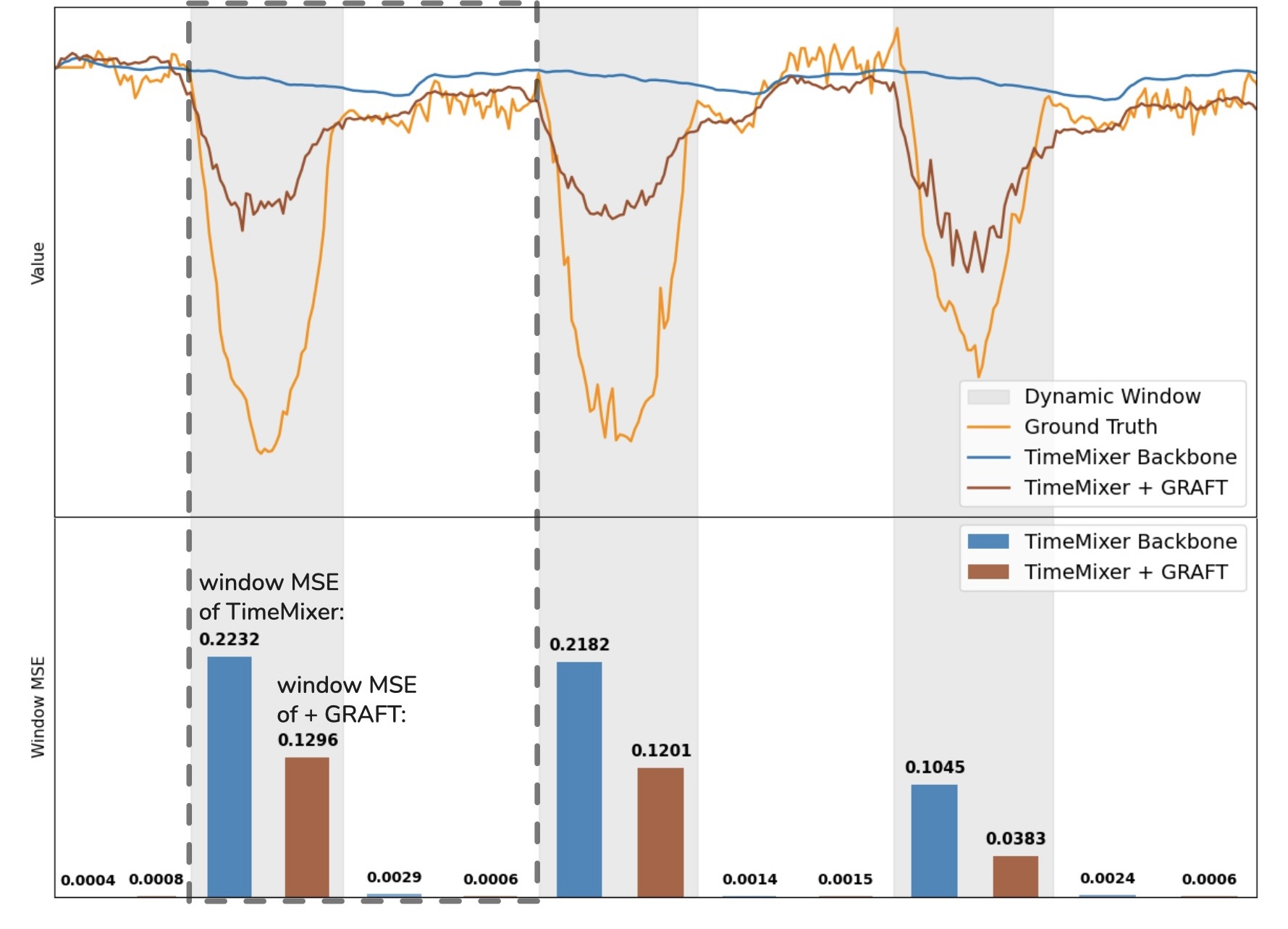}}
     \caption{Dynamic-window analysis on ETTm1}
     \label{fig:dynamic_window}
 \end{subfigure}
 \hfill
 \begin{minipage}[c]{0.49\textwidth}
     \centering

     \begin{subfigure}[t]{0.48\textwidth}
         \centering
         \includegraphics[width=\textwidth]{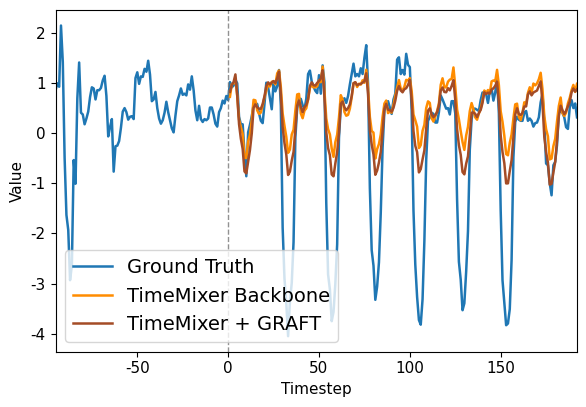}
         \caption{ETTh1}
         \label{fig:right_top_left}
     \end{subfigure}
     \hfill
     \begin{subfigure}[t]{0.48\textwidth}
         \centering        
         \includegraphics[width=\textwidth]{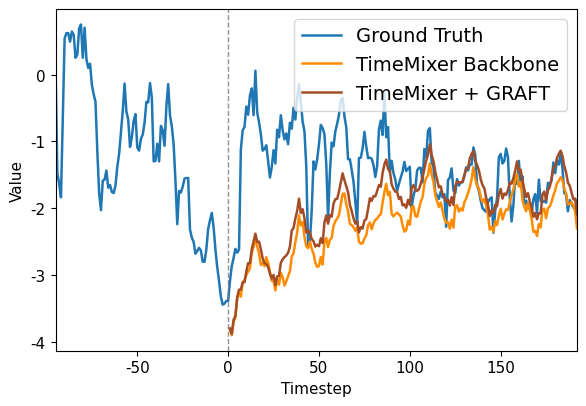}
         \caption{ETTh2}
         \label{fig:right_top_right}
     \end{subfigure}

     \begin{subfigure}[t]{0.48\textwidth}
         \centering
         \includegraphics[width=\textwidth]{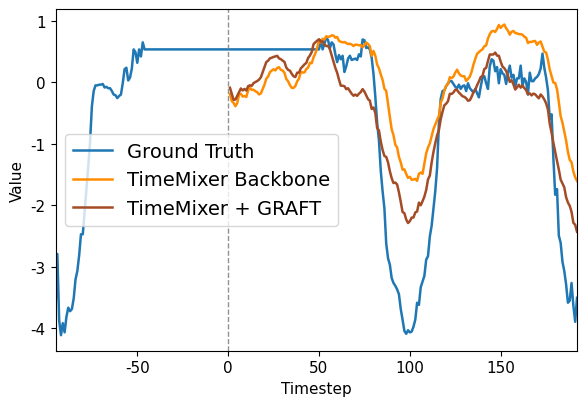}
         \caption{ETTm1}
         \label{fig:right_bottom_left}
     \end{subfigure}
     \hfill
     \begin{subfigure}[t]{0.48\textwidth}
         \centering
         \includegraphics[width=\textwidth]{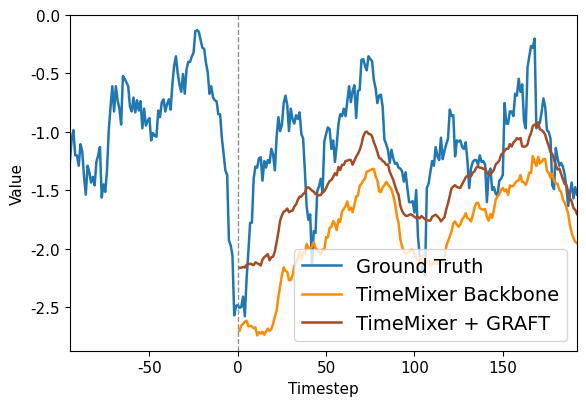}
         \caption{ETTm2}
         \label{fig:right_bottom_right}
     \end{subfigure}
 \end{minipage}
 \vspace{-0.5em}
 \caption{Dynamic recovery analysis. (a) Window-level analysis on ETTm1,
 (b)--(e) Additional cases across datasets.}
 \label{fig:case_visualization}
 \vspace{-1.5em}
 \end{figure}

\begin{table}[h]
\centering
\caption{
Dynamics-stratified error reduction after applying GRAFT.
}

\label{tab:dynamic_recovery_model_dataset}
\scriptsize
\renewcommand{\arraystretch}{0.6}
\setlength{\tabcolsep}{3.0pt}
\resizebox{\linewidth}{!}{
\begin{tabular}{ll|ccccc|ccccc}
\toprule
\multirow{2}{*}{\textbf{Backbone}} 
& \multirow{2}{*}{\textbf{Dataset}}
& \multicolumn{5}{c|}{Group-wise $\Delta_{\mathrm{MSE}}$}
& \multicolumn{5}{c}{Relative Contribution Share} \\
\cmidrule(lr){3-7} \cmidrule(lr){8-12}
& & Q1 & Q2 & Q3 & Q4 & Q5
& Q1 & Q2 & Q3 & Q4 & Q5 \\

\midrule
\multirow{5}{*}{TimesNet}
& ETTh1
& 0.0106 & 0.0042 & 0.0136 & 0.0168 & \textbf{0.0760}
& 8.8\% & 3.4\% & 11.2\% & 13.8\% & \textbf{62.7\%} \\
& ETTh2
& 0.0021 & 0.0167 & 0.0192 & 0.0258 & \textbf{0.0362}
& 2.1\% & 16.7\% & 19.2\% & 25.8\% & \textbf{36.2\%} \\
& ETTm1
& 0.0030 & 0.0012 & 0.0006 & 0.0015 & \textbf{0.0040}
& 28.9\% & 11.9\% & 6.1\% & 14.3\% & \textbf{38.8\%} \\
& ETTm2
& 0.0018 & 0.0097 & 0.0108 & 0.0096 & \textbf{0.0142}
& 4.0\% & 21.1\% & 23.4\% & 20.8\% & \textbf{30.7\%} \\
& Weather
& 0.0086 & 0.0048 & 0.0057 & 0.0070 & \textbf{0.0100}
& 23.8\% & 13.2\% & 15.7\% & 19.4\% & \textbf{27.8\%} \\

\midrule
\multirow{5}{*}{TimeMixer}
& ETTh1
& 0.0133 & 0.0239 & 0.0193 & 0.0336 & \textbf{0.0627}
& 8.7\% & 15.6\% & 12.6\% & 22.0\% & \textbf{41.0\%} \\
& ETTh2
& 0.0020 & 0.0093 & 0.0157 & \textbf{0.0203} & \textbf{0.0203}
& 3.0\% & 13.7\% & 23.2\% & \textbf{30.1\%} & \textbf{30.1\%} \\
& ETTm1
& 0.0002 & 0.0100 & \textbf{0.0134} & 0.0121 & 0.0121
& 0.4\% & 20.9\% & \textbf{28.0\%} & 25.3\% & 25.4\% \\
& ETTm2
& 0.0029 & 0.0094 & 0.0107 & 0.0102 & \textbf{0.0118}
& 6.4\% & 20.9\% & 23.8\% & 22.6\% & \textbf{26.4\%} \\
& Weather
& 0.0007 & -0.0013 & \textbf{0.0043} & 0.0010 & 0.0012 
 & 11.6\% & -23.2\% & \textbf{74.4\%} & 16.6\% & 20.5\%  \\
\bottomrule
\end{tabular}
}
\end{table}

Aggregate MSE/MAE cannot reveal whether a model recovers high-utility local dynamics or merely improves average point-wise accuracy. 
We therefore examine where GRAFT reduces error and whether these reductions occur in the dynamic regions that motivate our method.
Fig.~\ref{fig:case_visualization}(a) shows that GRAFT recovers sharp drops and rebounds smoothed by TimeMixer on ETTm1.
Window-level error reductions concentrate in the shaded dynamic regions, with small changes in smooth regions.
Additional cases in Fig.~\ref{fig:case_visualization}(b)--(e) illustrate similar recovery across datasets.

For quantitative analysis, we partition forecast points into five quantile groups by the absolute second-order difference of the ground truth, from smooth Q1 to highly dynamic Q5.
Table~\ref{tab:dynamic_recovery_model_dataset} reports group-wise MSE reductions,
$\Delta_{\mathrm{MSE}}=\mathrm{MSE}_{\mathrm{Backbone}}-\mathrm{MSE}_{\mathrm{Backbone+GRAFT}}$,
and their relative contribution shares.
Q5 contributes the largest error reduction on all five datasets for TimesNet; for TimeMixer, the largest reductions occur in Q3--Q5.
Together, these results support the concentration of GRAFT's gains in dynamic regions.
Further details are provided in Appendix~\ref{app:dynamic_recovery_details}.

\section{Conclusion}
\vspace{-0.7em}
We introduced GRAFT, a plug-and-play forecasting enhancement framework built on \emph{Predict Backbone, Graft Dynamics}.
Its methodological contribution is to treat dynamic recovery as a distinct forecasting responsibility: preserving the complete backbone forecast as a structural anchor while selectively restoring historically grounded local dynamics through grounded retrieval, phase- and channel-aware calibration, confidence-gated release, and restoration-cooperative optimization.
Experiments across standard benchmarks, forecasting horizons, and backbone families show that GRAFT improves aggregate accuracy while recovering peaks, valleys, fluctuations, and turning regions that are often suppressed by point-wise optimization. 
These findings highlight an underemphasized opportunity in forecasting: explicitly modeling the recovery of sparse dynamics alongside reliable structural prediction.

\subsection*{AI use statement}

We used generative AI tools solely to assist with translation
into English and language polishing of author-written text.
These tools were not used to develop or substantiate the
core ideas or mechanisms, formulate hypotheses, design
experiments, implement methods, process data, interpret
results, or construct theoretical arguments or proofs.
They were also not used to search for or suggest references.
All references were checked by the authors.

The authors manually reviewed and revised all AI-assisted
passages to verify that they accurately conveyed the intended
meaning. The authors take full responsibility for the final
content of this paper.

\bibliographystyle{iclr2027_conference}
\bibliography{references.bib}

\appendix

\section{Additional Method Details}
\label{app:method_details}

\subsection{MSE Rewards and the Risk of Local-Event Recovery}
\label{app:mse_reward_risk}

We formalize two aspects of time-series forecasting relevant to our motivation:
the larger loss reduction available from high-energy structural residuals,
and the limited recoverable reward and misalignment cost of uncertain
local events.

\paragraph{Analytical setup.}
Consider a target $\mathbf{y}=\mathbf{b}+\mathbf{d}\in\mathbb{R}^{H}$,
where the integer $H\geq 1$ is the prediction horizon, $\mathbf{b}$ denotes the
macroscopic backbone, and $\mathbf{d}$ denotes local dynamics.
For a prediction $\widehat{\mathbf{y}}_{\theta}=f_{\theta}(\mathbf{X})$,
write $\mathbf{r}_{\theta}=\mathbf{y}-\widehat{\mathbf{y}}_{\theta}$.
For the first result, use fixed orthogonal projectors
$P_b,P_d\in\mathbb{R}^{H\times H}$ with
$P_bP_d=0$ and $P_b+P_d=I_H$ to define the analytical residual components
$\mathbf{r}_b=P_b\mathbf{r}_{\theta}$ and
$\mathbf{r}_d=P_d\mathbf{r}_{\theta}$.
Here $P_b$ selects a prescribed structural subspace and $P_d$ its
orthogonal complement; this does not require separate network output heads
or an exact separation of smooth signals and sparse events.
Orthogonality yields
\begin{equation}
    \mathcal{L}(\theta)
    =\frac{1}{H}\|\mathbf{r}_{\theta}\|_2^2
    =\underbrace{\frac{1}{H}\|\mathbf{r}_b\|_2^2}_{\mathcal{L}_b(\theta)}
    +\underbrace{\frac{1}{H}\|\mathbf{r}_d\|_2^2}_{\mathcal{L}_d(\theta)}.
    \label{eq:app_mse_components}
\end{equation}
For $C$ channels, the same argument uses the Frobenius norm and the
normalization $1/(HC)$, with projections applied along time.

\begin{lemma}[Energy-weighted MSE reward]
\label{lem:app_backbone_reward}
Suppose $\|\mathbf{r}_b\|_2\geq\Gamma\|\mathbf{r}_d\|_2$ for
some $\Gamma>1$. Consider an update $\theta\mapsto\theta^{+}$ such that
\[
    \|\mathbf{r}_b^{+}\|_2^2=(1-\alpha)\|\mathbf{r}_b\|_2^2,
    \qquad
    \|\mathbf{r}_d^{+}\|_2^2=(1-\beta)\|\mathbf{r}_d\|_2^2,
    \qquad \alpha,\beta\in[0,1],
\]
where $\mathbf{r}_u^{+}=P_u\mathbf{r}_{\theta^{+}}$ for $u\in\{b,d\}$.
Define $\Delta\mathcal{L}_u=\mathcal{L}_u(\theta)-\mathcal{L}_u(\theta^{+})$.
Then
\begin{equation}
    \Delta\mathcal{L}_b=\frac{\alpha}{H}\|\mathbf{r}_b\|_2^2,
    \qquad
    \Delta\mathcal{L}_d=\frac{\beta}{H}\|\mathbf{r}_d\|_2^2.
    \label{eq:app_mse_reward}
\end{equation}
If $\beta>0$ and $\|\mathbf{r}_d\|_2>0$, then
\begin{equation}
    \frac{\Delta\mathcal{L}_b}{\Delta\mathcal{L}_d}
    =\frac{\alpha}{\beta}
      \frac{\|\mathbf{r}_b\|_2^2}{\|\mathbf{r}_d\|_2^2}
    \geq\frac{\alpha\Gamma^2}{\beta}.
    \label{eq:app_mse_reward_ratio}
\end{equation}
In particular, equal positive relative reductions in squared residual
energy yield at least $\Gamma^2$ times more MSE reduction from the
structural component.
\end{lemma}

\begin{proof}
Subtract the post-update component losses in
\eqref{eq:app_mse_components} from their pre-update values to obtain
\eqref{eq:app_mse_reward}. Dividing these expressions and applying
the assumed energy inequality gives~\eqref{eq:app_mse_reward_ratio}.
If $\beta=0$ or $\mathbf{r}_d=\mathbf{0}$, the dynamic contribution
is zero and no ratio is needed.
\end{proof}

This is a conditional loss-reduction statement.
Parseval's identity preserves these energies under a unitary Fourier
transform, so the same accounting holds in either domain for the same
components; it does not assign an intrinsic preference to low frequencies.

\paragraph{Uncertain local events.}
We next isolate event-prediction error, equivalently assuming that the
backbone has been subtracted exactly. This is a separate event model;
the event need not lie in the range of $P_d$.
Fix a context $\mathbf{X}$, an integer $1\leq M\leq H$, distinct candidate positions
$\tau_1,\ldots,\tau_M\in\{1,\ldots,H\}$, and an amplitude $a>0$.
Let
\begin{equation}
    \mathbf{d}=za\mathbf{e}_{\tau},\qquad
    \Pr(z=1\mid\mathbf{X})=p,\qquad
    \Pr(\tau=\tau_j\mid z=1,\mathbf{X})=q_j,
    \label{eq:app_local_event}
\end{equation}
where $\mathbf{e}_t$ is the $t$th standard basis vector,
$p\in(0,1]$, $q_j\geq0$, and $\sum_{j=1}^{M}q_j=1$.
Define the conditional event risk of a deterministic prediction
$\mathbf{v}\in\mathbb{R}^{H}$ by
$\mathcal{R}_{\mathbf{X}}(\mathbf{v})
=H^{-1}\mathbb{E}[\|\mathbf{d}-\mathbf{v}\|_2^2\mid\mathbf{X}]$.

\begin{lemma}[Recoverable event reward and misalignment cost]
\label{lem:app_event_risk}
Under~\eqref{eq:app_local_event}, the unique minimizer of
$\mathcal{R}_{\mathbf{X}}$ over $\mathbb{R}^{H}$ is
$\mathbf{v}^{\star}=pa\sum_{j=1}^{M}q_j\mathbf{e}_{\tau_j}$.
Relative to omission, its absolute and relative MSE reductions are
\begin{align}
    \mathcal{R}_{\mathbf{X}}(\mathbf{0})
    -\mathcal{R}_{\mathbf{X}}(\mathbf{v}^{\star})
    &=\frac{p^2a^2}{H}\sum_{j=1}^{M}q_j^2,
    \label{eq:app_event_absolute_reward}\\
    \frac{\mathcal{R}_{\mathbf{X}}(\mathbf{0})
    -\mathcal{R}_{\mathbf{X}}(\mathbf{v}^{\star})}
    {\mathcal{R}_{\mathbf{X}}(\mathbf{0})}
    &=p\sum_{j=1}^{M}q_j^2.
    \label{eq:app_event_relative_reward}
\end{align}
For uniform phase probabilities $q_j=1/M$, the relative reduction is $p/M$.
Furthermore, for a realized event $\mathbf{d}=a\mathbf{e}_{\tau}$,
predicting the correct amplitude at a different valid position $t\neq\tau$
incurs
\begin{equation}
    \mathcal{L}_{\mathrm{shift}}
    =\frac{1}{H}\|a\mathbf{e}_{\tau}-a\mathbf{e}_t\|_2^2
    =\frac{2a^2}{H}
    =2\mathcal{L}_{\mathrm{omit}},
    \qquad
    \mathcal{L}_{\mathrm{omit}}=\frac{a^2}{H}.
    \label{eq:app_event_shift}
\end{equation}
\end{lemma}

\begin{proof}
The finite event model gives
$\mathbb{E}[\mathbf{d}\mid\mathbf{X}]
=pa\sum_jq_j\mathbf{e}_{\tau_j}=\mathbf{v}^{\star}$ and
$\mathbb{E}[\|\mathbf{d}\|_2^2\mid\mathbf{X}]=pa^2$.
Expanding the squared error and completing the square yields
\[
    \mathcal{R}_{\mathbf{X}}(\mathbf{v})
    =\frac{1}{H}\left(
       pa^2-\|\mathbf{v}^{\star}\|_2^2
       +\|\mathbf{v}-\mathbf{v}^{\star}\|_2^2\right).
\]
Thus $\mathbf{v}^{\star}$ is the unique minimizer.
Since the candidate basis vectors are orthonormal,
$\|\mathbf{v}^{\star}\|_2^2=p^2a^2\sum_jq_j^2$.
Subtracting the minimum risk from
$\mathcal{R}_{\mathbf{X}}(\mathbf{0})=pa^2/H$ proves
\eqref{eq:app_event_absolute_reward}; division by this positive
omission risk proves~\eqref{eq:app_event_relative_reward}.
Finally, $\langle\mathbf{e}_{\tau},\mathbf{e}_t\rangle=0$ for
$t\neq\tau$, which gives~\eqref{eq:app_event_shift}.
\end{proof}

If $p=0$, the event is absent almost surely: the optimum and absolute
gain are zero, while the relative gain is undefined.
The factor of two in~\eqref{eq:app_event_shift} is specific to
nonoverlapping impulses, rather than arbitrary shifted waveforms.

\paragraph{Implication for dynamic recovery.}
Lemma~\ref{lem:app_backbone_reward} identifies the larger MSE payoff
of structural fitting when structural residual energy dominates and
relative progress is comparable.
Lemma~\ref{lem:app_event_risk} shows that low occurrence probability
and diffuse phase probabilities reduce the fraction of event error
recoverable by a deterministic predictor, while a misplaced sharp
response can cost more than omission.
Together, these results motivate treating dynamic recovery separately
and conditioning sharp corrections on reliable event and phase evidence,
as in GRAFT. They characterize MSE rewards and event-level risk.

\subsection{Complete Responses vs. Residualized Retrieval Targets}
\label{app:full_response_memory}

This section further examines the memory-target design in Sec.~\ref{sec:historical_recurrence}.
Our historical memory is defined as
\begin{equation}
\mathcal{M}
=
\{(\mathbf{K}_i,\mathbf{V}_i)\}_{i=1}^{N},
\end{equation}
where $\mathbf{K}_i\in\mathbb{R}^{L\times C}$ is a historical look-back window and
$\mathbf{V}_i\in\mathbb{R}^{H\times C}$ is its associated future response.
GRAFT stores the complete observed response,
\begin{equation}
\mathbf{V}_i^{\mathrm{full}}
=
\mathbf{Y}_i.
\end{equation}

A natural alternative is to residualize the memory value with respect to the backbone:
\begin{equation}
\mathbf{V}_i^{\mathrm{res}}
=
\mathbf{Y}_i
-
f_{\theta}(\mathbf{K}_i),
\end{equation}
where $f_{\theta}(\mathbf{K}_i)$ denotes the backbone forecast for the historical context.
Although this representation appears directly aligned with correction learning, it changes the semantics of the memory.
The complete response records \emph{what actually followed} a historical state, whereas the residual records \emph{what a particular backbone failed to predict}.
Consequently, residualized memory is inherently coupled to the architecture, parameters, and approximation errors of the forecaster being enhanced.

This distinction is important because a forecasting residual is not necessarily equivalent to a transferable local dynamic.
It may contain genuine transient variations, but also structural estimation errors, systematic backbone bias, and noise.
Residualization therefore performs an early decomposition in which the historical dynamic target is defined by a particular backbone before the current query is observed.
In contrast, GRAFT preserves predictor-independent historical evidence and postpones the identification of transferable dynamics until grafting time.

Preserving the complete response also retains the context in which a local event occurs.
The meaning of a spike, drop, rebound, or turning point depends on its surrounding level, trend direction, amplitude scale, and temporal phase.
Two locally similar variations may imply different dynamics when embedded in different structural trajectories.
Rather than discarding this context through backbone-dependent residualization, GRAFT retrieves the complete historical response and determines its transferable contribution conditioned on the current forecasting state.

Specifically, the retrieved prototype is not used as an alternative forecast or directly copied to the prediction.
It is transformed into a controlled response contribution through phase modulation, confidence gating, and channel-wise dynamic magnitude control:
\begin{equation}
\widehat{\mathbf{Y}}
=
\widehat{\mathbf{Y}}_{\mathrm{base}}
+
\underbrace{
\left(
\boldsymbol{\beta}_{\mathrm{gate}}
\odot
\boldsymbol{\Lambda}_{\mathrm{anchor}}
\right)
}_{\text{learned dynamic transfer field}}
\odot
\mathbf{R}.
\end{equation}
Thus, memory construction and dynamic transfer play different roles:
the memory preserves historically observed response evidence, while the grafting mechanism determines which part of that evidence is compatible with the current context and should complement the structural backbone.

We directly evaluate this design choice in Table~\ref{tab:ablation}.
The \textbf{w/ Residualized Memory} variant keeps the TimeMixer backbone, historical keys, retrieval rule, grafting mechanism, and optimization objectives unchanged, and replaces only the complete-response memory value with the backbone residual.
Complete-response memory yields better results on ETTh1 and ETTm1, while residualized memory remains competitive on ETTh2.
These results suggest that residualized targets can be useful when backbone errors themselves exhibit reusable structure, but their utility is tied to the particular forecaster that produces them.
Complete responses provide a more generally reliable and predictor-independent evidence space for dynamic transfer.

This experiment also complements the \textbf{w/ Residual MLP} control in Sec.~\ref{sec:ablation_studies}.
The former tests \emph{what historical evidence should be stored}, whereas the latter tests \emph{how retrieved evidence should be transferred}.
Together, they support the central design principle of GRAFT:
historical memory should preserve predictor-independent evidence of what occurred, while the grafting path should determine, at prediction time, which compatible dynamics should be released onto the backbone forecast.

\subsection{Clarification on Data Leakage and Shortcut-Free Retrieval}
\label{app:leakage_free_retrieval}

In retrieval-augmented time series forecasting, a critical concern is the risk of future data leakage.
We first clarify that our memory bank is built \emph{exclusively from the training split}, which eliminates any conventional future leakage in the standard evaluation protocol.

However, we identify and address a more subtle optimization pathology induced by the sliding-window construction: \emph{degenerate shortcut learning}.
Long-term forecasting datasets are typically generated via a sliding-window scheme, leading to heavy temporal overlap between adjacent samples.
If the current query $\mathbf{X}$ is permitted to search the entire memory bank without constraint, the retrieval module can trivially match the query to its exact self or a nearly identical shifted window.
In this case, the retrieved value $\mathbf{V}$ would nearly perfectly align with the current target $\mathbf{Y}$.
The model can then minimize the loss by exploiting a trivial lookup shortcut and completely bypass meaningful feature extraction and dynamic grafting.

To prevent this degenerate behavior and enforce meaningful matching, we explicitly mask the temporal neighborhood of the query during training.
Concretely, for a query window starting at index $t$ with input length $L$, we exclude any memory entry $(\mathbf{K}_i, \mathbf{V}_i)$ whose index $i$ lies in the range:
\begin{equation}
i \in [t - L, \, t + L].
\end{equation}
By setting the similarity scores in this interval to $-\infty$, we ensure that retrieved prototypes are drawn from \emph{truly distinct historical contexts} rather than trivially overlapping fragments.
\subsection{Channel-Adaptive Similarity Fusion and Prototype Construction}
\label{app:prototype_construction}

\begin{figure}[t]
    \centering
    \includegraphics[width=0.7\linewidth]{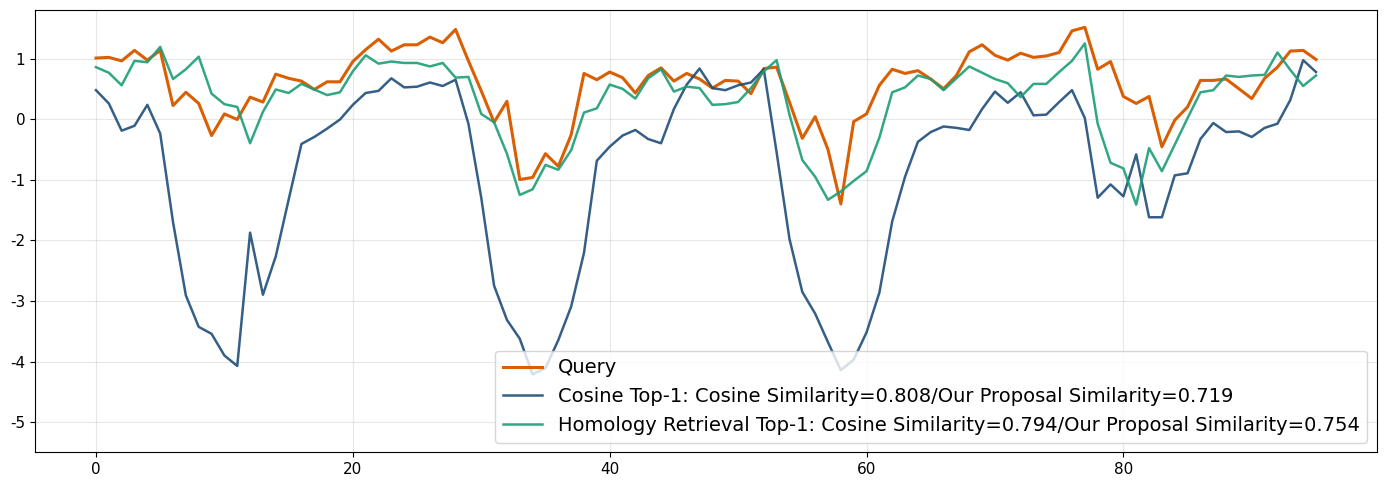}
    \caption{Illustration of pseudo-similar trajectories. Raw-value-only similarity may retrieve candidates with inconsistent temporal shapes, mutation patterns, or evolution trends.}
    \label{fig:pseudo_similarity}
\end{figure}

This section provides additional details for the homology retrieval mechanism introduced in Sec.~\ref{sec:historical_recurrence}.
The main text motivates retrieval from the issue of \emph{pseudo-similar} trajectories: historical candidates that appear close to the query under a single raw-value similarity measure but exhibit inconsistent temporal shapes, mutation patterns, or periodic structures.
Such candidates are unreliable for dynamic grafting because their future responses may contain local dynamics that are not transferable to the current query.
To reduce pseudo-similar retrieval, as illustrated in Figure~\ref{fig:pseudo_similarity}, we measure historical homology from three complementary views: time-domain shape, local differential evolution, and phase-invariant spectral composition.

For a query channel $\mathbf{q}^{(c)}\in\mathbb{R}^{L}$ and a historical key channel $\mathbf{k}_i^{(c)}\in\mathbb{R}^{L}$, our retrieval score combines the following three views.

\paragraph{Time-domain shape similarity.}
\begin{equation}
\mathrm{Sim}_{\mathrm{shape}}^{(i,c)}
=
\mathrm{CosSim}
\left(
\mathbf{q}^{(c)},
\mathbf{k}_i^{(c)}
\right).
\end{equation}
This metric measures macroscopic contour consistency between the query and a historical key in the original signal space.
It helps identify candidates that follow similar overall rising, falling, or oscillatory patterns.
This view is important because two segments may be close at several point-wise positions but still follow different global evolution trends.
By enforcing contour-level consistency, the shape view filters candidates whose future responses may be structurally misaligned with the current query.

\paragraph{Local differential similarity.}
\begin{equation}
\mathrm{Sim}_{\mathrm{diff}}^{(i,c)}
=
\mathrm{CosSim}
\left(
\Delta\mathbf{q}^{(c)},
\Delta\mathbf{k}_i^{(c)}
\right),
\end{equation}
where $\Delta$ denotes the discrete difference operator along the temporal dimension.
This metric focuses on local changes rather than absolute values.
It helps identify historical contexts with similar slope transitions, mutation tendencies, and short-term dynamic evolution.
This view is particularly important for dynamic grafting because transferable information often lies not in the absolute level of a segment, but in how the segment changes locally.
For example, two contexts may share a similar level or coarse shape, yet one may be approaching a turning point while the other remains smooth.
The differential view helps distinguish such cases by comparing local evolution patterns.

\paragraph{Phase-invariant spectral similarity.}
\begin{equation}
\mathrm{Sim}_{\mathrm{freq}}^{(i,c)}
=
\mathrm{CosSim}
\left(
|\mathcal{F}(\mathbf{q}^{(c)})|,
|\mathcal{F}(\mathbf{k}_i^{(c)})|
\right),
\end{equation}
where $\mathcal{F}(\cdot)$ denotes the 1D Discrete Fourier Transform.
By comparing amplitude spectra, this metric captures the periodic composition of the sequences while reducing sensitivity to minor temporal shifts.
It focuses on which periodic components are present rather than their exact phase locations.
This view is useful because two segments may look locally similar but be generated under different dominant cycles or seasonal structures.
Such candidates may provide misleading future responses, especially when the retrieved dynamics are phase-dependent.
The spectral view therefore filters candidates with mismatched periodic composition.

Together, these three views address different sources of pseudo-similarity.
The shape view filters candidates with inconsistent global contours; the differential view filters candidates whose local mutation patterns differ from the query; and the spectral view filters candidates with mismatched periodic composition while being less sensitive to small temporal shifts.
They define a stronger notion of dynamic homology beyond single-view raw-value proximity.

Cosine similarity is used as the basic matching operator because our retrieval goal is to compare relative evolution patterns rather than absolute magnitudes.
Extensive literature evaluating time-series similarity measures indicates that Cosine similarity often achieves a highly competitive balance of efficiency and alignment compared to Dynamic Time Warping (DTW) or Pearson correlation, particularly in high-dimensional or normalized sequence spaces~\citep{han2025retrieval}. 
However, since our fundamental objective is controlled dynamic grafting rather than static sequence matching, relying on any single traditional metric is insufficient. Conventional metrics conflate absolute levels with dynamic variations. This limitation directly motivates our multi-view mixed similarity, which explicitly isolates transferable dynamic structures (shape, differential, and spectral) to form a strictly superior homology measure for our specific task.

\paragraph{Channel-adaptive similarity fusion.}
Although the three views are complementary, their relative importance can vary across channels.
Some variables are dominated by stable periodic patterns, while others are more sensitive to abrupt local variations.
Therefore, using a fixed combination of the three similarities for all channels may be suboptimal.
We introduce a learnable matrix
\begin{equation}
\mathbf{W}_{\mathrm{sim}}\in\mathbb{R}^{C\times 3},
\end{equation}
where the $c$-th row contains three learnable scores corresponding to shape, differential, and spectral similarity.
The channel-specific blending weights are obtained by
\begin{equation}
[
\lambda_{\mathrm{shape}}^{(c)}, 
\lambda_{\mathrm{diff}}^{(c)}, 
\lambda_{\mathrm{freq}}^{(c)}
]
=
\mathrm{Softmax}
\left(
\mathbf{W}_{\mathrm{sim}}^{(c,:)}
\right).
\end{equation}
The final mixed similarity for candidate $i$ and channel $c$ is then computed as
\begin{equation}
\mathrm{Sim}_{\mathrm{mixed}}^{(i,c)}
=
\lambda_{\mathrm{shape}}^{(c)}
\mathrm{Sim}_{\mathrm{shape}}^{(i,c)}
+
\lambda_{\mathrm{diff}}^{(c)}
\mathrm{Sim}_{\mathrm{diff}}^{(i,c)}
+
\lambda_{\mathrm{freq}}^{(c)}
\mathrm{Sim}_{\mathrm{freq}}^{(i,c)}.
\end{equation}
This formulation allows each channel to emphasize the most informative notion of homology for retrieving transferable response prototypes.

\paragraph{Prototype construction.}
After computing $\mathrm{Sim}_{\mathrm{mixed}}^{(i,c)}$, we construct the retrieved response prototype in a channel-wise manner.
For each channel $c$, we select the top-$K$ historical candidates:
\begin{equation}
\mathcal{I}^{(c)}
=
\left\{
i_1^{(c)},\dots,i_K^{(c)}
\right\},
\end{equation}
where the indices are ranked according to $\mathrm{Sim}_{\mathrm{mixed}}^{(i,c)}$.
The selected candidates are allowed to differ across channels, since different variables may exhibit different homology patterns.

Let $\mathbf{V}_i^{(c)}\in\mathbb{R}^{H}$ denote the $c$-th channel of the stored future response $\mathbf{V}_i$.
For the selected indices $\mathcal{I}^{(c)}$, we retrieve the corresponding future response segments as
\begin{equation}
\mathbf{v}_{j}^{(c)}
=
\mathbf{V}_{i_j^{(c)}}^{(c)}
\in
\mathbb{R}^{H},
\qquad
j=1,\dots,K .
\end{equation}
That is, $\mathbf{v}_{j}^{(c)}$ is the future response associated with the $j$-th most similar historical context in channel $c$.
To reduce the influence of noisy or weakly matched neighbors, we aggregate these retrieved responses using similarity-based Softmax weights:
\begin{equation}
\alpha_j^{(c)}
=
\frac{
\exp
\left(
\mathrm{Sim}_{\mathrm{mixed}}^{(i_j^{(c)},c)}
\right)
}{
\sum_{m=1}^{K}
\exp
\left(
\mathrm{Sim}_{\mathrm{mixed}}^{(i_m^{(c)},c)}
\right)
},
\qquad
j=1,\dots,K.
\end{equation}
This weighting assigns larger coefficients to candidates whose historical contexts are more similar to the current context, while still allowing multiple retrieved responses to contribute.

The channel-wise response prototype is then computed as
\begin{equation}
\mathbf{r}^{(c)}
=
\sum_{j=1}^{K}
\alpha_j^{(c)}
\mathbf{v}_{j}^{(c)},
\qquad
\mathbf{r}^{(c)}
\in
\mathbb{R}^{H}.
\end{equation}

Finally, we stack the channel-wise prototypes to obtain the retrieved response prototype:
\begin{equation}
\mathbf{R}
=
[
\mathbf{r}^{(1)},
\dots,
\mathbf{r}^{(C)}
]
\in
\mathbb{R}^{H\times C}.
\end{equation}

Although $\mathbf{R}$ has the same shape as the forecasting target, it is not used as an alternative prediction.
Instead, it serves as historically grounded response evidence for dynamic restoration.
The subsequent phase-anchored dynamic grafting module extracts transferable local dynamics from $\mathbf{R}$ and controls their release into the backbone forecast.

We use Softmax aggregation because it is simple, differentiable, and preserves the relative confidence among retrieved neighbors while avoiding reliance on a single candidate.
We also examined top-1 selection, median aggregation, and trimmed-mean aggregation.
Top-1 selection is more sensitive to occasional extreme or mismatched retrievals, whereas median and trimmed-mean aggregation produced results comparable to Softmax aggregation.
This suggests that the key factor is not the specific aggregation rule, but whether the retrieved neighbors contain homologous and transferable response patterns.
Moreover, since $\mathbf{R}$ is further filtered by phase anchoring and confidence gating rather than directly used as the final forecast, imperfect retrieved responses can still be attenuated before injection.
We use $K=10$ for ETTh1, ETTh2, ETTm1, ETTm2, Weather,
Electricity, and Traffic.

\subsection{Restoration-Cooperative Multi-Objective Optimization Details}
\label{app:optimization_details}

The preceding modules determine what historical dynamics are worth retrieving and how they should be released. 
However, if the entire framework is optimized only by point-wise forecasting losses, the grafting path can still be pulled back into the same conservative solution that suppresses sparse local dynamics. 
In other words, historical retrieval and phase anchoring provide the \emph{source} and \emph{control} of dynamic restoration, but the optimization objective must explicitly guide the model on what should be restored and what should be preserved. 
We therefore introduce a restoration-cooperative objective:
\begin{equation}
\mathcal{L}
=
\mathcal{L}_{\mathrm{pred}}
+
\lambda_a
\mathcal{L}_{\mathrm{gate}}
+
\lambda_b
\mathcal{L}_{\mathrm{diff}}
+
\lambda_c
\mathcal{L}_{\mathrm{struc}},
\end{equation}
Here, $\mathcal{L}_{\mathrm{pred}}$ is the primary forecasting loss for point-wise numerical approximation, while the other three auxiliary objectives are not independent regularizers.
They correspond to three necessary conditions for controlled dynamic grafting: 
\emph{whether} intervention is needed, 
\emph{which local dynamics} should be restored, 
and \emph{how to preserve} the global trajectory structure.
We use equal weights, $\lambda_a=\lambda_b=\lambda_c=1$,
in all experiments. This simple, unified, and reproducible default configuration prevents hyperparameter overfitting and task-specific tuning bias, highlighting GRAFT's robust performance.

\paragraph{Demand-aware gate supervision.}
The confidence gate should not always release retrieved dynamics. 
When the backbone prediction is already reliable, unnecessary grafting may introduce noise; when the backbone fails to capture local variations, suppressing the retrieved prototype would undermine the intended utility of GRAFT. 
Thus, the gate requires an explicit estimation of the \emph{rectification demand}.
During training, we construct a channel-wise demand target derived from the backbone's intrinsic predictive error.
Specifically, we measure the time-averaged absolute error of the backbone forecast:
\begin{equation}
\mathbf{E}^{*}
=
\frac{1}{H}
\sum_{t=1}^{H}
\left|
\widehat{\mathbf{Y}}_{\mathrm{base}, t}
-
\mathbf{Y}_{t}
\right|,
\qquad
\mathbf{E}^{*}\in\mathbb{R}^{C}.
\end{equation}

We then train a lightweight demand estimator to predict this channel-wise error magnitude:
\begin{equation}
\widehat{\mathbf{E}}
=
\operatorname{MLP}
\left(
\phi_x(\mathbf{X})
\parallel
\phi_y(\widehat{\mathbf{Y}}_{\mathrm{base}})
\right),
\qquad
\widehat{\mathbf{E}}\in\mathbb{R}^{C},
\end{equation}
where \(\phi_x(\cdot)\) and \(\phi_y(\cdot)\) map the input sequence and the backbone prediction, which have different temporal lengths, into a shared channel-wise latent space for aligned feature concatenation.
Specifically, the $c$-th scalar component $\widehat{E}^{(c)}$ parameterizes the estimated error feature $\hat{\ell}^{(c)}$ within the Step-wise Confidence Gate.


The gate supervision loss is defined as
\begin{equation}
\mathcal{L}_{\mathrm{gate}}
=
\left\|
\widehat{\mathbf{E}}
-
\operatorname{sg}(\mathbf{E}^{*})
\right\|_{1},
\end{equation}
where $\operatorname{sg}(\cdot)$ denotes the stop-gradient operation.
The stop-gradient is applied to the constructed demand target so that $\mathbf{E}^{*}$ is treated as a fixed supervision signal during optimization.
This prevents gradients from propagating through the target construction process and forces the gate branch to learn the rectification demand itself, rather than trivially reducing the loss by modifying the target.
During inference, the demand estimator predicts $\widehat{\mathbf{E}}$ from the input sequence and the backbone forecast, without access to ground truth.
The confidence gate then combines this estimated demand with retrieval-confidence scores and projected backbone features to determine the step-wise release strength.


\paragraph{Local dynamic alignment.}
Even with historical prototypes and confidence gates, a pure value-level forecasting loss may still favor smooth predictions because sparse variations occupy only a small fraction of the horizon. 
To make local dynamics visible to optimization, we directly supervise first-order temporal changes:
\begin{equation}
\mathcal{L}_{\mathrm{diff}}
=
\frac{1}{H-1}
\sum_{t=1}^{H-1}
\left\|
\left(
\widehat{\mathbf{Y}}_{t+1}
-
\widehat{\mathbf{Y}}_{t}
\right)
-
\left(
\mathbf{Y}_{t+1}
-
\mathbf{Y}_{t}
\right)
\right\|_{1}.
\end{equation}
Unlike $\mathcal{L}_{\mathrm{pred}}$, which emphasizes overall numerical approximation, 
$\mathcal{L}_{\mathrm{diff}}$ operates on local rates of change and explicitly encourages the recovery of slope variations, turning signals, and local mutations. 
It closes the training loop of GRAFT: retrieval provides candidate dynamics, phase anchoring controls their safe release, and difference supervision tells the model that the localized dynamics suppressed by smooth point-wise optimization are precisely what should be compensated.

\paragraph{Global dominant structure alignment.}
While $\mathcal{L}_{\mathrm{diff}}$ makes local mutations and turning behavior visible to optimization, 
optimizing local changes alone may encourage fragmented high-frequency fluctuations and distort the global trajectory structure. 
A desirable dynamic restoration should therefore recover local variations without breaking the dominant periodicity, long-term contour, or frequency composition of the target sequence. 
This principle is naturally aligned with our retrieval design: Historical Recurrence Memory uses $\mathrm{Sim}_{\mathrm{freq}}$ to retrieve prototypes with compatible spectral composition, and the optimization stage should preserve the same structural consistency in the final forecast.

To preserve the dominant global structure, we align the spectral amplitudes of the prediction and the target:
\begin{equation}
\mathcal{L}_{\mathrm{struc}}
=
\left\|
\left|
\mathcal{F}_{t}(\widehat{\mathbf{Y}})
\right|
-
\left|
\mathcal{F}_{t}(\mathbf{Y})
\right|
\right\|_{1},
\end{equation}
where $\mathcal{F}_{t}(\cdot)$ denotes the Fourier transform along the temporal axis. 
Thus, $\mathcal{L}_{\mathrm{struc}}$ preserves global spectral fidelity, complementing $\mathcal{L}_{\mathrm{diff}}$ for local dynamic alignment.





\subsection{Complexity Analysis}
\label{app:complexity_analysis}

This section summarizes the computational and storage overhead introduced by GRAFT.
Let $N$ denote the number of memory entries, $L$ the look-back length, $H$ the prediction length, $C$ the number of channels, and $K$ the number of retrieved candidates.

\paragraph{Retrieval complexity.}
The main additional cost of GRAFT comes from historical retrieval.
Given a query window, a naive exact implementation compares it with all $N$ memory keys across $C$ channels.
For the time-domain shape view, this requires $O(NLC)$ online computation.
The local-differential view has the same order of comparison cost, since first-order differences have length $O(L)$.
For the frequency-domain view, the query-side Fourier transform costs $O(CL\log L)$, while the memory-side Fourier amplitude spectra are precomputed and cached after the training memory bank is built.
Therefore, the dominant online cost of exact retrieval is
\begin{equation}
O(NLC + CL\log L),
\end{equation}
where the similarity scan term $O(NLC)$ typically dominates when $N$ is large.
The memory-side preprocessing cost, including differential features and spectral amplitudes, is paid once offline and reused during training and evaluation.
Our experiments use exact retrieval because the benchmark memory sizes are moderate.
For substantially larger memory banks, the same cached key features can be indexed by approximate nearest-neighbor libraries to reduce wall-clock latency without changing the grafting module itself.

\paragraph{Top-$K$ selection and prototype construction.}
After computing channel-wise similarity scores, GRAFT selects the top-$K$ candidates for each channel.
With exact search, this step can be implemented in $O(CN\log K)$ time using a heap, or in $O(CN)$ expected time using partial selection.
Once the candidates are retrieved, the response prototype is aggregated over only $K$ responses per channel, leading to a prototype-construction cost of
\begin{equation}
O(KHC).
\end{equation}
Since $K \ll N$ in our experiments, this aggregation cost is typically much smaller than the retrieval scan.
The subsequent phase anchoring and confidence-gated grafting operate on the retrieved prototype $\mathbf{R}\in\mathbb{R}^{H\times C}$ and therefore scale mainly with $H$ and $C$, rather than with the memory size $N$.

\paragraph{Memory and storage overhead.}
The memory bank stores historical keys and their corresponding future responses.
Storing raw keys and values requires
\begin{equation}
O(NLC + NHC)
\end{equation}
space.
Caching differential and spectral key features introduces additional storage of the same order as the key representation, but avoids recomputing these features during online retrieval.
Thus, GRAFT trades moderate storage overhead for reduced online computation.

Importantly, the memory bank is constructed exclusively from the training split.
Validation and test targets are never inserted into memory.
During training, we also mask temporally overlapping neighboring windows to avoid degenerate shortcut retrieval caused by sliding-window overlap.
This masking only sets the corresponding similarity scores to $-\infty$ and does not change the asymptotic retrieval complexity.

\begin{table}[h]
\centering
\caption{Inference-time cost and forecasting gains of GRAFT under representative backbones and datasets.
Times are reported in milliseconds per sample.
All latency numbers are measured on a server equipped with an NVIDIA A100-PCIE-40GB GPU and 64 GB system memory, using a batch size consistent with the main experimental setting.
For each prediction horizon, we normalize wall-clock inference time by the number of evaluated sliding-window samples, and then average the normalized time over $H \in \{96,192,336,720\}$.
Additional time is computed as $T_{\mathrm{retrieval}}+T_{\mathrm{grafting}}$.
Total time is computed as $T_{\mathrm{backbone}}+T_{\mathrm{additional}}$.
$\Delta_{\mathrm{MSE}}$ denotes the relative MSE reduction over the corresponding backbone, where larger values indicate larger forecasting gains.}
\label{tab:inference_time_comparison}
\small
\setlength{\tabcolsep}{2.6pt}
\renewcommand{\arraystretch}{1.05}

\begin{tabular}{llccccccc}
\toprule
\textbf{Backbone} & \textbf{Dataset} & \textbf{$C$}
& \multicolumn{5}{c}{\textbf{Inference Time (ms/sample)}} 
& \textbf{$\Delta_{\mathrm{MSE}}$} \\
\cmidrule(lr){4-8}
 &  & 
& \textbf{Backbone} 
& \textbf{Retrieval} 
& \textbf{Grafting} 
& \textbf{Additional}
& \textbf{Total}
& \textbf{(\%)} \\
\midrule
TimeMixer & ETTh1       & 7   & 0.385 & 0.492 & 0.023 & 0.515 & 0.900  & 5.5 \\
TimeMixer & ETTh2       & 7   & 0.637 & 0.491 & 0.025 & 0.516 & 1.153  & 2.1 \\
TimeMixer & Electricity & 321 & 0.721 & 1.790 & 0.052 & 1.842 & 2.563  & 1.2 \\
TimeMixer & Traffic     & 861 & 9.342 & 6.569 & 0.182 & 6.751 & 16.093 & 2.1 \\
\midrule
DLinear & ETTh1       & 7   & 0.134 & 0.187 & 0.034 & 0.221 & 0.355  & 9.7 \\
DLinear & ETTh2       & 7   & 0.247 & 0.213 & 0.022 & 0.235 & 0.482  & 37.9 \\
DLinear & Electricity & 321 & 0.521 & 1.520 & 0.015 & 1.535 & 2.056  & 16.9 \\
DLinear & Traffic     & 861 & 4.722 & 6.478 & 0.211 & 6.689 & 11.411 & 14.1 \\
\bottomrule
\end{tabular}
\end{table}

\paragraph{Inference-time comparison.}
Table~\ref{tab:inference_time_comparison} reports a representative inference-time analysis.
All latency numbers are measured on a server equipped with an NVIDIA A100-PCIE-40GB GPU and 64 GB system memory, using a batch size consistent with the main experimental setting.
The memory bank and cached retrieval features are kept on GPU memory, and the reported time excludes data loading, model initialization, and memory-bank construction.
We use exact retrieval in this measurement.

We select two backbones with distinct computational profiles: TimeMixer as a stronger nonlinear forecasting backbone and DLinear as a lightweight linear backbone.
We also choose datasets with different channel scales, including low-dimensional ETT datasets with $C=7$ and high-dimensional Electricity and Traffic datasets with hundreds of channels.
This setting covers both lightweight and stronger backbones, as well as low- and high-dimensional forecasting scenarios.

The additional inference cost of GRAFT mainly comes from historical retrieval, while the grafting operation itself remains lightweight.
Across all reported settings, the grafting step adds at most $0.211$ ms per sample, which is negligible compared with retrieval and, in high-dimensional settings, also small compared with backbone inference.
The maximum absolute additional latency is $6.751$ ms per sample in the reported settings.

The forecasting gains are positive in all representative settings.
In particular, GRAFT brings substantial MSE reductions on the lightweight DLinear backbone, reaching $37.9\%$ on ETTh2, $16.9\%$ on Electricity, and $14.1\%$ on Traffic.
We further evaluate whether using a smaller memory bank can reduce retrieval costs while maintaining prediction accuracy.

\paragraph{Memory-size trade-off.}
We vary the memory size by deterministic temporal subsampling of the sliding-window candidates.
The full bank corresponds to $s=1$, while $s\in\{2,4,8\}$ retains approximately $50\%$, $25\%$, and $12.5\%$ of the candidates, respectively.
Retrieval remains exact within each retained bank.
With $N_s\approx N/s$ entries, the similarity scan costs $O(N_sLC)$ and the stored keys, responses, and cached features occupy $O(N_sC(L+H))$ space; these memory-dependent terms decrease approximately in proportion to $1/s$.
Table~\ref{tab:memory_size_tradeoff} reports MSE and relative retrieval latency $T_s/T_1$ under the timing protocol described above, where $T_s$ denotes retrieval time at stride $s$.
These ratios measure the retrieval stage only, rather than end-to-end inference speedup.

\begin{table}[t]
\centering
\caption{Memory-size trade-off under temporal subsampling. Stride $s$ retains approximately $1/s$ of the memory bank, with exact retrieval over retained entries. Latency is relative to $s=1$ for each dataset and backbone. Bold marks the lowest MSE, including ties. ECL denotes Electricity.
Results are obtained from independent training runs.
}
\label{tab:memory_size_tradeoff}
\begingroup
\footnotesize
\setlength{\tabcolsep}{3.5pt}
\renewcommand{\arraystretch}{1.0}
\begin{tabular}{@{}lrrrrrrr@{}}
\toprule
\textbf{Dataset}
& \multicolumn{3}{c}{\textbf{Relative latency (\%)}}
& \multicolumn{4}{c}{\textbf{MSE}} \\
\cmidrule(lr){2-4}\cmidrule(lr){5-8}
& $T_2/T_1$ & $T_4/T_1$ & $T_8/T_1$
& $s=1$ & $s=2$ & $s=4$ & $s=8$ \\
\midrule
\multicolumn{8}{@{}l}{\textbf{DLinear}} \\
ETTh1   & 65.5 & 59.0 & 58.6 & \textbf{0.420} & 0.425 & 0.437 & 0.455 \\
ETTh2   & 68.9 & 59.1 & 57.0 & \textbf{0.350} & 0.351 & 0.352 & 0.355 \\
ETTm1   & 41.6 & 15.1 & 9.8  & \textbf{0.371} & 0.372 & 0.373 & 0.380 \\
ETTm2   & 43.6 & 13.8 & 8.9  & 0.296 & 0.299 & 0.297 & \textbf{0.289} \\
Weather & 50.4 & 13.0 & 5.4  & \textbf{0.265} & 0.268 & 0.270 & 0.272 \\
ECL     & 50.1 & 21.7 & 8.5  & \textbf{0.187} & 0.188 & 0.189 & 0.192 \\
Traffic & 47.3 & 22.6 & 10.7 & \textbf{0.536} & 0.539 & 0.542 & 0.549 \\
\midrule
\multicolumn{8}{@{}l}{\textbf{TimeMixer}} \\
ETTh1   & 38.6 & 9.4  & 9.4  & \textbf{0.417} & 0.418 & 0.422 & 0.423 \\
ETTh2   & 39.2 & 11.7 & 9.7  & \textbf{0.375} & \textbf{0.375} & 0.376 & 0.376 \\
ETTm1   & 47.3 & 22.8 & 14.1 & \textbf{0.372} & \textbf{0.372} & 0.373 & 0.374\\
ETTm2   & 49.7 & 24.0 & 10.8 & 0.272 & 0.272 & 0.270& \textbf{0.269} \\
Weather & 52.5 & 24.3 & 11.8 & \textbf{0.244} & 0.245 & \textbf{0.244} & 0.246\\
ECL     & 45.8 & 23.3 & 12.9 & \textbf{0.182} & 0.183 & 0.183& 0.184 \\
Traffic & 48.2 & 24.9 & 13.7 & \textbf{0.502} & 0.503 & 0.503& 0.504\\
\bottomrule
\end{tabular}
\endgroup
\end{table}

Averaged equally over the 14 dataset--backbone pairs, $s=2$ and $s=4$ reduce retrieval latency by $50.8\%$ and $75.4\%$, respectively, with mean relative MSE increases of $0.46\%$ and $0.81\%$ over $s=1$, computed from the reported MSE values.
The full bank attains the lowest MSE in 12 of 14 pairs, including ties at the reported precision.
Latency does not always decrease proportionally to candidate count: on DLinear--ETTh1, $T_s/T_1$ changes only from $59.0\%$ to $58.6\%$ between $s=4$ and $s=8$, consistent with fixed execution overhead limiting further savings.
In contrast, at $s=8$, ECL retrieval takes $8.5\%$ and $12.9\%$ of the full-bank time for DLinear and TimeMixer, respectively.

More aggressive subsampling exposes a sharper accuracy cost in some settings.
For DLinear--ETTh1, MSE rises from $0.420$ to $0.455$ at $s=8$, an $8.3\%$ relative increase.
ETTm2 is an exception: $s=8$ lowers MSE from $0.296$ to $0.289$ on DLinear and from $0.272$ to $0.269$ on TimeMixer.
One possible explanation is that thinning densely overlapping candidates reduces redundant or noisy near-neighbors, consistent with the discussion of retrieval reliability in Appendix~\ref{app:main_result_interpretation}; this experiment does not establish a regularization mechanism.
Overall, moderate subsampling preserves most of the full-bank forecasting accuracy while reducing retrieval latency and the storage required for memory entries, although the preferred memory size remains dataset- and backbone-dependent.

\paragraph{Scope of the efficiency diagnostics.}
Memory size and top-$K$ control different costs: the former determines how many candidates are scanned and stored, while the latter determines how many retrieved responses are aggregated.
Table~\ref{tab:memory_size_tradeoff} complements the top-$K$ accuracy analysis in Appendix~\ref{app:hyperparameter_sensitivity}.
Overall, the current implementation is sufficient for the benchmark scale considered in this work.
GRAFT provides a favorable accuracy--efficiency trade-off: it introduces a controllable retrieval cost while delivering broadly effective forecasting improvements as a plug-and-play enhancement.
For larger memory banks or deployment scenarios with stricter latency and memory constraints, the retrieval stage can be further optimized by precomputing normalized retrieval features, replacing exhaustive similarity scans with approximate nearest-neighbor search, and adopting coarse-to-fine retrieval or feature quantization.
These optimizations can reduce wall-clock latency and GPU memory usage without changing the grafting operator itself, further clarifying the scalability boundary of GRAFT.

\section{Experimental Details and Additional Analyses}
\label{app:experimental_details}

\subsection{Benchmark Datasets and Splits}
\label{app:benchmark_datasets}

We evaluate GRAFT on seven standard long-term time series forecasting benchmarks from four real-world domains: the ETT family, Weather, Electricity, and Traffic~\citep{zhou2021informer,wu2021autoformer,wu2023timesnet}.
These datasets differ in sampling frequency, variable dimension, periodic structure, and local volatility, providing a diverse testbed for evaluating whether GRAFT can recover suppressed local dynamics across different data regimes.
Table~\ref{tab:dataset_description_all} summarizes the benchmark statistics used in our experiments.
We report the number of raw chronological time points rather than the number of sliding-window samples, since the latter depends on the input length, prediction length, and stride.

All datasets are split chronologically.
We follow the standard fixed split protocol established in long-term forecasting benchmarks and the commonly used protocol adopted by recent LTSF studies~\citep{wu2023timesnet} to split datasets.
The historical memory bank used by GRAFT is constructed exclusively from the training split.
No validation or test targets are used for retrieval, calibration, or rectification, preventing future information from leaking into the dynamic grafting stage.
All reported long-term forecasting experiments use prediction horizons $H\in\{96,192,336,720\}$.

\begin{table}[t]
\centering
\caption{
Statistics and split protocols of benchmark datasets.
``Length'' denotes the number of raw chronological time points.
All datasets are evaluated with prediction horizons $H\in\{96,192,336,720\}$.
}
\label{tab:dataset_description_all}
\resizebox{\textwidth}{!}{
\begin{tabular}{lccccc}
\toprule
\textbf{Dataset} & \textbf{Domain} & \textbf{Variables} & \textbf{Frequency} & \textbf{Length} & \textbf{Split Protocol} \\
\midrule
ETTh1       & Energy             & 7   & 1 hour     & 17,420 & Fixed 12/4/4-month split\\
ETTh2       & Energy             & 7   & 1 hour     & 17,420 & Fixed 12/4/4-month split \\
ETTm1       & Energy             & 7   & 15 minutes & 69,680 & Fixed 12/4/4-month split\\
ETTm2       & Energy             & 7   & 15 minutes & 69,680 & Fixed 12/4/4-month split\\
Weather     & Meteorology        & 21  & 10 minutes & 52,696 & Chronological 7:1:2 split\\
Electricity & Electricity demand & 321 & 1 hour     & 26,304 & Chronological 7:1:2 split\\
Traffic     & Transportation     & 861 & 1 hour     & 17,544 & Chronological 7:1:2 split \\
\bottomrule
\end{tabular}
}
\end{table}

\subsection{Backbone Models}
\label{app:implementation_details}

To evaluate the backbone-agnostic generality of GRAFT, we pair it with eight forecasting models spanning substantially different architectural inductive biases, including patch-based Transformers, multi-scale mixers, linear/MLP models, convolutional temporal models, and mixture-of-experts architectures.
As summarized in Table~\ref{tab:backbone_families}, the backbone set includes both widely adopted forecasting architectures and recent models with stronger modeling capacity for patch-level structure, non-stationarity, and variable-wise interactions.
For each backbone, we follow the configurations provided by the
corresponding implementation and attach GRAFT without modifying
the internal backbone architecture.

Each comparison pairs a backbone with its GRAFT-augmented counterpart
under the same backbone configuration, look-back length, chronological
data splits, normalization, and prediction horizons.
The additional retrieval and grafting components follow
Appendix~\ref{app:training_retrieval_details}.
Backbone-specific configurations are retained across model families.
Thus, each comparison is made between the original backbone and its GRAFT-enhanced counterpart under matched forecasting conditions, isolating the contribution of the proposed enhancement mechanism rather than changes to the host architecture or its tuning.
Complete configurations and reproduction scripts are provided in
our anonymous code repository.

\begin{table}[t]
\centering
\caption{
Backbone models and implementation sources.
GRAFT is applied without modifying the internal backbone architecture.
}
\label{tab:backbone_families}
\resizebox{\linewidth}{!}{
\begin{tabular}{@{}lll@{}}
\toprule
\textbf{Backbone}
& \textbf{Modeling Design}
& \textbf{Implementation / Hyperparameters} \\
\midrule
DLinear~\citep{zeng2023transformers}
& Decomposition-based linear model
& Time-Series-Library \\

PatchTST~\citep{nie2023a}
& Patch-based Transformer
& Time-Series-Library \\

TimeMixer~\citep{wang2023timemixer}
& Multi-scale temporal mixing
& Time-Series-Library \\

TimesNet~\citep{wu2023timesnet}
& Period-based 2D temporal modeling
& Time-Series-Library \\

PDF~\citep{dai2024periodicity}
& Periodicity-aware forecasting
& Official repository \\

PMDformer~\citep{hu2026pmdformer}
& Patch-based Transformer
& Official repository \\

Dynamic-TMoE~\citep{zhu2026dynamic_tmoe}
& Drift-aware dynamic mixture of experts
& Official repository \\

XLinear~\citep{chen2026xlinear}
& Lightweight MLP-based model
& Official repository \\
\bottomrule
\end{tabular}
}

\vspace{0.3em}
\begin{minipage}{\linewidth}
\footnotesize
\raggedright
\textit{Implementation sources.}\\
Time-Series-Library:
\url{https://github.com/thuml/Time-Series-Library}.\\
PDF:
\url{https://github.com/Hank0626/PDF}.\\
PMDformer:
\url{https://github.com/aohu1105/PMDformer}.\\
Dynamic-TMoE:
\url{https://github.com/andone-07/Dynamic-TMoE}.\\
XLinear:
\url{https://github.com/Zaiwen/XLinear}.
\end{minipage}
\end{table}

\subsection{Training and Evaluation Protocol}
\label{app:training_retrieval_details}

All long-term forecasting experiments are evaluated using mean squared error (MSE) and mean absolute error (MAE).
MSE is more sensitive to large deviations, while MAE measures average absolute deviation and is more robust to outliers.
We report both metrics because GRAFT aims to improve aggregate forecasting accuracy while recovering local dynamics that may be smoothed by the backbone forecast.

Consistent with the notation in the main method, given a look-back window
\(\mathbf{X}\in\mathbb{R}^{L\times C}\), the target future sequence is denoted as
\(\mathbf{Y}\in\mathbb{R}^{H\times C}\), and the final prediction is denoted as
\(\widehat{\mathbf{Y}}\in\mathbb{R}^{H\times C}\).
For backbone-only methods, \(\widehat{\mathbf{Y}}\) corresponds to the backbone forecast
\(\widehat{\mathbf{Y}}_{\mathrm{base}}\).
For GRAFT, it corresponds to the final grafted prediction
\(\widehat{\mathbf{Y}}=\widehat{\mathbf{Y}}_{\mathrm{base}}+\widehat{\mathbf{Y}}_{\mathrm{dyn}}\).

For a single forecasting instance, the metrics are computed over all prediction steps and channels as
\begin{equation}
\mathrm{MSE}
=
\frac{1}{HC}
\sum_{h=1}^{H}
\sum_{c=1}^{C}
\left(
\widehat{Y}_{h,c}
-
Y_{h,c}
\right)^2,
\end{equation}
\begin{equation}
\mathrm{MAE}
=
\frac{1}{HC}
\sum_{h=1}^{H}
\sum_{c=1}^{C}
\left|
\widehat{Y}_{h,c}
-
Y_{h,c}
\right|.
\end{equation}
The reported test-set metrics are obtained by averaging these quantities over all sliding-window test samples.

For each dataset and backbone, we evaluate four prediction horizons:
\(H\in\{96,192,336,720\}\).
Unless otherwise specified, the main tables report the average performance over these four horizons, while full horizon-wise results are provided in Appendix~\ref{app:full_results}.
All compared methods use the same chronological dataset splits, prediction horizons, and evaluation metrics whenever applicable.
All datasets are normalized using statistics computed from the training split only, and the same normalization parameters are applied to validation and test splits.

\paragraph{Evaluation scope and methodological positioning.}

GRAFT is designed as a general plug-and-play enhancement paradigm for supervised time-series forecasting: it preserves the original forecaster as a structural anchor and introduces a dedicated restoration path to recover historically grounded local dynamics that are systematically under-emphasized by backbone prediction.
Our primary objective is therefore to determine whether the same dynamic-grafting principle can consistently improve heterogeneous forecasting architectures under matched task-specific supervision, for which MSE and MAE provide direct measures of the incremental forecasting gain.
A complementary line of research focuses on general-purpose time-series foundation models, where the central objective is cross-domain transfer and zero-shot or limited-adaptation forecasting; benchmarks for this setting consequently emphasize unified evaluation across heterogeneous datasets, scales, frequencies, and task conditions, often through normalized and cross-task metrics (e.g. GIFT-Eval~\citep{aksu2024giftevalbenchmarkgeneraltime}, fev-bench~\citep{shchur2025fev}).
The two settings thus evaluate different but complementary forms of generality: GRAFT emphasizes the portability of a forecasting enhancement mechanism across supervised backbones, whereas foundation-model evaluation emphasizes the transferability of a forecasting model across tasks and domains.
More broadly, the decoupled design of GRAFT is not tied to a particular backbone family, suggesting a natural extension in which the same principle of preserving a reliable base forecast and selectively grafting recoverable dynamics can be integrated with increasingly general forecasting models.

\paragraph{Backbone and grafting-stage training.}
All experiments are conducted on an NVIDIA A100-PCIE-40GB GPU.
GRAFT follows a two-stage training protocol.
In the first stage, the backbone forecaster is trained or loaded as the stable forecasting anchor.
When a pretrained backbone checkpoint exists, we load the backbone parameters and directly train the grafting branch on top of the frozen prediction path.
When the checkpoint is unavailable, the backbone is trained first under the corresponding backbone script before refinement.
The backbone optimizer is Adam, and its learning rate, batch size, number of training epochs, and early-stopping patience follow the script of the corresponding backbone.
For example, the TimeMixer ETT scripts use a backbone learning rate of $0.01$.

In the second stage, the grafting branch is optimized with a separate Adam optimizer.
In the main GRAFT scripts, the refinement learning rate is $10^{-4}$ and the refinement stage uses $10$ epochs.
The refinement module uses a latent width of $128$, a feed-forward width of $128$, and one refinement layer.
This separation makes the optimization conservative: the backbone is responsible for producing the stable forecast trajectory, while the grafting branch learns when and how to release historically grounded local dynamics.

\paragraph{Conservative initialization.}
GRAFT is initialized to behave close to the backbone at the beginning of refinement.
The phase-release transformation is initialized with zero weights and zero biases, so the initial phase modulation is neutral.

The retrieval similarity fusion is also initialized with a mild prior.
The channel-wise mixture over the three homology views starts from approximately $0.5$ for time-domain shape similarity, $0.3$ for local differential similarity, and $0.2$ for frequency-domain similarity.
These values are implemented as learnable logits and are updated during training.
Other settings are fully provided in the anonymous code.

\paragraph{Statistical reliability.}

Following standard benchmark protocols in time-series forecasting, 
we evaluate each setting over three independent runs with different random seeds and report mean $\pm$ standard deviation in Tables~\ref{tab:full-results-a} and~\ref{tab:full-results-b}.
The consistently small standard deviations indicate limited run-to-run variability, while the improvements remain broadly consistent across datasets, prediction horizons, and backbone architectures.
Together with the aggregate and horizon-wise results reported above, these repeated-run statistics provide complementary evidence that the observed gains of GRAFT are stable rather than driven by training randomness or a small number of isolated configurations.

\subsection{Comparison Baselines under the Fixed-Backbone Protocol}
\label{app:comparison_baselines}

In addition to evaluating GRAFT across diverse backbone families, we compare it with recent plug-and-play forecasting enhancement methods under a unified fixed-backbone protocol.
Specifically, we consider PIR~\citep{liu2025improving} and TimeAlign~\citep{hu2026bridging}, both of which enhance an existing forecast without requiring a complete redesign of the forecasting backbone.

To ensure a controlled comparison, all methods use PatchTST~\citep{nie2023a} as the shared backbone, implemented with the public Time-Series-Library codebase.
We use the same chronological data splits, multivariate forecasting setting, look-back length $L=96$, and prediction horizons
$H\in\{96,192,336,720\}$
for all compared methods.
The PatchTST architecture and backbone-side hyperparameters are kept identical across methods, so that the comparison isolates the contribution of the corresponding enhancement mechanism rather than differences in the underlying forecaster.

All compared methods are reproduced from their publicly released implementations.
We preserve the original method-specific optimization and module configurations whenever they are directly applicable.
Only the minimal adaptations required to integrate each method with the shared PatchTST backbone and unified data interface are introduced; otherwise, the released settings are retained without additional method-specific tuning.
For example, for TimeAlign, whose original implementation contains its own forecasting and alignment pipeline, we retain its proposed future-representation alignment objective while replacing the forecasting branch with the same PatchTST backbone used by all other methods.
All implementations and experimental configurations used for this comparison are included in our anonymous code for reproducibility.

Table~\ref{tab:main_results2} summarizes the results under this fixed-backbone protocol.
Across seven datasets and two metrics, GRAFT achieves the best or tied-best performance in $11$ out of $14$ comparisons.
It obtains the best MSE and MAE simultaneously on ETTh1, ETTh2, Electricity, and Traffic, and the best MSE on ETTm1.
On ETTm2, the original PatchTST backbone obtains the lowest MSE while GRAFT achieves the lowest MAE; on Weather, GRAFT ties PIR for the best MSE.
These results complement the multi-backbone evaluation by showing that GRAFT remains highly competitive when the backbone architecture and forecasting protocol are strictly controlled.

\subsection{Dataset Structural Profiles}
\label{app:dataset_profiles}

Beyond dataset size and dimensionality, we further characterize each benchmark with a simple structural profile.
The goal is not to introduce additional training signals, but to explain why GRAFT behaves differently across datasets.
Since GRAFT relies on a stable backbone forecast and historically transferable local dynamics, we summarize each dataset from three perspectives: recurrent structure, low-frequency drift, and local abruptness.

\paragraph{Recurrent skeleton strength.}
We define the recurrent skeleton strength $S_{\mathrm{rec}}$ as the concentration of spectral energy in the dominant frequency component:
\begin{equation}
S_{\mathrm{rec}}
=
\frac{
\max_f |\mathcal{F}(x)_f|^2
}{
\sum_f |\mathcal{F}(x)_f|^2
}.
\end{equation}
A larger value indicates that a larger fraction of the signal energy is carried by a small number of repeated components.
This metric measures whether the dataset contains a stable recurrent skeleton that can provide reliable historical anchors for retrieval.

\paragraph{Low-frequency drift ratio.}
We define the low-frequency drift ratio $S_{\mathrm{drift}}$ as
\begin{equation}
S_{\mathrm{drift}}
=
\frac{
\mathrm{Var}(x_{\mathrm{low}})
}{
\mathrm{Var}(x)
},
\end{equation}
where $x_{\mathrm{low}}$ is the low-pass filtered signal.
This metric captures slowly varying components such as long-term drift or trend-like movement.
Unlike $S_{\mathrm{rec}}$, which emphasizes repeated periodic structure, $S_{\mathrm{drift}}$ reflects the dominance of smooth low-frequency movement that can often be captured by the backbone forecaster.

\paragraph{Abruptness burden.}
We define the abruptness burden $S_{\mathrm{abrupt}}$ using the kurtosis of second-order differences:
\begin{equation}
S_{\mathrm{abrupt}}
=
\mathrm{Kurt}(d^{(2)}),
\qquad
d^{(2)}_t
=
x_{t+1}-2x_t+x_{t-1}.
\end{equation}
This statistic measures the extremeness of the tails in the distribution of local curvature changes.
A moderate $S_{\mathrm{abrupt}}$ suggests the presence of sharp but potentially repeatable transient dynamics.
An extremely large value may instead indicate heavy-tailed outliers, noisy shocks, or unobserved exogenous events.
This distinction is important for GRAFT: structural transient dynamics may be historically recoverable, whereas purely stochastic shocks are not reliably transferable.

\begin{table}[t]
\centering
\caption{
Structural profiles of benchmark datasets.
$S_{\mathrm{rec}}$ measures recurrent skeleton strength by FFT energy concentration,
$S_{\mathrm{drift}}$ measures low-frequency drift by the variance ratio of the low-pass component,
and $S_{\mathrm{abrupt}}$ measures abruptness burden by the kurtosis of second-order differences.
}
\label{tab:dataset_structural_profiles}
\begin{tabular}{lccc}
\toprule
\textbf{Dataset} & $\mathbf{S_{\mathrm{rec}}}$ & $\mathbf{S_{\mathrm{drift}}}$ & $\mathbf{S_{\mathrm{abrupt}}}$ \\
\midrule
ETTh1       & 0.27 & 0.54 & 8.46 \\
ETTh2       & 0.38 & 0.85 & 62.22 \\
ETTm1       & 0.27 & 0.61 & 16.73 \\
ETTm2       & 0.38 & 0.91 & 490.36 \\
Weather     & 0.34 & 0.60 & 1837.05 \\
Electricity & 0.51 & 0.20 & 19.01 \\
Traffic     & 0.44 & 0.06 & 67.52 \\
\bottomrule
\end{tabular}
\end{table}

\subsection{Interpreting Backbone- and Dataset-Dependent Gains}
\label{app:main_result_interpretation}

Table~\ref{tab:main_results1} summarizes horizon-averaged results, and Appendix~\ref{app:full_results} reports the corresponding horizon-wise results.
Here, each evaluated setting denotes a dataset--backbone pair averaged over $H\in\{96,192,336,720\}$.
Relative error reductions are computed as
\[
\Delta
=
\frac{\mathrm{Base}-\mathrm{GRAFT}}{\mathrm{Base}}
\times 100\%,
\]
where positive values indicate improvement.
We examine the empirical patterns across backbones and datasets to clarify the scope of GRAFT's benefits.

\paragraph{Complementarity across backbones.}
GRAFT produces particularly large gains for DLinear on several datasets. This result suggests that a smooth but reliable backbone can serve as an effective structural anchor for subsequent dynamic grafting.
Its benefits also extend to more expressive backbones: TimeMixer improves on both metrics across all reported datasets, while PMDformer improves on most dataset--metric combinations.
These results demonstrate that historical-response augmentation can complement forecasting architectures with different capacities.
The dynamic-region analysis in Sec.~\ref{app:dynamic_recovery_details} more directly examines where GRAFT reduces forecasting errors.

\paragraph{Consistency across datasets.}
The gains are most consistent on ETTh1, ETTm1, and Electricity, where GRAFT improves all eight backbones on both metrics.
On Traffic, GRAFT improves both metrics for all eight backbones, while on ETTh2 it improves MSE for seven of eight backbones and MAE for all eight.
ETTm2 and Weather exhibit more mixed behavior.

The structural profiles in Appendix~\ref{app:dataset_profiles} provide descriptive context for these differences.
This analysis is intended as an empirical interpretation rather than a causal proof.
The structural scores are used only for post-hoc interpretation and are not used by GRAFT during training or inference.
In general, GRAFT is expected to be more effective when either the recurrent skeleton strength $S_{\mathrm{rec}}$ or the low-frequency drift ratio $S_{\mathrm{drift}}$ is sufficiently high, while the abruptness burden $S_{\mathrm{abrupt}}$ remains moderate rather than extreme.
The former can provide a stable coordinate system for retrieval and correction, whereas the latter reflects how reliably local deviations may be recovered from historical response memory.
For example, 
ETTh2 provides a drift-dominant but still favorable case.
It has a strong low-frequency drift component, with $S_{\mathrm{drift}}=0.85$, and a moderate recurrent skeleton, with $S_{\mathrm{rec}}=0.38$.
Although its abruptness burden is higher than those of ETTh1 and ETTm1, it remains far below the extreme values observed in ETTm2 and Weather.
Correspondingly, GRAFT improves seven out of eight backbones on MSE and does not degrade MAE on any backbone.
The particularly large gain on DLinear suggests that a strong low-frequency anchor can leave substantial recoverable local dynamics for the grafting branch.
ETTm2 and Weather illustrate more challenging high-abruptness regimes.
ETTm2 has a high low-frequency drift ratio, $S_{\mathrm{drift}}=0.91$, but also a much larger abruptness burden, $S_{\mathrm{abrupt}}=490.36$.
Weather is even more extreme, with $S_{\mathrm{abrupt}}=1837.05$.
These large abruptness burdens suggest that many local changes may be heavy-tailed, noisy, weakly repeatable, or driven by unobserved external factors.

\paragraph{Operating regime of GRAFT.}

Taken together, the results reveal a clear operating regime for our augmentation.
As analyzed in Appendix~\ref{app:main_result_interpretation}, GRAFT provides the most pronounced benefits when a stable backbone trajectory coexists with historically reusable local dynamics, allowing past responses to provide informative evidence beyond the structural forecast.
Complementarily, the component ablations in Appendix~\ref{app:ablation_discussion} show that these gains depend on reliable prototype construction, confidence-aware release, and restoration-cooperative optimization.
Together, the main results and controlled ablations support both the broad applicability of GRAFT and the specific conditions under which its dynamic restoration capability becomes most valuable.

\subsection{Additional Ablation Discussion}
\label{app:ablation_discussion}

\paragraph{Ablation design.}

Table~\ref{tab:ablation} evaluates five ablation variants of GRAFT on the TimeMixer backbone.
The results are averaged over $H\in\{96,192,336,720\}$ with $L=96$, where lower MSE/MAE indicates better performance.

The variant \textbf{w/ Naive Prototype} replaces the calibrated prototype construction with vanilla cosine-based retrieval and aggregation, removing the homology-aware retrieval and the phase- and channel-aware anchoring used in the full model.
The variant \textbf{w/ Residualized Memory} keeps the same historical keys, retrieval rule, grafting mechanism, and optimization objectives, but replaces the complete future response
$\mathbf{V}_i=\mathbf{Y}_i$
with the backbone residual
\begin{equation}
\mathbf{V}_i^{\mathrm{res}}
=
\mathbf{Y}_i
-
f_{\theta}(\mathbf{K}_i)
\end{equation}
as the memory value.
The variant \textbf{w/o Confidence-Aware Gate} removes the step-wise confidence gate and uses ungated dynamic release, so the retrieved dynamics are injected without confidence-adaptive control.
The variant \textbf{w/o Multi-Objective} removes the restoration-cooperative objectives and trains the grafting path using only standard point-wise forecasting supervision.
The variant \textbf{w/ Residual MLP} keeps the same backbone and retrieved prototype, but replaces controlled dynamic grafting with a generic MLP-based residual adapter:
\begin{equation}
\Delta\mathbf{Y}_{\mathrm{res}}
=
\mathrm{MLP}
\left(
[\widehat{\mathbf{Y}}_{\mathrm{base}}\parallel \mathbf{R}]
\right),
\qquad
\widehat{\mathbf{Y}}
=
\widehat{\mathbf{Y}}_{\mathrm{base}}
+
\Delta\mathbf{Y}_{\mathrm{res}}.
\end{equation}

\paragraph{Historical evidence construction.}

The \textbf{w/ Naive Prototype} variant consistently degrades performance, showing that superficially similar historical responses are insufficient for reliable dynamic transfer.
The homology-aware construction and temporal anchoring in GRAFT are therefore important for identifying transferable historical evidence.

The \textbf{w/ Residualized Memory} variant further examines what historical memory should represent.
Unlike complete responses, which record what actually followed a historical context, residualized values encode what a particular backbone failed to predict.
They therefore mix potentially transferable local dynamics with backbone-specific approximation errors and structural bias.
The complete-response design is generally more reliable across the evaluated settings, while residualized memory remains competitive in some cases.
This supports preserving predictor-independent historical evidence in memory and postponing the identification of transferable dynamics until grafting time, where the evidence can be conditioned on the current context and backbone state.

\paragraph{Confidence-aware release.}

Removing the confidence-aware gate degrades performance across the reported datasets.
This confirms that retrieved dynamics should be selectively released according to their compatibility with the current forecast rather than uniformly injected into the backbone prediction.

\paragraph{Restoration-cooperative optimization.}

The \textbf{w/o Multi-Objective} variant also performs consistently worse than full GRAFT.
The additional restoration objectives therefore complement point-wise forecasting supervision by explicitly guiding the grafting path toward local dynamic recovery.

\paragraph{Residual correction control.}

We further include \textbf{w/ Residual MLP} as a direct control for generic residual correction.
With the same TimeMixer backbone and retrieved prototype, this variant replaces GRAFT's controlled grafting mechanism with a learned residual adapter.
GRAFT outperforms this variant in all ten dataset--metric comparisons, while the residual adapter underperforms the original backbone in eight.
This controlled comparison shows that retrieved historical responses alone are insufficient: the gains depend on how historical dynamics are calibrated and selectively grafted onto the backbone rather than being treated as generic residual corrections.

Importantly, \textbf{w/ Residualized Memory} and \textbf{w/ Residual MLP} isolate two distinct design choices:
the former tests \emph{what historical evidence is stored}, whereas the latter tests \emph{how retrieved evidence is transferred}.
Together, they support GRAFT's decoupled design:
historical memory preserves predictor-independent response evidence, while the grafting mechanism determines which compatible dynamics should complement the structural backbone.

\subsection{Hyperparameter Sensitivity Analysis}
\label{app:hyperparameter_sensitivity}

GRAFT follows the decoupled principle of \emph{Predict Backbone, Graft Dynamics} and introduces only a small number of additional hyperparameters beyond the fixed forecasting backbone.
Among them, two factors are most relevant to optimization stability and retrieval quality: the training batch size and the top-$K$ candidate number used in historical retrieval.
The batch size controls the stochasticity of optimization for the grafting branch, while top-$K$ determines how many historical response candidates are aggregated to construct the dynamic prototype.
We therefore conduct sensitivity experiments on ETTh1 and ETTh2 using TimeMixer as the backbone.

\begin{figure}[!htbp]
\centering

\begin{subfigure}{0.90\textwidth}
    \centering
    \includegraphics[width=\linewidth]{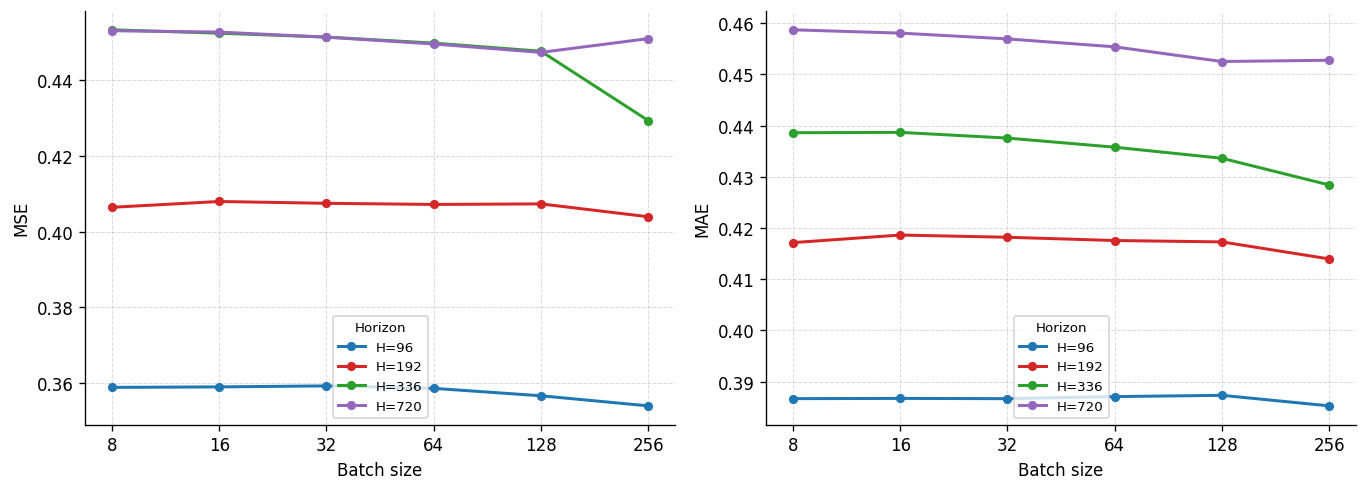}
    \caption{Batch size on ETTh1}
\end{subfigure}

\vspace{0.6em}

\begin{subfigure}{0.90\textwidth}
    \centering
    \includegraphics[width=\linewidth]{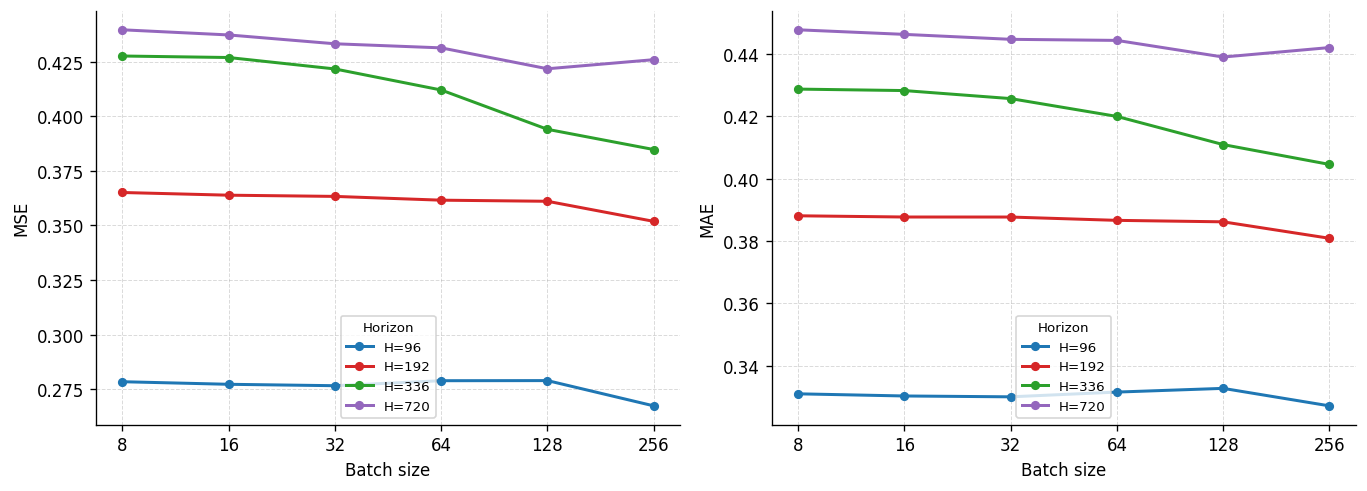}
    \caption{Batch size on ETTh2}
\end{subfigure}

\vspace{0.6em}

\begin{subfigure}{0.90\textwidth}
    \centering
    \includegraphics[width=\linewidth]{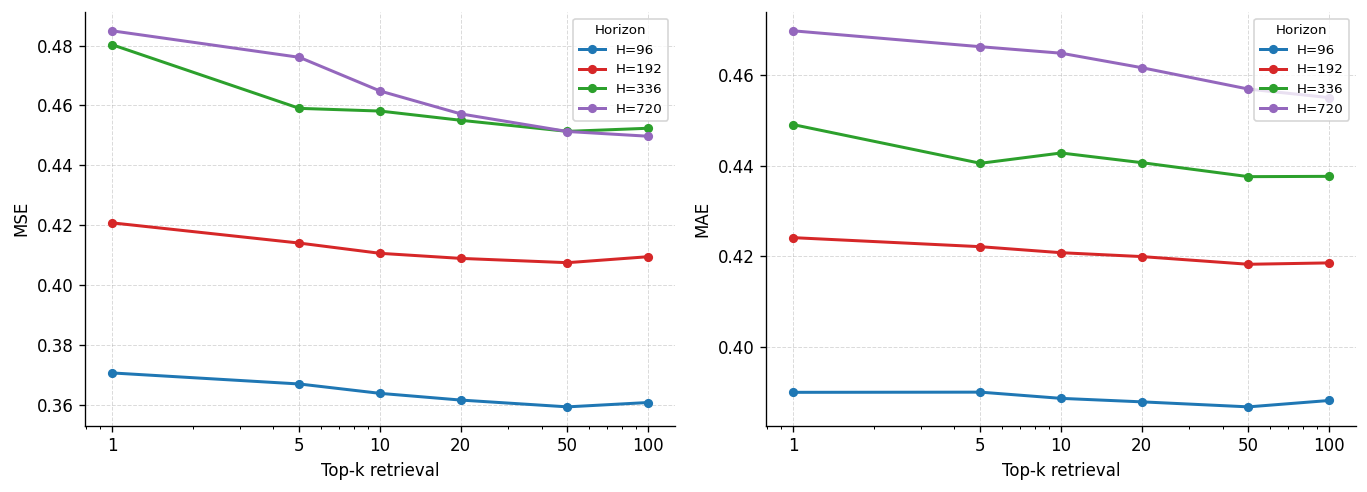}
    \caption{Top-$K$ candidates on ETTh1}
\end{subfigure}

\vspace{0.6em}

\begin{subfigure}{0.90\textwidth}
    \centering
    \includegraphics[width=\linewidth]{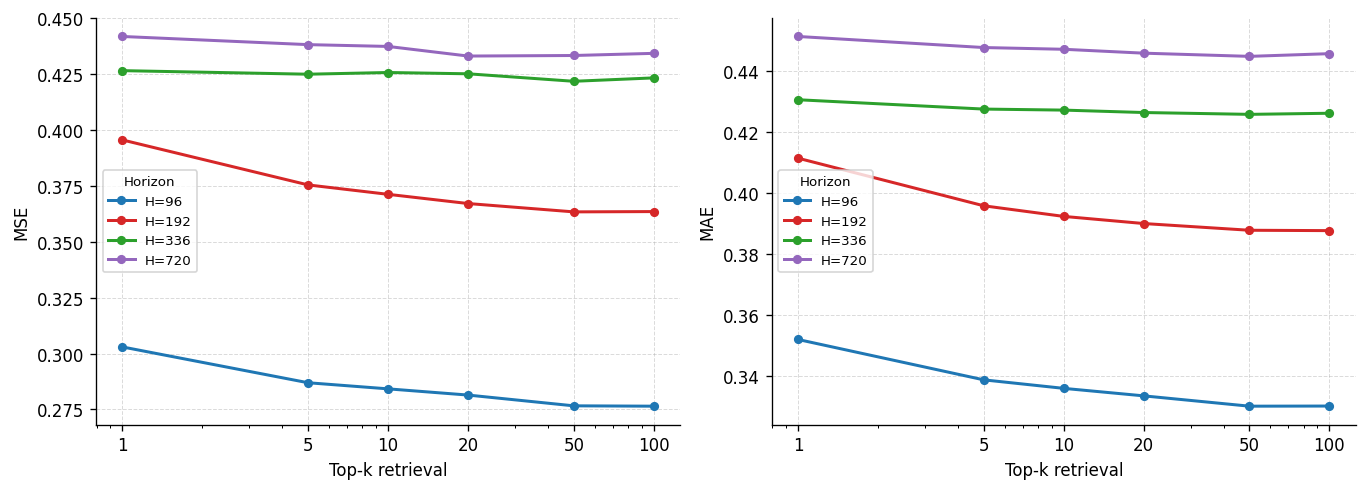}
    \caption{Top-$K$ candidates on ETTh2}
\end{subfigure}

\caption{
Sensitivity analysis of GRAFT on the TimeMixer backbone.
Panels (a)--(b) vary the training batch size, while panels (c)--(d) vary the top-$K$ candidate number used in historical retrieval.
GRAFT remains generally stable across a wide range of batch sizes, suggesting that its gains are not caused by a fragile optimization setting.
Moderate top-$K$ retrieval usually improves over top-1 retrieval, showing the benefit of aggregating multiple historical response candidates.
Lower MSE and MAE indicate better performance.
}
\label{fig:hyperparameter_sensitivity}
\end{figure}

Figure~\ref{fig:hyperparameter_sensitivity} first evaluates the effect of training batch size.
Across ETTh1 and ETTh2, the MSE and MAE curves vary smoothly as the batch size changes from $8$ to $256$.
This indicates that GRAFT is not highly sensitive to a narrow batch-size choice, which is desirable for a post-hoc enhancement module.
In several settings, larger batch sizes yield slightly better results, especially at medium and long horizons.
This suggests that more stable mini-batch optimization can help the grafting branch learn more reliable correction behavior.
However, the trend is not strictly monotonic across all horizons, implying that batch size mainly affects optimization stability rather than changing the core retrieval-and-grafting mechanism.

The top-$K$ sensitivity shows a more direct connection to the design of GRAFT.
When top-$K$ is set to 1, performance is often worse, especially at longer horizons.
This suggests that a single nearest neighbor can be noisy, accidentally matched, or insufficient to represent reusable future dynamics.
Increasing top-$K$ generally improves both MSE and MAE, because aggregating multiple historically similar responses provides a more stable dynamic prototype.
This result supports our motivation that local correction should be grounded in reusable historical response evidence rather than in an isolated nearest neighbor.

At the same time, the best top-$K$ value is dataset- and horizon-dependent.
On ETTh1, performance generally improves as $K$ increases and tends to saturate at a moderate-to-large top-$K$ value.
On ETTh2, the improvement from very small $K$ to larger $K$ is also clear, but the curves become flatter after a certain point.
This indicates a trade-off between evidence diversity and retrieval purity.
A larger top-$K$ can provide more response evidence and reduce the instability of top-1 matching, but excessively increasing $K$ may introduce less compatible candidates and incur additional memory and computational costs.
Therefore, $K$ should be large enough to stabilize the dynamic prototype, but not so large that weakly related historical responses dilute the retrieved evidence.

Based on these observations, we adopt a conservative top-$K$ strategy in the main experiments and keep $K$ within $50$.
This choice balances evidence sufficiency, retrieval purity, computational efficiency, and memory usage.

Overall, the sensitivity results support two conclusions.
First, GRAFT is robust to a broad range of training batch sizes, indicating that its improvements are not driven by fragile hyperparameter tuning.
Second, top-$K$ matters because it controls the trade-off between historical evidence diversity and retrieval purity.
Moderate top-$K$ retrieval is usually preferable to top-1 retrieval, while excessively enlarging the candidate pool can lead to diminishing returns.

\subsection{Additional Dynamic Recovery Discussion}
\label{app:dynamic_recovery_details}

GRAFT is designed to preserve the backbone trajectory while providing controlled dynamic compensation from historically grounded local responses.
To further understand its empirical behavior, we analyze where the net error reduction comes from.
Specifically, we ask whether the improvement after applying GRAFT is uniformly distributed over all forecast points, or whether it is concentrated in regions with stronger local temporal variation.
This analysis is diagnostic: it stratifies forecast errors by ground-truth local dynamic intensity and examines where the net improvement appears.
It does not directly measure gate activation or retrieval reliability.

\paragraph{Local-dynamics conditional analysis.}
Forecasting datasets are evaluated using a sliding-window protocol.
Let \(n\in\{1,\ldots,N_{\mathrm{test}}\}\) index the sliding-window test samples, where each sample contains a look-back window and its corresponding \(H\)-step future target.
For the \(n\)-th test sample, let \(y_{n,t}^{(c)}\) denote the ground-truth value of the \(c\)-th channel at the \(t\)-th position of the future prediction horizon.

For each test sample, forecast position, and channel, we measure point-wise local dynamic intensity using the absolute second-order difference of the ground-truth future sequence.
For each valid interior position \(t=2,\ldots,H-1\), we define
\begin{equation}
    \kappa_{n,t}^{(c)}
    =
    \left|
    y_{n,t+1}^{(c)}
    -
    2y_{n,t}^{(c)}
    +
    y_{n,t-1}^{(c)}
    \right|.
\end{equation}
This quantity measures the local curvature of the target trajectory.
Small values correspond to smooth regions, while large values usually indicate peaks, valleys, turning points, or abrupt local fluctuations.

We sort all evaluated forecast points by \(\kappa_{n,t}^{(c)}\) and split them into five equal-sized quantile groups, denoted by Q1,\(\ldots\),Q5.
Q1 contains the smoothest points, while Q5 contains the strongest local dynamic regions.
This is the same Q1--Q5 notation used in Table~\ref{tab:dynamic_recovery_model_dataset}.

For each dynamic group \(Q_r\), we compute the group-wise MSE reduction as
\begin{equation}
    \Delta_{\mathrm{MSE}}(Q_r)
    =
    \mathrm{MSE}_{\mathrm{Backbone}}(Q_r)
    -
    \mathrm{MSE}_{\mathrm{Backbone+GRAFT}}(Q_r).
\end{equation}
Specifically,
\begin{equation}
    \mathrm{MSE}_{\mathrm{Backbone}}(Q_r)
    =
    \frac{1}{|Q_r|}
    \sum_{(n,t,c)\in Q_r}
    \left(
    \widehat{y}_{\mathrm{base},n,t}^{(c)}
    -
    y_{n,t}^{(c)}
    \right)^2,
\end{equation}
and
\begin{equation}
    \mathrm{MSE}_{\mathrm{Backbone+GRAFT}}(Q_r)
    =
    \frac{1}{|Q_r|}
    \sum_{(n,t,c)\in Q_r}
    \left(
    \widehat{y}_{n,t}^{(c)}
    -
    y_{n,t}^{(c)}
    \right)^2.
\end{equation}
Here, \(\widehat{y}_{\mathrm{base},n,t}^{(c)}\) is the backbone prediction, while \(\widehat{y}_{n,t}^{(c)}\) is the final prediction after grafting.
A positive \(\Delta_{\mathrm{MSE}}(Q_r)\) means that GRAFT reduces the prediction error in the \(r\)-th dynamic group, while a negative value means that the correction increases the error in that group.

We define the signed relative contribution share of \(Q_r\) as
\begin{equation}
    \mathrm{Share}(Q_r)
    =
    \frac{
    \Delta_{\mathrm{MSE}}(Q_r)
    }{
    \sum_{u=1}^{5}
    \Delta_{\mathrm{MSE}}(Q_u)
    }.
\end{equation}
Since the five quantile groups have equal size, the denominator is proportional to the total group-level MSE reduction over all dynamic groups.
This share is signed and can be negative when GRAFT worsens a specific group.
When positive and negative reductions coexist, the share reflects contribution to the net improvement and may be affected by cancellation across groups.
Therefore, we interpret the relative contribution share together with the raw group-wise reduction \(\Delta_{\mathrm{MSE}}(Q_r)\), rather than treating it as a standalone performance metric.

\paragraph{Results.}
Table~\ref{tab:dynamic_recovery_model_dataset} shows that GRAFT's net error reduction
\footnote{``Net error reduction'' denotes the realized decrease in forecasting error after applying GRAFT, computed as \(\mathrm{MSE}_{\mathrm{Backbone}}-\mathrm{MSE}_{\mathrm{Backbone+GRAFT}}\) on a given group of forecast points.
}
is mainly concentrated in medium-to-high dynamic regions, rather than being uniformly distributed over the entire prediction horizon.
For TimesNet, the largest group-wise \(\Delta_{\mathrm{MSE}}\) occurs in Q5 across all five evaluated datasets.
For example, Q5 accounts for 62.7\% of the net reduction on ETTh1 and remains the largest contributor across all five datasets.
This indicates that, for TimesNet, the largest net error reductions are consistently observed in regions with the strongest local dynamic intensity.

For TimeMixer, the gains generally shift away from the smoothest group Q1 and are distributed over Q2--Q5, with the largest or near-largest contributions usually appearing in Q3--Q5.
This pattern should be interpreted together with the structural properties of the ETT datasets.
ETT series contain strong periodic components, such as daily and weekly operating patterns, while also exhibiting local deviations caused by load changes, transient fluctuations, and short-term regime variations.
Therefore, their medium-to-high dynamic regions are more likely to include recoverable local changes than purely random noise, making them suitable cases for evaluating whether GRAFT can complement a strong backbone.

On ETTh1, Q5 alone accounts for 41.0\% of the total group-level MSE reduction, showing that the largest net benefit appears in the most dynamic group.
On ETTh2, the improvement is concentrated in Q4--Q5, with both groups contributing 30.1\%.
On ETTm1, Q1 contributes almost nothing, while the reductions are distributed across Q2--Q5.
These observations suggest that, for TimeMixer, GRAFT's additional benefit is limited in the smoothest regions where the backbone is already relatively reliable.
Instead, the net gains mainly appear in regions with stronger local variation, where the backbone is more likely to produce overly smooth predictions.
This pattern is consistent with the intended role of GRAFT as a post-hoc dynamic compensation module rather than a uniform error-correction layer.

Weather exhibits a less concentrated pattern than the ETT datasets.
For TimeMixer on Weather, Q2 has a negative contribution share, while the positive gains are mainly concentrated in Q3--Q5.
For TimesNet on Weather, all groups are positive, but the contribution is less sharply concentrated in Q5 than on ETTh1 or ETTh2.
This pattern is consistent with the view that Weather contains more complex and less reproducible local variations.
Some high-intensity changes may correspond to noisy or exogenous events rather than reusable historical patterns.
Therefore, GRAFT can still recover part of the local dynamics, but the gain is less cleanly concentrated in the most abrupt group.

\paragraph{Relation to Abruptness Burden.}

The second-order difference used in this analysis is related to, but different from, the Abruptness Burden introduced in Appendix~\ref{app:dataset_profiles}.

Abruptness Burden is defined as the kurtosis of dataset-level second-order differences:

\[
    S_{\mathrm{abrupt}}
    =
    \mathrm{Kurt}\big(\{d_{t}^{(2,c)}\}_{t,c}\big),
    \qquad
    d_{t}^{(2,c)}
    =
    x_{t+1}^{(c)}
    -
    2x_t^{(c)}
    +
    x_{t-1}^{(c)}.
\]

Here, \(\{d_t^{(2,c)}\}_{t,c}\) denotes the collection of second-order differences computed over all valid time steps and channels in the dataset.
Thus, \(S_{\mathrm{abrupt}}\) measures how heavy-tailed the distribution of local curvature changes is over an entire dataset.
A moderate value may indicate sharp but potentially repeatable transient dynamics, whereas an extremely large value may indicate outliers, noisy shocks, or unobserved exogenous events that are difficult to recover from historical memory.
In contrast, \(\kappa_{n,t}^{(c)}\) is a point-level diagnostic computed on the ground-truth horizon of each evaluated forecast sample.
It uses the magnitude of the local second-order difference to stratify forecast points into Q1--Q5 within a fixed dataset and backbone setting.

Thus, \(S_{\mathrm{abrupt}}\) characterizes the dataset-level distribution of abruptness, while \(\kappa_{n,t}^{(c)}\)-based grouping identifies where GRAFT's net error reduction occurs after prediction.
This distinction is important because abruptness does not necessarily imply recoverability.
A dataset with high \(S_{\mathrm{abrupt}}\) may contain many abrupt movements, but some of them may be caused by noise, outliers, measurement errors, or exogenous shocks rather than recurrent temporal patterns.
Such events can produce large point-wise \(\kappa_{n,t}^{(c)}\), but they are difficult to recover from historical evidence because they do not correspond to reusable response patterns.
Therefore, high Abruptness Burden does not contradict the local dynamic recovery behavior observed in the Q1--Q5 analysis.
It only indicates that high-dynamic regions may contain a larger fraction of non-reproducible events.
GRAFT is designed to recover the subset of local dynamics that are historically recurrent and phase-compatible, rather than arbitrary abrupt changes.
\paragraph{Summary.}
Overall, the dynamic recovery analysis shows that GRAFT's net improvement is not uniformly distributed over all forecast points.
Instead, the improvement is mainly concentrated in medium-to-high dynamic regions, where the backbone forecast is more likely to be overly smooth.
This result should be interpreted as diagnostic evidence: it shows where the net benefit of applying GRAFT appears, rather than directly proving larger gate activations or correction magnitudes in those regions.
Nevertheless, the observed pattern is consistent with the intended role of GRAFT as a post-hoc dynamic compensation module.
The backbone provides a stable structural trajectory, while the grafting branch brings additional benefit mainly in regions where sparse local dynamics are harder for the backbone to recover and may benefit from historical response evidence.

\subsection{Full Horizon-wise Forecasting Results}
\label{app:full_results}

\begin{table}[t]
\centering
\scriptsize
\setlength{\tabcolsep}{2.6pt}
\renewcommand{\arraystretch}{1.05}
\caption{Full forecasting results for PMDformer, Dynamic-TMoE, XLinear, and PatchTST.
For each backbone, we compare the original model and its GRAFT-enhanced counterpart.
Values are reported as mean $\pm$ standard deviation over repeated runs.
Lower MSE and MAE indicate better performance. Boldface is determined before rounding.}
\label{tab:full-results-a}
\resizebox{\textwidth}{!}{%
\begin{tabular}{lc|cccc|cccc|cccc|cccc}
\toprule
\multirow{2}{*}{Dataset} & \multirow{2}{*}{Pred.} & \multicolumn{2}{c}{PMDformer} & \multicolumn{2}{c}{+GRAFT} & \multicolumn{2}{c}{Dynamic-TMoE} & \multicolumn{2}{c}{+GRAFT} & \multicolumn{2}{c}{XLinear} & \multicolumn{2}{c}{+GRAFT} & \multicolumn{2}{c}{PatchTST} & \multicolumn{2}{c}{+GRAFT} \\
\cmidrule(lr){3-4}\cmidrule(lr){5-6}\cmidrule(lr){7-8}\cmidrule(lr){9-10}\cmidrule(lr){11-12}\cmidrule(lr){13-14}\cmidrule(lr){15-16}\cmidrule(lr){17-18}
 & & MSE & MAE & MSE & MAE & MSE & MAE & MSE & MAE & MSE & MAE & MSE & MAE & MSE & MAE & MSE & MAE \\
\midrule
\multirow{4}{*}{\rotatebox{90}{ETTh1}}
 & 96 & 0.354{\scriptsize $\pm$0.001} & 0.389{\scriptsize $\pm$0.001} & \textbf{0.351}{\scriptsize $\pm$0.001} & \textbf{0.386}{\scriptsize $\pm$0.001} & 0.370{\scriptsize $\pm$0.001} & 0.393{\scriptsize $\pm$0.001} & \textbf{0.359}{\scriptsize $\pm$0.002} & \textbf{0.388}{\scriptsize $\pm$0.001} & 0.369{\scriptsize $\pm$0.001} & 0.393{\scriptsize $\pm$0.002} & \textbf{0.360}{\scriptsize $\pm$0.001} & \textbf{0.389}{\scriptsize $\pm$0.001} & 0.409{\scriptsize $\pm$0.001} & 0.415{\scriptsize $\pm$0.002} & \textbf{0.369}{\scriptsize $\pm$0.001} & \textbf{0.395}{\scriptsize $\pm$0.002} \\
 & 192 & 0.398{\scriptsize $\pm$0.002} & 0.416{\scriptsize $\pm$0.001} & \textbf{0.389}{\scriptsize $\pm$0.001} & \textbf{0.415}{\scriptsize $\pm$0.001} & 0.415{\scriptsize $\pm$0.002} & \textbf{0.420}{\scriptsize $\pm$0.001} & \textbf{0.415}{\scriptsize $\pm$0.001} & 0.421{\scriptsize $\pm$0.001} & 0.421{\scriptsize $\pm$0.001} & 0.423{\scriptsize $\pm$0.001} & \textbf{0.403}{\scriptsize $\pm$0.001} & \textbf{0.416}{\scriptsize $\pm$0.003} & 0.457{\scriptsize $\pm$0.002} & 0.443{\scriptsize $\pm$0.001} & \textbf{0.413}{\scriptsize $\pm$0.001} & \textbf{0.422}{\scriptsize $\pm$0.002} \\
 & 336 & 0.411{\scriptsize $\pm$0.002} & 0.425{\scriptsize $\pm$0.000} & \textbf{0.400}{\scriptsize $\pm$0.001} & \textbf{0.423}{\scriptsize $\pm$0.002} & 0.460{\scriptsize $\pm$0.002} & 0.446{\scriptsize $\pm$0.001} & \textbf{0.446}{\scriptsize $\pm$0.001} & \textbf{0.438}{\scriptsize $\pm$0.002} & 0.455{\scriptsize $\pm$0.003} & 0.440{\scriptsize $\pm$0.001} & \textbf{0.443}{\scriptsize $\pm$0.001} & \textbf{0.437}{\scriptsize $\pm$0.002} & 0.498{\scriptsize $\pm$0.001} & 0.464{\scriptsize $\pm$0.002} & \textbf{0.454}{\scriptsize $\pm$0.001} & \textbf{0.446}{\scriptsize $\pm$0.001} \\
 & 720 & 0.425{\scriptsize $\pm$0.002} & 0.450{\scriptsize $\pm$0.003} & \textbf{0.424}{\scriptsize $\pm$0.002} & \textbf{0.450}{\scriptsize $\pm$0.002} & 0.480{\scriptsize $\pm$0.003} & 0.469{\scriptsize $\pm$0.003} & \textbf{0.459}{\scriptsize $\pm$0.002} & \textbf{0.461}{\scriptsize $\pm$0.003} & 0.453{\scriptsize $\pm$0.003} & \textbf{0.456}{\scriptsize $\pm$0.002} & \textbf{0.446}{\scriptsize $\pm$0.002} & 0.456{\scriptsize $\pm$0.001} & 0.496{\scriptsize $\pm$0.002} & 0.485{\scriptsize $\pm$0.001} & \textbf{0.467}{\scriptsize $\pm$0.001} & \textbf{0.473}{\scriptsize $\pm$0.002} \\
\midrule
\multirow{4}{*}{\rotatebox{90}{ETTh2}}
 & 96 & 0.270{\scriptsize $\pm$0.001} & 0.329{\scriptsize $\pm$0.001} & \textbf{0.267}{\scriptsize $\pm$0.001} & \textbf{0.328}{\scriptsize $\pm$0.000} & 0.287{\scriptsize $\pm$0.001} & 0.339{\scriptsize $\pm$0.001} & \textbf{0.276}{\scriptsize $\pm$0.001} & \textbf{0.328}{\scriptsize $\pm$0.002} & 0.286{\scriptsize $\pm$0.001} & 0.337{\scriptsize $\pm$0.001} & \textbf{0.274}{\scriptsize $\pm$0.001} & \textbf{0.326}{\scriptsize $\pm$0.001} & 0.299{\scriptsize $\pm$0.001} & 0.347{\scriptsize $\pm$0.002} & \textbf{0.277}{\scriptsize $\pm$0.001} & \textbf{0.329}{\scriptsize $\pm$0.001} \\
 & 192 & 0.334{\scriptsize $\pm$0.001} & \textbf{0.374}{\scriptsize $\pm$0.000} & \textbf{0.328}{\scriptsize $\pm$0.002} & 0.374{\scriptsize $\pm$0.001} & 0.363{\scriptsize $\pm$0.001} & 0.393{\scriptsize $\pm$0.001} & \textbf{0.353}{\scriptsize $\pm$0.002} & \textbf{0.382}{\scriptsize $\pm$0.002} & 0.363{\scriptsize $\pm$0.001} & 0.388{\scriptsize $\pm$0.001} & \textbf{0.354}{\scriptsize $\pm$0.001} & \textbf{0.381}{\scriptsize $\pm$0.002} & 0.383{\scriptsize $\pm$0.002} & 0.397{\scriptsize $\pm$0.001} & \textbf{0.359}{\scriptsize $\pm$0.001} & \textbf{0.382}{\scriptsize $\pm$0.002} \\
 & 336 & 0.352{\scriptsize $\pm$0.002} & 0.394{\scriptsize $\pm$0.001} & \textbf{0.343}{\scriptsize $\pm$0.001} & \textbf{0.393}{\scriptsize $\pm$0.002} & 0.406{\scriptsize $\pm$0.002} & 0.425{\scriptsize $\pm$0.002} & \textbf{0.404}{\scriptsize $\pm$0.001} & \textbf{0.419}{\scriptsize $\pm$0.001} & 0.378{\scriptsize $\pm$0.001} & \textbf{0.407}{\scriptsize $\pm$0.002} & \textbf{0.377}{\scriptsize $\pm$0.001} & 0.408{\scriptsize $\pm$0.002} & 0.418{\scriptsize $\pm$0.001} & 0.429{\scriptsize $\pm$0.002} & \textbf{0.400}{\scriptsize $\pm$0.001} & \textbf{0.415}{\scriptsize $\pm$0.001} \\
 & 720 & 0.388{\scriptsize $\pm$0.002} & \textbf{0.428}{\scriptsize $\pm$0.002} & \textbf{0.385}{\scriptsize $\pm$0.003} & 0.428{\scriptsize $\pm$0.003} & 0.425{\scriptsize $\pm$0.002} & 0.444{\scriptsize $\pm$0.002} & \textbf{0.416}{\scriptsize $\pm$0.001} & \textbf{0.438}{\scriptsize $\pm$0.003} & 0.408{\scriptsize $\pm$0.002} & 0.431{\scriptsize $\pm$0.002} & \textbf{0.406}{\scriptsize $\pm$0.002} & \textbf{0.429}{\scriptsize $\pm$0.003} & 0.426{\scriptsize $\pm$0.002} & 0.443{\scriptsize $\pm$0.001} & \textbf{0.407}{\scriptsize $\pm$0.002} & \textbf{0.430}{\scriptsize $\pm$0.001} \\
\midrule
\multirow{4}{*}{\rotatebox{90}{ETTm1}}
 & 96 & 0.280{\scriptsize $\pm$0.001} & 0.330{\scriptsize $\pm$0.001} & \textbf{0.277}{\scriptsize $\pm$0.002} & \textbf{0.329}{\scriptsize $\pm$0.002} & 0.315{\scriptsize $\pm$0.000} & 0.353{\scriptsize $\pm$0.002} & \textbf{0.304}{\scriptsize $\pm$0.000} & \textbf{0.343}{\scriptsize $\pm$0.002} & 0.311{\scriptsize $\pm$0.001} & 0.351{\scriptsize $\pm$0.001} & \textbf{0.302}{\scriptsize $\pm$0.001} & \textbf{0.344}{\scriptsize $\pm$0.001} & 0.352{\scriptsize $\pm$0.002} & 0.383{\scriptsize $\pm$0.001} & \textbf{0.318}{\scriptsize $\pm$0.002} & \textbf{0.355}{\scriptsize $\pm$0.003} \\
 & 192 & 0.326{\scriptsize $\pm$0.001} & 0.359{\scriptsize $\pm$0.002} & \textbf{0.315}{\scriptsize $\pm$0.003} & \textbf{0.359}{\scriptsize $\pm$0.001} & 0.358{\scriptsize $\pm$0.002} & 0.379{\scriptsize $\pm$0.002} & \textbf{0.350}{\scriptsize $\pm$0.003} & \textbf{0.371}{\scriptsize $\pm$0.002} & 0.353{\scriptsize $\pm$0.001} & 0.376{\scriptsize $\pm$0.001} & \textbf{0.346}{\scriptsize $\pm$0.000} & \textbf{0.370}{\scriptsize $\pm$0.001} & 0.376{\scriptsize $\pm$0.003} & 0.395{\scriptsize $\pm$0.002} & \textbf{0.362}{\scriptsize $\pm$0.001} & \textbf{0.379}{\scriptsize $\pm$0.002} \\
 & 336 & 0.361{\scriptsize $\pm$0.002} & \textbf{0.385}{\scriptsize $\pm$0.002} & \textbf{0.357}{\scriptsize $\pm$0.002} & 0.385{\scriptsize $\pm$0.001} & 0.390{\scriptsize $\pm$0.002} & 0.403{\scriptsize $\pm$0.002} & \textbf{0.387}{\scriptsize $\pm$0.003} & \textbf{0.400}{\scriptsize $\pm$0.001} & 0.382{\scriptsize $\pm$0.002} & 0.398{\scriptsize $\pm$0.002} & \textbf{0.375}{\scriptsize $\pm$0.0012} & \textbf{0.393}{\scriptsize $\pm$0.002} & 0.408{\scriptsize $\pm$0.002} & 0.416{\scriptsize $\pm$0.001} & \textbf{0.393}{\scriptsize $\pm$0.003} & \textbf{0.400}{\scriptsize $\pm$0.002} \\
 & 720 & 0.421{\scriptsize $\pm$0.003} & 0.416{\scriptsize $\pm$0.002} & \textbf{0.402}{\scriptsize $\pm$0.003} & \textbf{0.413}{\scriptsize $\pm$0.002} & 0.451{\scriptsize $\pm$0.003} & 0.442{\scriptsize $\pm$0.001} & \textbf{0.444}{\scriptsize $\pm$0.002} & \textbf{0.437}{\scriptsize $\pm$0.001} & 0.444{\scriptsize $\pm$0.002} & 0.436{\scriptsize $\pm$0.003} & \textbf{0.436}{\scriptsize $\pm$0.002} & \textbf{0.432}{\scriptsize $\pm$0.003} & 0.465{\scriptsize $\pm$0.001} & 0.445{\scriptsize $\pm$0.003} & \textbf{0.452}{\scriptsize $\pm$0.002} & \textbf{0.437}{\scriptsize $\pm$0.001} \\
\midrule
\multirow{4}{*}{\rotatebox{90}{ETTm2}}
 & 96 & \textbf{0.156}{\scriptsize $\pm$0.002} & \textbf{0.241}{\scriptsize $\pm$0.002} & 0.156{\scriptsize $\pm$0.001} & 0.241{\scriptsize $\pm$0.000} & 0.168{\scriptsize $\pm$0.000} & 0.253{\scriptsize $\pm$0.001} & \textbf{0.167}{\scriptsize $\pm$0.001} & \textbf{0.251}{\scriptsize $\pm$0.000} & 0.167{\scriptsize $\pm$0.001} & 0.250{\scriptsize $\pm$0.001} & \textbf{0.166}{\scriptsize $\pm$0.001} & \textbf{0.250}{\scriptsize $\pm$0.001} & 0.177{\scriptsize $\pm$0.000} & 0.262{\scriptsize $\pm$0.001} & \textbf{0.175}{\scriptsize $\pm$0.000} & \textbf{0.258}{\scriptsize $\pm$0.001} \\
 & 192 & 0.214{\scriptsize $\pm$0.001} & \textbf{0.282}{\scriptsize $\pm$0.001} & \textbf{0.211}{\scriptsize $\pm$0.001} & 0.284{\scriptsize $\pm$0.001} & 0.237{\scriptsize $\pm$0.000} & 0.299{\scriptsize $\pm$0.001} & \textbf{0.234}{\scriptsize $\pm$0.001} & \textbf{0.296}{\scriptsize $\pm$0.001} & 0.233{\scriptsize $\pm$0.001} & 0.293{\scriptsize $\pm$0.001} & \textbf{0.231}{\scriptsize $\pm$0.001} & \textbf{0.293}{\scriptsize $\pm$0.000} & 0.242{\scriptsize $\pm$0.001} & 0.305{\scriptsize $\pm$0.000} & \textbf{0.240}{\scriptsize $\pm$0.001} & \textbf{0.300}{\scriptsize $\pm$0.000} \\
 & 336 & 0.267{\scriptsize $\pm$0.001} & \textbf{0.320}{\scriptsize $\pm$0.001} & \textbf{0.263}{\scriptsize $\pm$0.002} & 0.322{\scriptsize $\pm$0.001} & \textbf{0.291}{\scriptsize $\pm$0.001} & \textbf{0.335}{\scriptsize $\pm$0.001} & 0.293{\scriptsize $\pm$0.002} & 0.338{\scriptsize $\pm$0.001} & \textbf{0.290}{\scriptsize $\pm$0.001} & \textbf{0.332}{\scriptsize $\pm$0.001} & 0.292{\scriptsize $\pm$0.001} & 0.332{\scriptsize $\pm$0.002} & \textbf{0.303}{\scriptsize $\pm$0.000} & \textbf{0.345}{\scriptsize $\pm$0.001} & 0.305{\scriptsize $\pm$0.000} & 0.349{\scriptsize $\pm$0.001} \\
 & 720 & 0.347{\scriptsize $\pm$0.002} & 0.376{\scriptsize $\pm$0.001} & \textbf{0.347}{\scriptsize $\pm$0.003} & \textbf{0.376}{\scriptsize $\pm$0.001} & \textbf{0.393}{\scriptsize $\pm$0.003} & \textbf{0.396}{\scriptsize $\pm$0.003} & 0.395{\scriptsize $\pm$0.002} & 0.399{\scriptsize $\pm$0.002} & \textbf{0.388}{\scriptsize $\pm$0.003} & \textbf{0.390}{\scriptsize $\pm$0.002} & 0.390{\scriptsize $\pm$0.003} & 0.392{\scriptsize $\pm$0.002} & \textbf{0.400}{\scriptsize $\pm$0.001} & \textbf{0.401}{\scriptsize $\pm$0.000} & 0.404{\scriptsize $\pm$0.001} & 0.402{\scriptsize $\pm$0.000} \\
\midrule
\multirow{4}{*}{\rotatebox{90}{Weather}}
 & 96 & 0.141{\scriptsize $\pm$0.002} & \textbf{0.180}{\scriptsize $\pm$0.003} & \textbf{0.139}{\scriptsize $\pm$0.002} & 0.180{\scriptsize $\pm$0.001} & 0.155{\scriptsize $\pm$0.001} & 0.201{\scriptsize $\pm$0.003} & \textbf{0.152}{\scriptsize $\pm$0.002} & \textbf{0.196}{\scriptsize $\pm$0.002} & 0.149{\scriptsize $\pm$0.002} & 0.198{\scriptsize $\pm$0.003}& \textbf{0.148}{\scriptsize $\pm$0.001} & \textbf{0.195}{\scriptsize $\pm$0.004} & 0.177{\scriptsize $\pm$0.004} & 0.218{\scriptsize $\pm$0.002} & \textbf{0.169}{\scriptsize $\pm$0.003} & \textbf{0.216}{\scriptsize $\pm$0.002} \\
 & 192 & 0.185{\scriptsize $\pm$0.003} & \textbf{0.226}{\scriptsize $\pm$0.003} & \textbf{0.184}{\scriptsize $\pm$0.001} & 0.226{\scriptsize $\pm$0.003} & 0.208{\scriptsize $\pm$0.002} & 0.250{\scriptsize $\pm$0.002} & \textbf{0.208}{\scriptsize $\pm$0.001} & \textbf{0.247}{\scriptsize $\pm$0.003} & 0.201{\scriptsize $\pm$0.004} & 0.244{\scriptsize $\pm$0.003} & \textbf{0.200}{\scriptsize $\pm$0.002} & \textbf{0.243}{\scriptsize $\pm$0.002} & 0.225{\scriptsize $\pm$0.003} & \textbf{0.259}{\scriptsize $\pm$0.004} & \textbf{0.217}{\scriptsize $\pm$0.005} & \textbf{0.259}{\scriptsize $\pm$0.002} \\
 & 336 & 0.237{\scriptsize $\pm$0.001} & \textbf{0.275}{\scriptsize $\pm$0.003} & \textbf{0.236}{\scriptsize $\pm$0.003} & 0.275{\scriptsize $\pm$0.002} & 0.263{\scriptsize $\pm$0.001} & 0.291{\scriptsize $\pm$0.001} & \textbf{0.261}{\scriptsize $\pm$0.002} & \textbf{0.288}{\scriptsize $\pm$0.002} & 0.259{\scriptsize $\pm$0.001} & 0.288{\scriptsize $\pm$0.002} & \textbf{0.258}{\scriptsize $\pm$0.003} & \textbf{0.286}{\scriptsize $\pm$0.003} & 0.279{\scriptsize $\pm$0.005} & \textbf{0.298}{\scriptsize $\pm$0.003} & \textbf{0.274}{\scriptsize $\pm$0.004} & 0.301{\scriptsize $\pm$0.002} \\
 & 720 & 0.305{\scriptsize $\pm$0.004} & \textbf{0.323}{\scriptsize $\pm$0.003} & \textbf{0.303}{\scriptsize $\pm$0.003} & 0.323{\scriptsize $\pm$0.002} & 0.351{\scriptsize $\pm$0.004} & \textbf{0.349}{\scriptsize $\pm$0.003} & \textbf{0.350}{\scriptsize $\pm$0.003} & 0.351{\scriptsize $\pm$0.005} & 0.339{\scriptsize $\pm$0.003} & \textbf{0.340}{\scriptsize $\pm$0.002} & \textbf{0.338}{\scriptsize $\pm$0.004} & 0.341{\scriptsize $\pm$0.003} & \textbf{0.353}{\scriptsize $\pm$0.003} & \textbf{0.347}{\scriptsize $\pm$0.004} & 0.357{\scriptsize $\pm$0.002} & 0.357{\scriptsize $\pm$0.005} \\
\midrule
\multirow{4}{*}{\rotatebox{90}{Electricity}}
 & 96 & 0.122{\scriptsize $\pm$0.000} & 0.214{\scriptsize $\pm$0.001} & \textbf{0.121}{\scriptsize $\pm$0.001} & \textbf{0.212}{\scriptsize $\pm$0.002} & 0.147{\scriptsize $\pm$0.001} & 0.246{\scriptsize $\pm$0.002} & \textbf{0.145}{\scriptsize $\pm$0.001} & \textbf{0.240}{\scriptsize $\pm$0.002} & 0.138{\scriptsize $\pm$0.001} & 0.237{\scriptsize $\pm$0.001} & \textbf{0.136}{\scriptsize $\pm$0.001} & \textbf{0.232}{\scriptsize $\pm$0.002} & 0.193{\scriptsize $\pm$0.002} & 0.283{\scriptsize $\pm$0.003} & \textbf{0.172}{\scriptsize $\pm$0.002} & \textbf{0.259}{\scriptsize $\pm$0.001} \\
 & 192 & 0.140{\scriptsize $\pm$0.001} & 0.231{\scriptsize $\pm$0.002} & \textbf{0.138}{\scriptsize $\pm$0.001} & \textbf{0.229}{\scriptsize $\pm$0.001} & 0.165{\scriptsize $\pm$0.002} & 0.264{\scriptsize $\pm$0.002} & \textbf{0.162}{\scriptsize $\pm$0.001} & \textbf{0.261}{\scriptsize $\pm$0.002} & 0.156{\scriptsize $\pm$0.000} & 0.251{\scriptsize $\pm$0.002} & \textbf{0.152}{\scriptsize $\pm$0.001} & \textbf{0.247}{\scriptsize $\pm$0.001} & 0.197{\scriptsize $\pm$0.003} & 0.289{\scriptsize $\pm$0.002} & \textbf{0.181}{\scriptsize $\pm$0.004} & \textbf{0.269}{\scriptsize $\pm$0.001} \\
 & 336 & 0.152{\scriptsize $\pm$0.002} & \textbf{0.245}{\scriptsize $\pm$0.002} & \textbf{0.150}{\scriptsize $\pm$0.001} & 0.245{\scriptsize $\pm$0.002} & 0.177{\scriptsize $\pm$0.002} & 0.274{\scriptsize $\pm$0.003} & \textbf{0.175}{\scriptsize $\pm$0.002} & \textbf{0.273}{\scriptsize $\pm$0.003} & 0.171{\scriptsize $\pm$0.002} & 0.267{\scriptsize $\pm$0.003} & \textbf{0.170}{\scriptsize $\pm$0.001} & \textbf{0.262}{\scriptsize $\pm$0.002} & 0.213{\scriptsize $\pm$0.004} & 0.304{\scriptsize $\pm$0.002} & \textbf{0.196}{\scriptsize $\pm$0.003} & \textbf{0.284}{\scriptsize $\pm$0.001} \\
 & 720 & 0.177{\scriptsize $\pm$0.002} & 0.272{\scriptsize $\pm$0.003} & \textbf{0.176}{\scriptsize $\pm$0.001} & \textbf{0.271}{\scriptsize $\pm$0.001} & 0.212{\scriptsize $\pm$0.002} & 0.298{\scriptsize $\pm$0.002} & \textbf{0.210}{\scriptsize $\pm$0.002} & \textbf{0.297}{\scriptsize $\pm$0.001} & 0.206{\scriptsize $\pm$0.002} & 0.297{\scriptsize $\pm$0.002} & \textbf{0.205}{\scriptsize $\pm$0.003} & \textbf{0.290}{\scriptsize $\pm$0.002} & \textbf{0.255}{\scriptsize $\pm$0.002} & \textbf{0.336}{\scriptsize $\pm$0.004} & 0.256{\scriptsize $\pm$0.003} & 0.340{\scriptsize $\pm$0.001} \\
\midrule
\multirow{4}{*}{\rotatebox{90}{Traffic}}
 & 96 & 0.338{\scriptsize $\pm$0.003} & 0.212{\scriptsize $\pm$0.004} & \textbf{0.333}{\scriptsize $\pm$0.004} & \textbf{0.210}{\scriptsize $\pm$0.005} & 0.462{\scriptsize $\pm$0.005} & 0.282{\scriptsize $\pm$0.003} & \textbf{0.454}{\scriptsize $\pm$0.003} & \textbf{0.280}{\scriptsize $\pm$0.003} & 0.432{\scriptsize $\pm$0.004} & 0.281{\scriptsize $\pm$0.003} & \textbf{0.411}{\scriptsize $\pm$0.003} & \textbf{0.273}{\scriptsize $\pm$0.003} & 0.470{\scriptsize $\pm$0.005} & 0.307{\scriptsize $\pm$0.004} & \textbf{0.468}{\scriptsize $\pm$0.006} & \textbf{0.304}{\scriptsize $\pm$0.005} \\
 & 192 & 0.367{\scriptsize $\pm$0.004} & 0.227{\scriptsize $\pm$0.006} & \textbf{0.363}{\scriptsize $\pm$0.005} & \textbf{0.225}{\scriptsize $\pm$0.005} & 0.472{\scriptsize $\pm$0.004} & 0.284{\scriptsize $\pm$0.003} & \textbf{0.466}{\scriptsize $\pm$0.005} & \textbf{0.282}{\scriptsize $\pm$0.004} & 0.449{\scriptsize $\pm$0.004} & 0.289{\scriptsize $\pm$0.003} & \textbf{0.438}{\scriptsize $\pm$0.004} & \textbf{0.283}{\scriptsize $\pm$0.003} & 0.477{\scriptsize $\pm$0.006} & \textbf{0.308}{\scriptsize $\pm$0.005} & \textbf{0.474}{\scriptsize $\pm$0.007} & \textbf{0.308}{\scriptsize $\pm$0.004} \\
 & 336 & 0.379{\scriptsize $\pm$0.005} & \textbf{0.235}{\scriptsize $\pm$0.003} & \textbf{0.377}{\scriptsize $\pm$0.004} & 0.236{\scriptsize $\pm$0.003} & 0.478{\scriptsize $\pm$0.006} & 0.295{\scriptsize $\pm$0.002} & \textbf{0.473}{\scriptsize $\pm$0.004} & \textbf{0.295}{\scriptsize $\pm$0.002} & 0.465{\scriptsize $\pm$0.007} & 0.295{\scriptsize $\pm$0.005} & \textbf{0.457}{\scriptsize $\pm$0.004} & \textbf{0.289}{\scriptsize $\pm$0.003} & 0.493{\scriptsize $\pm$0.007} & 0.315{\scriptsize $\pm$0.006} & \textbf{0.483}{\scriptsize $\pm$0.005} & \textbf{0.311}{\scriptsize $\pm$0.007} \\
 & 720 & 0.426{\scriptsize $\pm$0.005} & \textbf{0.262}{\scriptsize $\pm$0.003} & \textbf{0.422}{\scriptsize $\pm$0.004} & 0.264{\scriptsize $\pm$0.003} & 0.515{\scriptsize $\pm$0.007} & 0.303{\scriptsize $\pm$0.005} & \textbf{0.511}{\scriptsize $\pm$0.004} & \textbf{0.302}{\scriptsize $\pm$0.004} & 0.507{\scriptsize $\pm$0.005} & 0.315{\scriptsize $\pm$0.003} & \textbf{0.499}{\scriptsize $\pm$0.006} & \textbf{0.310}{\scriptsize $\pm$0.004} & 0.526{\scriptsize $\pm$0.005} & \textbf{0.333}{\scriptsize $\pm$0.007} & \textbf{0.521}{\scriptsize $\pm$0.006} & 0.334{\scriptsize $\pm$0.004} \\
\bottomrule
\end{tabular}%
}
\end{table}

\begin{table}[t]
\centering
\scriptsize
\setlength{\tabcolsep}{2.6pt}
\renewcommand{\arraystretch}{1.05}
\caption{Full forecasting results for TimeMixer, DLinear, TimesNet, and PDF.
For each backbone, we compare the original model and its GRAFT-enhanced counterpart.
Values are reported as mean $\pm$ standard deviation over repeated runs.
Lower MSE and MAE indicate better performance.}
\label{tab:full-results-b}
\resizebox{\textwidth}{!}{%
\begin{tabular}{lc|cccc|cccc|cccc|cccc}
\toprule
\multirow{2}{*}{Dataset} & \multirow{2}{*}{Pred.} & \multicolumn{2}{c}{TimeMixer} & \multicolumn{2}{c}{+GRAFT} & \multicolumn{2}{c}{DLinear} & \multicolumn{2}{c}{+GRAFT} & \multicolumn{2}{c}{TimesNet} & \multicolumn{2}{c}{+GRAFT} & \multicolumn{2}{c}{PDF} & \multicolumn{2}{c}{+GRAFT} \\
\cmidrule(lr){3-4}\cmidrule(lr){5-6}\cmidrule(lr){7-8}\cmidrule(lr){9-10}\cmidrule(lr){11-12}\cmidrule(lr){13-14}\cmidrule(lr){15-16}\cmidrule(lr){17-18}
 & & MSE & MAE & MSE & MAE & MSE & MAE & MSE & MAE & MSE & MAE & MSE & MAE & MSE & MAE & MSE & MAE \\
\midrule
\multirow{4}{*}{\rotatebox{90}{ETTh1}}
 & 96 & 0.386{\scriptsize $\pm$0.002} & 0.403{\scriptsize $\pm$0.001} & \textbf{0.359}{\scriptsize $\pm$0.001} & \textbf{0.387}{\scriptsize $\pm$0.002} & 0.398{\scriptsize $\pm$0.001} & 0.413{\scriptsize $\pm$0.002} & \textbf{0.363}{\scriptsize $\pm$0.001} & \textbf{0.393}{\scriptsize $\pm$0.002} & 0.393{\scriptsize $\pm$0.002} & 0.414{\scriptsize $\pm$0.001} & \textbf{0.368}{\scriptsize $\pm$0.001} & \textbf{0.395}{\scriptsize $\pm$0.002} & 0.370{\scriptsize $\pm$0.001} & 0.395{\scriptsize $\pm$0.002} & \textbf{0.361}{\scriptsize $\pm$0.001} & \textbf{0.390}{\scriptsize $\pm$0.002} \\
 & 192 & 0.430{\scriptsize $\pm$0.001} & 0.428{\scriptsize $\pm$0.001} & \textbf{0.407}{\scriptsize $\pm$0.002} & \textbf{0.418}{\scriptsize $\pm$0.001} & 0.455{\scriptsize $\pm$0.002} & 0.451{\scriptsize $\pm$0.001} & \textbf{0.413}{\scriptsize $\pm$0.002} & \textbf{0.420}{\scriptsize $\pm$0.001} & 0.450{\scriptsize $\pm$0.001} & 0.447{\scriptsize $\pm$0.002} & \textbf{0.408}{\scriptsize $\pm$0.001} & \textbf{0.423}{\scriptsize $\pm$0.002} & 0.406{\scriptsize $\pm$0.002} & 0.416{\scriptsize $\pm$0.001} & \textbf{0.397}{\scriptsize $\pm$0.002} & \textbf{0.413}{\scriptsize $\pm$0.001} \\
 & 336 & 0.476{\scriptsize $\pm$0.001} & 0.443{\scriptsize $\pm$0.002} & \textbf{0.451}{\scriptsize $\pm$0.001} & \textbf{0.438}{\scriptsize $\pm$0.002} & 0.491{\scriptsize $\pm$0.001} & 0.469{\scriptsize $\pm$0.002} & \textbf{0.449}{\scriptsize $\pm$0.001} & \textbf{0.437}{\scriptsize $\pm$0.002} & 0.498{\scriptsize $\pm$0.002} & 0.471{\scriptsize $\pm$0.001} & \textbf{0.455}{\scriptsize $\pm$0.002} & \textbf{0.447}{\scriptsize $\pm$0.001} & 0.418{\scriptsize $\pm$0.001} & 0.425{\scriptsize $\pm$0.002} & \textbf{0.414}{\scriptsize $\pm$0.001} & \textbf{0.424}{\scriptsize $\pm$0.002} \\
 & 720 & 0.474{\scriptsize $\pm$0.002} & 0.461{\scriptsize $\pm$0.001} & \textbf{0.451}{\scriptsize $\pm$0.002} & \textbf{0.457}{\scriptsize $\pm$0.001} & 0.514{\scriptsize $\pm$0.002} & 0.512{\scriptsize $\pm$0.001} & \textbf{0.453}{\scriptsize $\pm$0.002} & \textbf{0.458}{\scriptsize $\pm$0.001} & 0.510{\scriptsize $\pm$0.001} & 0.489{\scriptsize $\pm$0.002} & \textbf{0.483}{\scriptsize $\pm$0.001} & \textbf{0.479}{\scriptsize $\pm$0.002} & 0.444{\scriptsize $\pm$0.002} & 0.465{\scriptsize $\pm$0.001} & \textbf{0.432}{\scriptsize $\pm$0.002} & \textbf{0.459}{\scriptsize $\pm$0.001} \\
\midrule
\multirow{4}{*}{\rotatebox{90}{ETTh2}}
 & 96 & 0.290{\scriptsize $\pm$0.002} & 0.344{\scriptsize $\pm$0.001} & \textbf{0.278}{\scriptsize $\pm$0.001} & \textbf{0.331}{\scriptsize $\pm$0.002} & 0.352{\scriptsize $\pm$0.002} & 0.403{\scriptsize $\pm$0.001} & \textbf{0.280}{\scriptsize $\pm$0.002} & \textbf{0.330}{\scriptsize $\pm$0.001} & 0.319{\scriptsize $\pm$0.001} & 0.364{\scriptsize $\pm$0.002} & \textbf{0.300}{\scriptsize $\pm$0.001} & \textbf{0.343}{\scriptsize $\pm$0.002} & 0.282{\scriptsize $\pm$0.002} & 0.341{\scriptsize $\pm$0.001} & \textbf{0.277}{\scriptsize $\pm$0.002} & \textbf{0.337}{\scriptsize $\pm$0.001} \\
 & 192 & 0.372{\scriptsize $\pm$0.001} & 0.393{\scriptsize $\pm$0.002} & \textbf{0.363}{\scriptsize $\pm$0.001} & \textbf{0.387}{\scriptsize $\pm$0.002} & 0.482{\scriptsize $\pm$0.001} & 0.479{\scriptsize $\pm$0.002} & \textbf{0.358}{\scriptsize $\pm$0.001} & \textbf{0.381}{\scriptsize $\pm$0.002} & 0.404{\scriptsize $\pm$0.002} & 0.415{\scriptsize $\pm$0.001} & \textbf{0.377}{\scriptsize $\pm$0.002} & \textbf{0.393}{\scriptsize $\pm$0.001} & 0.343{\scriptsize $\pm$0.001} & 0.383{\scriptsize $\pm$0.002} & \textbf{0.340}{\scriptsize $\pm$0.001} & \textbf{0.382}{\scriptsize $\pm$0.002} \\
 & 336 & 0.427{\scriptsize $\pm$0.002} & 0.430{\scriptsize $\pm$0.001} & \textbf{0.425}{\scriptsize $\pm$0.001} & \textbf{0.427}{\scriptsize $\pm$0.002} & 0.597{\scriptsize $\pm$0.002} & 0.543{\scriptsize $\pm$0.001} & \textbf{0.368}{\scriptsize $\pm$0.002} & \textbf{0.396}{\scriptsize $\pm$0.001} & 0.465{\scriptsize $\pm$0.001} & 0.462{\scriptsize $\pm$0.002} & \textbf{0.422}{\scriptsize $\pm$0.001} & \textbf{0.431}{\scriptsize $\pm$0.002} & \textbf{0.331}{\scriptsize $\pm$0.002} & \textbf{0.386}{\scriptsize $\pm$0.001} & 0.342{\scriptsize $\pm$0.002} & 0.390{\scriptsize $\pm$0.001} \\
 & 720 & 0.443{\scriptsize $\pm$0.001} & 0.450{\scriptsize $\pm$0.002} & \textbf{0.434}{\scriptsize $\pm$0.001} & \textbf{0.445}{\scriptsize $\pm$0.001} & 0.821{\scriptsize $\pm$0.001} & 0.653{\scriptsize $\pm$0.002} & \textbf{0.393}{\scriptsize $\pm$0.001} & \textbf{0.420}{\scriptsize $\pm$0.002} & 0.425{\scriptsize $\pm$0.002} & 0.445{\scriptsize $\pm$0.001} & \textbf{0.406}{\scriptsize $\pm$0.002} & \textbf{0.431}{\scriptsize $\pm$0.001} & \textbf{0.384}{\scriptsize $\pm$0.001} & \textbf{0.424}{\scriptsize $\pm$0.002} & 0.387{\scriptsize $\pm$0.001} & \textbf{0.424}{\scriptsize $\pm$0.002} \\
\midrule
\multirow{4}{*}{\rotatebox{90}{ETTm1}}
 & 96 & 0.337{\scriptsize $\pm$0.003} & 0.372{\scriptsize $\pm$0.001} & \textbf{0.307}{\scriptsize $\pm$0.002} & \textbf{0.349}{\scriptsize $\pm$0.001} & 0.345{\scriptsize $\pm$0.002} & 0.372{\scriptsize $\pm$0.003} & \textbf{0.306}{\scriptsize $\pm$0.001} & \textbf{0.345}{\scriptsize $\pm$0.002} & 0.334{\scriptsize $\pm$0.003} & 0.373{\scriptsize $\pm$0.001} & \textbf{0.331}{\scriptsize $\pm$0.002} & \textbf{0.368}{\scriptsize $\pm$0.003} & 0.297{\scriptsize $\pm$0.001} & 0.347{\scriptsize $\pm$0.002} & \textbf{0.284}{\scriptsize $\pm$0.003} & \textbf{0.337}{\scriptsize $\pm$0.001} \\
 & 192 & 0.361{\scriptsize $\pm$0.002} & 0.383{\scriptsize $\pm$0.001} & \textbf{0.353}{\scriptsize $\pm$0.002} & \textbf{0.375}{\scriptsize $\pm$0.003} & 0.382{\scriptsize $\pm$0.001} & 0.391{\scriptsize $\pm$0.002} & \textbf{0.352}{\scriptsize $\pm$0.003} & \textbf{0.369}{\scriptsize $\pm$0.001} & \textbf{0.381}{\scriptsize $\pm$0.002} & 0.396{\scriptsize $\pm$0.003} & 0.387{\scriptsize $\pm$0.001} & \textbf{0.393}{\scriptsize $\pm$0.002} & 0.333{\scriptsize $\pm$0.003} & 0.367{\scriptsize $\pm$0.001} & \textbf{0.323}{\scriptsize $\pm$0.002} & \textbf{0.362}{\scriptsize $\pm$0.003} \\
 & 336 & 0.388{\scriptsize $\pm$0.002} & 0.404{\scriptsize $\pm$0.003} & \textbf{0.381}{\scriptsize $\pm$0.001} & \textbf{0.397}{\scriptsize $\pm$0.002} & 0.415{\scriptsize $\pm$0.003} & 0.416{\scriptsize $\pm$0.001} & \textbf{0.384}{\scriptsize $\pm$0.002} & \textbf{0.391}{\scriptsize $\pm$0.003} & 0.445{\scriptsize $\pm$0.001} & 0.436{\scriptsize $\pm$0.002} & \textbf{0.403}{\scriptsize $\pm$0.003} & \textbf{0.409}{\scriptsize $\pm$0.001} & 0.369{\scriptsize $\pm$0.002} & 0.387{\scriptsize $\pm$0.003} & \textbf{0.354}{\scriptsize $\pm$0.001} & \textbf{0.381}{\scriptsize $\pm$0.002} \\
 & 720 & 0.463{\scriptsize $\pm$0.003} & 0.448{\scriptsize $\pm$0.002} & \textbf{0.445}{\scriptsize $\pm$0.001} & \textbf{0.436}{\scriptsize $\pm$0.002} & 0.473{\scriptsize $\pm$0.002} & 0.452{\scriptsize $\pm$0.003} & \textbf{0.443}{\scriptsize $\pm$0.001} & \textbf{0.428}{\scriptsize $\pm$0.002} & 0.472{\scriptsize $\pm$0.003} & 0.452{\scriptsize $\pm$0.001} & \textbf{0.447}{\scriptsize $\pm$0.002} & \textbf{0.437}{\scriptsize $\pm$0.003} & 0.424{\scriptsize $\pm$0.001} & 0.418{\scriptsize $\pm$0.002} & \textbf{0.402}{\scriptsize $\pm$0.003} & \textbf{0.411}{\scriptsize $\pm$0.001} \\
\midrule
\multirow{4}{*}{\rotatebox{90}{ETTm2}}
 & 96 & 0.178{\scriptsize $\pm$0.001} & 0.261{\scriptsize $\pm$0.000} & \textbf{0.170}{\scriptsize $\pm$0.001} & \textbf{0.253}{\scriptsize $\pm$0.000} & 0.196{\scriptsize $\pm$0.000} & 0.295{\scriptsize $\pm$0.001} & \textbf{0.173}{\scriptsize $\pm$0.000} & \textbf{0.255}{\scriptsize $\pm$0.001} & 0.187{\scriptsize $\pm$0.001} & 0.266{\scriptsize $\pm$0.000} & \textbf{0.179}{\scriptsize $\pm$0.001} & \textbf{0.260}{\scriptsize $\pm$0.000} & 0.163{\scriptsize $\pm$0.001} & 0.254{\scriptsize $\pm$0.000} & \textbf{0.162}{\scriptsize $\pm$0.001} & \textbf{0.248}{\scriptsize $\pm$0.000} \\
 & 192 & 0.236{\scriptsize $\pm$0.000} & 0.299{\scriptsize $\pm$0.001} & \textbf{0.233}{\scriptsize $\pm$0.000} & \textbf{0.296}{\scriptsize $\pm$0.001} & 0.280{\scriptsize $\pm$0.001} & 0.356{\scriptsize $\pm$0.000} & \textbf{0.238}{\scriptsize $\pm$0.001} & \textbf{0.298}{\scriptsize $\pm$0.000} & 0.256{\scriptsize $\pm$0.000} & 0.308{\scriptsize $\pm$0.001} & \textbf{0.246}{\scriptsize $\pm$0.000} & \textbf{0.302}{\scriptsize $\pm$0.001} & 0.221{\scriptsize $\pm$0.000} & 0.293{\scriptsize $\pm$0.001} & \textbf{0.216}{\scriptsize $\pm$0.000} & \textbf{0.285}{\scriptsize $\pm$0.001} \\
 & 336 & \textbf{0.294}{\scriptsize $\pm$0.001} & \textbf{0.337}{\scriptsize $\pm$0.000} & 0.296{\scriptsize $\pm$0.001} & 0.340{\scriptsize $\pm$0.000} & 0.386{\scriptsize $\pm$0.000} & 0.429{\scriptsize $\pm$0.001} & \textbf{0.304}{\scriptsize $\pm$0.000} & \textbf{0.344}{\scriptsize $\pm$0.001} & 0.323{\scriptsize $\pm$0.001} & 0.351{\scriptsize $\pm$0.000} & \textbf{0.317}{\scriptsize $\pm$0.001} & \textbf{0.349}{\scriptsize $\pm$0.000} & 0.275{\scriptsize $\pm$0.001} & 0.328{\scriptsize $\pm$0.000} & \textbf{0.269}{\scriptsize $\pm$0.001} & \textbf{0.322}{\scriptsize $\pm$0.000} \\
 & 720 & \textbf{0.389}{\scriptsize $\pm$0.000} & \textbf{0.395}{\scriptsize $\pm$0.001} & 0.390{\scriptsize $\pm$0.001} & 0.399{\scriptsize $\pm$0.000} & 0.548{\scriptsize $\pm$0.001} & 0.520{\scriptsize $\pm$0.000} & \textbf{0.470}{\scriptsize $\pm$0.001} & \textbf{0.425}{\scriptsize $\pm$0.000} & \textbf{0.420}{\scriptsize $\pm$0.000} & \textbf{0.406}{\scriptsize $\pm$0.001} & 0.422{\scriptsize $\pm$0.000} & 0.408{\scriptsize $\pm$0.001} & 0.362{\scriptsize $\pm$0.000} & 0.382{\scriptsize $\pm$0.001} & \textbf{0.359}{\scriptsize $\pm$0.000} & \textbf{0.379}{\scriptsize $\pm$0.001} \\
\midrule
\multirow{4}{*}{\rotatebox{90}{Weather}}
 & 96 & 0.164{\scriptsize $\pm$0.003} & 0.210{\scriptsize $\pm$0.005} & \textbf{0.162}{\scriptsize $\pm$0.002} & \textbf{0.207}{\scriptsize $\pm$0.003} & 0.196{\scriptsize $\pm$0.003} & 0.255{\scriptsize $\pm$0.004} & \textbf{0.181}{\scriptsize $\pm$0.002} & \textbf{0.227}{\scriptsize $\pm$0.005} & 0.169{\scriptsize $\pm$0.004} & 0.218{\scriptsize $\pm$0.003} & \textbf{0.162}{\scriptsize $\pm$0.005} & \textbf{0.212}{\scriptsize $\pm$0.002} & \textbf{0.151}{\scriptsize $\pm$0.004} & 0.200{\scriptsize $\pm$0.003} & 0.153{\scriptsize $\pm$0.005} & \textbf{0.194}{\scriptsize $\pm$0.002} \\
 & 192 & 0.210{\scriptsize $\pm$0.002} & 0.253{\scriptsize $\pm$0.004} & \textbf{0.208}{\scriptsize $\pm$0.003} & \textbf{0.249}{\scriptsize $\pm$0.005} & 0.236{\scriptsize $\pm$0.004} & 0.293{\scriptsize $\pm$0.002} & \textbf{0.225}{\scriptsize $\pm$0.005} & \textbf{0.263}{\scriptsize $\pm$0.003} & 0.224{\scriptsize $\pm$0.002} & 0.262{\scriptsize $\pm$0.004} & \textbf{0.217}{\scriptsize $\pm$0.003} & \textbf{0.257}{\scriptsize $\pm$0.005} & \textbf{0.199}{\scriptsize $\pm$0.002} & \textbf{0.243}{\scriptsize $\pm$0.004} & 0.205{\scriptsize $\pm$0.003} & 0.245{\scriptsize $\pm$0.005} \\
 & 336 & 0.265{\scriptsize $\pm$0.002} & 0.292{\scriptsize $\pm$0.004} & \textbf{0.264}{\scriptsize $\pm$0.003} & \textbf{0.291}{\scriptsize $\pm$0.005} & 0.282{\scriptsize $\pm$0.005} & 0.331{\scriptsize $\pm$0.003} & \textbf{0.279}{\scriptsize $\pm$0.004} & \textbf{0.302}{\scriptsize $\pm$0.002} & \textbf{0.282}{\scriptsize $\pm$0.005} & \textbf{0.305}{\scriptsize $\pm$0.003} & 0.284{\scriptsize $\pm$0.004} & \textbf{0.305}{\scriptsize $\pm$0.002} & \textbf{0.248}{\scriptsize $\pm$0.003} & \textbf{0.282}{\scriptsize $\pm$0.005} & 0.257{\scriptsize $\pm$0.004} & 0.290{\scriptsize $\pm$0.002} \\
 & 720 & \textbf{0.339}{\scriptsize $\pm$0.004} & \textbf{0.344}{\scriptsize $\pm$0.003} & 0.343{\scriptsize $\pm$0.005} & 0.349{\scriptsize $\pm$0.002} & \textbf{0.345}{\scriptsize $\pm$0.002} & 0.382{\scriptsize $\pm$0.004} & 0.379{\scriptsize $\pm$0.005} & \textbf{0.366}{\scriptsize $\pm$0.003} & 0.358{\scriptsize $\pm$0.004} & 0.355{\scriptsize $\pm$0.002} & \textbf{0.346}{\scriptsize $\pm$0.005} & \textbf{0.349}{\scriptsize $\pm$0.003} & \textbf{0.323}{\scriptsize $\pm$0.002} & \textbf{0.335}{\scriptsize $\pm$0.004} & 0.325{\scriptsize $\pm$0.005} & 0.339{\scriptsize $\pm$0.003} \\
\midrule
\multirow{4}{*}{\rotatebox{90}{Electricity}}
 & 96 & \textbf{0.156}{\scriptsize $\pm$0.003} & 0.248{\scriptsize $\pm$0.002} & \textbf{0.156}{\scriptsize $\pm$0.001} & \textbf{0.246}{\scriptsize $\pm$0.003} & 0.210{\scriptsize $\pm$0.002} & 0.301{\scriptsize $\pm$0.003} & \textbf{0.164}{\scriptsize $\pm$0.004} & \textbf{0.254}{\scriptsize $\pm$0.002} & 0.173{\scriptsize $\pm$0.003} & 0.275{\scriptsize $\pm$0.004} & \textbf{0.165}{\scriptsize $\pm$0.002} & \textbf{0.267}{\scriptsize $\pm$0.003} & 0.133{\scriptsize $\pm$0.004} & 0.228{\scriptsize $\pm$0.002} & \textbf{0.130}{\scriptsize $\pm$0.003} & \textbf{0.226}{\scriptsize $\pm$0.004} \\
 & 192 & 0.170{\scriptsize $\pm$0.003} & 0.261{\scriptsize $\pm$0.001} & \textbf{0.168}{\scriptsize $\pm$0.002} & \textbf{0.258}{\scriptsize $\pm$0.003} & 0.210{\scriptsize $\pm$0.003} & 0.304{\scriptsize $\pm$0.004} & \textbf{0.172}{\scriptsize $\pm$0.002} & \textbf{0.262}{\scriptsize $\pm$0.003} & 0.192{\scriptsize $\pm$0.004} & 0.289{\scriptsize $\pm$0.002} & \textbf{0.180}{\scriptsize $\pm$0.003} & \textbf{0.278}{\scriptsize $\pm$0.004} & \textbf{0.150}{\scriptsize $\pm$0.002} & 0.244{\scriptsize $\pm$0.003} & \textbf{0.150}{\scriptsize $\pm$0.004} & \textbf{0.240}{\scriptsize $\pm$0.002} \\
 & 336 & 0.185{\scriptsize $\pm$0.004} & 0.277{\scriptsize $\pm$0.001} & \textbf{0.182}{\scriptsize $\pm$0.003} & \textbf{0.273}{\scriptsize $\pm$0.002} & 0.222{\scriptsize $\pm$0.004} & 0.319{\scriptsize $\pm$0.002} & \textbf{0.188}{\scriptsize $\pm$0.003} & \textbf{0.277}{\scriptsize $\pm$0.004} & 0.193{\scriptsize $\pm$0.002} & 0.293{\scriptsize $\pm$0.003} & \textbf{0.184}{\scriptsize $\pm$0.004} & \textbf{0.284}{\scriptsize $\pm$0.002} & 0.167{\scriptsize $\pm$0.003} & 0.261{\scriptsize $\pm$0.004} & \textbf{0.166}{\scriptsize $\pm$0.002} & \textbf{0.258}{\scriptsize $\pm$0.003} \\
 & 720 & 0.225{\scriptsize $\pm$0.004} & 0.310{\scriptsize $\pm$0.003} & \textbf{0.221}{\scriptsize $\pm$0.002} & \textbf{0.307}{\scriptsize $\pm$0.004} & 0.257{\scriptsize $\pm$0.002} & 0.349{\scriptsize $\pm$0.004} & \textbf{0.223}{\scriptsize $\pm$0.003} & \textbf{0.307}{\scriptsize $\pm$0.002} & 0.219{\scriptsize $\pm$0.004} & 0.314{\scriptsize $\pm$0.003} & \textbf{0.207}{\scriptsize $\pm$0.002} & \textbf{0.303}{\scriptsize $\pm$0.004} & 0.207{\scriptsize $\pm$0.003} & 0.294{\scriptsize $\pm$0.002} & \textbf{0.201}{\scriptsize $\pm$0.004} & \textbf{0.289}{\scriptsize $\pm$0.003} \\
\midrule
\multirow{4}{*}{\rotatebox{90}{Traffic}}
 & 96 & 0.480{\scriptsize $\pm$0.007} & 0.298{\scriptsize $\pm$0.004} & \textbf{0.474}{\scriptsize $\pm$0.005} & \textbf{0.293}{\scriptsize $\pm$0.006} & 0.649{\scriptsize $\pm$0.005} & 0.396{\scriptsize $\pm$0.006} & \textbf{0.521}{\scriptsize $\pm$0.004} & \textbf{0.313}{\scriptsize $\pm$0.007} & 0.599{\scriptsize $\pm$0.006} & 0.317{\scriptsize $\pm$0.004} & \textbf{0.595}{\scriptsize $\pm$0.005} & \textbf{0.307}{\scriptsize $\pm$0.006} & 0.428{\scriptsize $\pm$0.007} & 0.274{\scriptsize $\pm$0.004} & \textbf{0.417}{\scriptsize $\pm$0.005} & \textbf{0.263}{\scriptsize $\pm$0.006} \\
 & 192 & 0.499{\scriptsize $\pm$0.005} & 0.300{\scriptsize $\pm$0.006} & \textbf{0.486}{\scriptsize $\pm$0.004} & \textbf{0.293}{\scriptsize $\pm$0.007} & 0.598{\scriptsize $\pm$0.006} & 0.370{\scriptsize $\pm$0.004} & \textbf{0.515}{\scriptsize $\pm$0.005} & \textbf{0.309}{\scriptsize $\pm$0.007} & 0.653{\scriptsize $\pm$0.004} & 0.341{\scriptsize $\pm$0.006} & \textbf{0.633}{\scriptsize $\pm$0.007} & \textbf{0.331}{\scriptsize $\pm$0.005} & 0.440{\scriptsize $\pm$0.004} & 0.278{\scriptsize $\pm$0.006} & \textbf{0.432}{\scriptsize $\pm$0.007} & \textbf{0.270}{\scriptsize $\pm$0.005} \\
 & 336 & 0.507{\scriptsize $\pm$0.005} & 0.301{\scriptsize $\pm$0.006} & \textbf{0.502}{\scriptsize $\pm$0.007} & \textbf{0.295}{\scriptsize $\pm$0.005} & 0.605{\scriptsize $\pm$0.007} & 0.373{\scriptsize $\pm$0.005} & \textbf{0.536}{\scriptsize $\pm$0.006} & \textbf{0.334}{\scriptsize $\pm$0.004} & 0.657{\scriptsize $\pm$0.005} & 0.348{\scriptsize $\pm$0.007} & \textbf{0.590}{\scriptsize $\pm$0.004} & \textbf{0.324}{\scriptsize $\pm$0.006} & 0.455{\scriptsize $\pm$0.005} & 0.283{\scriptsize $\pm$0.004} & \textbf{0.447}{\scriptsize $\pm$0.006} & \textbf{0.277}{\scriptsize $\pm$0.007} \\
 & 720 & 0.563{\scriptsize $\pm$0.004} & 0.319{\scriptsize $\pm$0.007} & \textbf{0.544}{\scriptsize $\pm$0.005} & \textbf{0.312}{\scriptsize $\pm$0.006} & 0.645{\scriptsize $\pm$0.004} & 0.395{\scriptsize $\pm$0.007} & \textbf{0.572}{\scriptsize $\pm$0.005} & \textbf{0.354}{\scriptsize $\pm$0.006} & 0.708{\scriptsize $\pm$0.007} & 0.362{\scriptsize $\pm$0.004} & \textbf{0.694}{\scriptsize $\pm$0.006} & \textbf{0.355}{\scriptsize $\pm$0.005} & 0.488{\scriptsize $\pm$0.006} & 0.303{\scriptsize $\pm$0.005} & \textbf{0.480}{\scriptsize $\pm$0.004} & \textbf{0.297}{\scriptsize $\pm$0.007} \\
\bottomrule
\end{tabular}%
}
\end{table}

Tables~\ref{tab:full-results-a} and~\ref{tab:full-results-b} report the complete results for eight backbones on seven datasets, with mean $\pm$ standard deviation over three independent random seeds to characterize run-to-run variability, which represents the standard reporting practice in long-term time series forecasting.
Across these reported runs, the standard deviations remain small (at most 0.007), supporting the stability of the observed forecasting gains.
Across the 224 dataset--backbone--horizon settings, GRAFT reduces MSE and MAE by approximately 3.88\% and 3.00\% on average, respectively.\footnote{Average relative reduction is computed as $\frac{100}{N}\sum_{i=1}^{N}(e_i^{\mathrm{Base}}-e_i^{\mathrm{GRAFT}})/e_i^{\mathrm{Base}}$, where $e_i$ denotes the corresponding metric and each setting receives equal weight.}

The improvements extend beyond DLinear, which exhibits the largest average gains.
After excluding DLinear, the remaining seven backbones still achieve average MSE/MAE reductions of 2.40\%/1.54\%, with lower errors in 173/196 (88.3\%) and 156/196 (79.6\%) settings, respectively; ties are excluded from the win counts.
These results show that GRAFT's effectiveness is not confined to the backbone with the largest gains.

The 2026 strong baselines PMDformer, Dynamic-TMoE, and XLinear experiments further support its applicability across backbones, with average MSE reductions of 1.30\%, 1.56\%, and 1.63\%, respectively.
When each dataset--backbone result is averaged over the four prediction horizons, GRAFT improves 20/21 MSE comparisons and 17/21 MAE comparisons, or 37/42 (88.1\%) overall.
The positive average gains on all eight backbones support using historical response evidence to complement backbone forecasts.
The magnitude of improvement varies across datasets, with average MSE reductions ranging from 0.74\% on Weather to 6.76\% on ETTh2.
Such variation is consistent with the design of GRAFT, whose benefit depends on whether historically observed local dynamics provide transferable evidence for the current forecast, consistent with the \emph{Predict Backbone, Graft Dynamics} design.


\end{document}